\documentclass{Araya}

\title{Active inference for action-unaware agents}

\author%
{\href{https://orcid.org/0000-0001-8790-2565}{\includegraphics[scale=0.06]{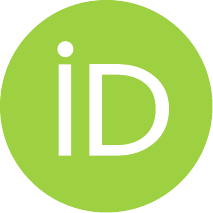}\hspace{1mm}Filippo Torresan}~$^{1, 2}$, \href{https://orcid.org/0000-0001-7014-8770}{\includegraphics[scale=0.06]{orcid.pdf}\hspace{1mm}Keisuke Suzuki}~$^{3}$, \href{https://orcid.org/0000-0002-0186-2687}{\includegraphics[scale=0.06]{orcid.pdf}\hspace{1mm}Ryota Kanai}~$^{1}$, \href{https://orcid.org/0000-0002-6086-4711}{\includegraphics[scale=0.06]{orcid.pdf}\hspace{1mm}Manuel Baltieri}~$^{1,2,}$\footnote[2]{Correspondence e-mail: \texttt{manuel\_baltieri@araya.org}}\\
\vspace{1em} 
\normalfont{\small $^{1}$ Araya Inc., Tokyo, Japan}\\
\normalfont{\small $^{2}$ School of Engineering and Informatics, University of Sussex, Brighton, UK}\\
\normalfont{\small $^{3}$ Center for Human Nature, Artificial Intelligence and Neuroscience (CHAIN), Hokkaido University, Sapporo, Japan}\\
}

\begin{document}

\maketitle
\thispagestyle{firstpagestyle} 

\begin{abstract}
    Active inference is a formal approach to study cognition based on the notion that adaptive agents can be seen as engaging in a process of approximate Bayesian inference, via the minimisation of variational and expected free energies.
    Minimising the former provides an account of perceptual processes and learning as evidence accumulation, while minimising the latter describes how agents select their actions over time.
    In this way, adaptive agents are able to maximise the likelihood of preferred observations or states, given a generative model of the environment.
    In the literature, however, different strategies have been proposed to describe how agents can plan their future actions.
    While they all share the notion that some kind of expected free energy offers an appropriate way to score policies, sequences of actions, in terms of their desirability, there are different ways to consider the contribution of \emph{past} motor experience to the agent's future behaviour.
    In some approaches, agents are assumed to know their own actions, and use such knowledge to better plan for the future.
    In other approaches, agents are unaware of their actions, and must infer their motor behaviour from recent observations in order to plan for the future.
    This difference reflects a standard point of departure in two leading frameworks in motor control based on the presence, or not, of an efference copy signal representing knowledge about an agent's own actions.
    In this work we compare the performances of action-aware and action-unaware agents in two navigations tasks, showing how action-unaware agents can achieve performances comparable to action-aware ones while at a severe disadvantage.
\end{abstract}


\section{Introduction}\label{sec:intro}

Active inference is a framework originally developed in cognitive science and theoretical neuroscience to account for the function(s) of adaptive agents and their nervous systems \parencite{Friston2005a,Friston2009c,Friston2017a,Parr2022b,Pezzulo2024a}.
Different mathematical formulations of its core ideas have been proposed, and have been used to formally account for the adaptive behaviour of agents in different domains, as well as to model neural and behavioural data in computational cognitive neuroscience \parencite{Adams2013a,Buckley2017a,DaCosta2020b,Friston2010b,Friston2012g,Friston2017c,Friston2018a,Heins2020a,Mirza2016a,Mirza2019a,Parr2017a,Parr2020a,Pezzulo2015a,Pezzulo2018a,Seth2016a}.
The framework has also received a lot of attention in philosophy of mind and cognitive science, with its key insights popularised under the banners of predictive processing and prediction error minimisation \parencite{Clark2013b,Clark2016a,Hohwy2013a,Hohwy2020a,Wiese2017c}.

The main idea driving active inference is that information processing in the brain can be explained by predictive activity that approximates a process of hierarchical dynamic Bayesian inference on the hidden states of the environment that produce sensory inputs for the agent \parencite{Lee2003a,Friston2008a,Friston2017a}.
On this view, the dynamics of brain states implement approximate Bayesian inference updates consistent with (the dynamics of) an implicit generative model of, i.e., a joint probability distribution over, sensory signals (observations), motor commands (actions), and internal configurations (states).
In turn, these updates allow an agent to infer its current predicament (perception), to infer the best sequence of actions (policies) to reach favourable states (planning/goal-directed decision making), and to learn what is possible in its eco-niche (learning relevant sensorimotor contingencies) \parencite{Clark2013b,Clark2015a,Pezzulo2015a,Catal2020a,Kaplan2018a,Bruineberg2017a}.

Active inference, its different implementations and ensuing applications have been presented and reviewed extensively in the literature.
For instance, \textcite{Buckley2017a} provides a review of active inference for continuous-time state-space models whereas~\textcite{DaCosta2020b} offer a synthesis of active inference based on the discrete-time framework of partially observable Markov decision processes (POMDPs).
The main difference between the two formulations revolves around the technicalities required to implement a (variational) Bayesian inference scheme according to the dynamical evolution of relevant quantities, occurring either in continuous time or at discrete time steps.

More recently,~\textcite{Smith2022a} presents a more beginner-friendly, yet technical tutorial introduction to the discrete-time formulation, with a special focus on empirical applications, i.e., how to fit active inference models to behavioural and neural data.
The recent implementations of active inference in Python~\parencite{Heins2022a} and in Julia~\parencite{Nehrer2025a}, together with their companion papers and tutorials, represent another excellent entry point and could be read alongside~\parencite{DaCosta2020b} for a deeper understanding of how the mathematical aspects of the theory have been implemented.
Additionally,~\textcite{Lanillos2021a} provides a survey of the approach with a special interest in robotics applications (especially involving the continuous-time formulation) whereas~\textcite{Mazzaglia2022a} offers a similar survey but examining more in detail the connections with related deep learning approaches.
Other works, such as~\textcite{Gottwald2020a}, provide an enlightening mathematical explanation of free-energy minimization, comparing the main versions of the active inference machinery (those that have appeared up to 2020), and also makes a comparison with other Bayesian approaches to adaptive decision-making such as control-as-inference \parencite{Kappen2012a,Toussaint2006a,Todorov2008a,Levine2018a}.
On the other hand,~\textcite{Millidge2021a} provides an introduction to the more foundational notion of the free-energy principle (i.e., tying free-energy minimization to self-organization in certain dynamical systems), from which a theory of sentient behaviour like active inference can be seen to emerge \parencite[see, also,][]{Friston2019b}, while~\textcite{Parr2022b} bring everything together in a thorough and accessible treatment of the approach and its applications.

Inspired by these and other relevant works in the area, in~\cref{sec:ae-interact,sec:approx-bi} we provide a self-contained introduction to the standard active inference framework, including in~\cref{secx:math-background,secx:percep-plan-act} further details and derivations of the active inference equations for perception, action selection and learning (of both transition dynamics and emission maps), with a breakdown of some its more under-explored aspects.
Our goal here is to investigate some assumptions that have appeared in parts of the active inference literature, and their implications for the study of adaptive behaviour.
In particular, we will focus on comparing two implementations for classes of agents we shall define as \textit{action-aware}, inspired by the control-as-inference literature~\cite{Todorov2008a, Kappen2012a, Levine2018a}, and \textit{action-unaware}, more closely related to classical active inference formulations that draw from work on the equilibrium point hypothesis and referent control~\parencite{Feldman2009a,Feldman2016a} to argue that classes of biological agents including humans do now have, or even need, access to explicit information about their motor signals~\parencite{Friston2010b, Friston2012b, Friston2011b, Adams2013a, Baltieri2018b, Baltieri2019c, Baltieri2019e}.
Agents of the first kind know precisely what actions they took in the past and only need to plan for the future, while the latter don't, and thus have to infer sequences of actions that best fit their past, accounting for their observations up to the present, as a pre-condition for inferring what is best to do in the future.
This means that action-aware agents can make use of more knowledge, as they don't need to infer what actions they took in the past.

We will highlight the main difference between these two strategies, related to how the agent's policies are conceived and used in perceptual inference and planning to infer relevant information from observations and evaluate/select future actions.
Action-unaware agents build on the standard treatment presented in~\parencite{DaCosta2020b,Parr2022b}, providing the bedrock of a computational and algorithmic framework in which agents that are unaware of their own actions (executed in the past), are required to infer (among other things) the most likely policy currently followed up until the present from evidence represented by past observations, and to decide subsequently whether to continue performing the same policy in the future.
On the other hand, more recent proposals \parencite{Heins2022a, Friston2021b} adopt a different, action-aware approach on policies by viewing them as sequences of actions in the future, since agents know exactly what actions they took so far (\emph{cf}., efference copy~\parencite{Crapse2008a}).
While the latter has become the most common approach to simulate active inference agents in discrete settings, a clear experimental comparison between the two is still missing.

We provide thus a Python implementation of these two variations of active inference, and unpack results from simulations that compare these two treatments, showing a detailed breakdown of what and how agents learn in simple navigation tasks, shedding light on the extent to which an agent's awareness of its past motor trajectory has an impact on its learning and adaptive behaviour.

In \cref{sec:ae-interact} we start with a brief overview of how the agent-environment interaction is formally modelled in a rigorous manner within the active inference framework.
Then, we explain in detail the optimisation problem that an active inference agent is designed to solve (\cref{sec:approx-bi}), and the various components of the active inference algorithms that go into solving that problem.
With two experiments, we illustrate the typical learning trajectories of action-unaware and action-aware agents in a simple grid-world environment (\cref{sec:experiments}).
We will conclude with a discussion of a few general points about active inference as well as a few more specific ones related to the findings of the experiments (\cref{sec:discussion}).

\section{Formalising the Agent-Environment Interaction}\label{sec:ae-interact}

Active inference proposes a formal approach to characterise cognition and adaptive behaviour starting from a few basic premises:

\begin{enumerate}
    \item biological and artificial agents can persist in a complex and ever changing world if and only if they keep sensory signal within certain \emph{viable} ranges, based on the definition of a set of \emph{preferred states} (or observations),
    \item an agent's internal states parametrise an implicit generative model of the surrounding environment,
    \item all the processes that constitute an agent, from perception to action, can be described as contributing to the minimization of a single quantity, i.e., variational free energy, for a particular class of preferred states and a given generative model.
\end{enumerate}

More formally, in the discrete state-space formulation of active inference, these intuitions are translated into the language of discrete-time partially observable Markov decision processes (POMDPs), which are used to describe mathematically both the relevant parts of the environment (the generative process) and an active inference agent interacting with it (whose dynamics encode parameters' updates consistent with probabilistic beliefs of an implicit generative model).
The characterisation of the agent also requires the specification of a probability distribution over preferred states or observations, thereby constraining its behaviour to be goal-directed.

\begin{definition}[POMDP in active inference, the generative process]\label{def:pomdp}
A POMDP is a six-element tuple, \((\statespace, \obsspace, \actionspace, \Transition, \Emission, \ntime)\), where:

    \begin{itemize}
        \item \(\statespace\) is a finite set of states,
        \item \(\obsspace\) is a finite set of observations,
        \item \(\actionspace\) is a finite set of admissible actions,
        \item \(\statevar_{i}, \obsvar_{i}, \actionvar_{i}\), with \(i \in [1, \ntime]\), are time-indexed random variables defined over the respective spaces, where the time index \(\ntime\) represents a terminal time step,
        \item \(\Transition: \statespace \times \actionspace \rightarrow \Delta(\statespace)\) is a transition function that maps state-action pairs to a probability distribution in the set \(\Delta(\statespace)\) of probability distribution defined over \(\statespace\)
        \item \(\Emission: \statespace  \rightarrow \Delta(\obsspace)\) is an emission function that maps a state to a probability distribution in the set \(\Delta(\obsspace)\) of probability distribution defined over \(\obsspace\).~\footnote{We note also that standard definitions of POMDPs \parencites[Ch.~16]{Russell2021a}[Ch.~34]{Murphy2023a}[Ch.~17]{Sutton2018a} include also a notion of \emph{reward} for an agent, here we don't however include them since active inference specifies targets for an agent in a different way, see \cref{def:aif-agent}.
        Formally, however, this can be easily accommodated in the above definition by stating that our observations \(\obsspace\) include both observations $\mathcal{Y}$ and rewards $\mathcal{R}$ of standard POMDP definitions: $\obsspace = \mathcal{Y} \times \mathcal{R}$.}
    \end{itemize}
\end{definition}

The transition and emission functions map state-action pairs and states to conditional probability distributions that will be denoted by \(\tprob{\cdot}\) and \(\eprob{\cdot}\), respectively.
These distributions define the dynamics of the POMDP where, \(\forall t \in [1, \ntime]\), state and observation random variables are sampled, \(\statevar_{t+1} \sim \tprob{\cdot}\) and \(\obsvar_{t} \sim \eprob{\cdot}\).
In particular, the former can be used to specify the probability that the state random variable at \(t+1\) takes on a certain value, \(\tprob{\statevar_{t+1} = \state_{t+1}}\), given particular values of state and action random variables at the previous time step.
The latter can instead be used to specify the probability that the observation random variable at time step \(t\) takes on a certain value, \(\eprob{\obsvar_{t} = \obs_{t}}\), given a particular value of the state random variable at the current time step.

We assume that \(\statevar_{t+1} \sim \tprob{\cdot}\) and \(\obsvar_{t} \sim \eprob{\cdot}\) correspond to categorical (i.e., discrete) random variables taking on a value from a finite set, i.e., the state space \(\statespace\), with a certain probability.
In active inference, the categorical distributions \(\tprob{\cdot}\) and \(\eprob{\cdot}\) are often indicated by \(\text{Cat}(\stateparams_{t+1})\) and \(\text{Cat}(\obsparams_{t})\), where \(\stateparams_{t+1} \in \mathbb{R}^{\card{\statespace}}\) and \(\obsparams_{t} \in \mathbb{R}^{\card{\obsspace}}\) are vectors of parameters of length \(\card{\statespace}\) and \(\card{\obsspace}\), respectively, and \(\card{\cdot}\) indicates the cardinality of a set (these probability distributions assign a probability to every state/observation in the respective spaces).
Also, it is worth highlighting that we use the words `state' and `observation' to indicate specifically the values of the corresponding random variables, i.e., the elements of the respective spaces, and not the random variable themselves; for the latter we use `state random variable' and `observation random variable'\label{pg:states-vs-srv}.


Given this formal setup, an active inference agent selects a sequence of actions \(\pi \in \Pi\) that can give access to one or more desired states or observations.
This is usually captured by postulating that the agent has goals in the form of preferred states or observations, formalised as the concentration of probability mass on a subset of the support for a probability distribution \(\pprob{\statevar}\) defined over \(\statespace\) or for \(\pprob{\obsvar}\) defined over \(\obsspace\) (\emph{cf}., first premise at the outset of this section).
More precisely, by selecting an appropriate sequence of action, the agent is trying to make the POMDP evolve or update in such a way that \(\pprob{\statevar}\) or \(\pprob{\obsvar}\) will be the “final” probability distribution over states or observations, i.e., the \emph{stationary distribution} of the Markov decision process in question.

In general, however, an agent starts with no knowledge about the POMDP dynamics and emission maps, i.e., of how the next state and observation relate to the current state and action (specified by the two conditional probability distributions introduced above).
Therefore, an active inference agent's task corresponds to the challenge of using observations from an environment (described formally by the POMDP of \cref{def:pomdp}) to learn the parameters of an approximate model that captures both the environment's transition dynamics of (hidden) states and how such states map to given observations.
This is usually called a \emph{generative model} because it allows the agent to predict or \emph{generate} the most likely next state given a state-action pair and the most likely observation resulting from being in that state (\emph{cf}., second premise).
The agent can rely on these predictive capabilities to implement a decision-making strategy to pick an action, or a sequence of actions, that allow it to obtain a preferred state or observation.

Therefore, an active inference agent can be defined in terms of the following components:

\begin{definition}[Active inference agent]
    An active inference agent is described by a five-element tuple, \((\pprob{\cdot}, \policyspace, \mathcal{X}, \decision, \genmodel)\), where:

    \begin{itemize}\label{def:aif-agent}
        \item \(\pprob{\cdot}\) is the preferred probability distribution over states $\statespace$ or observations $\obsspace$,
        \item \(\policyspace\) is a subset of all possible sequences of actions, or \emph{policies}, of length \(\polhorizon\), i.e., \(\Pi \subseteq \actionspace^{\polhorizon}\), where \(\actionspace^{\polhorizon}\) indicates the \(\polhorizon\)-fold Cartesian product \(\actionspace \times \actionspace \times \dots \times \actionspace = \{(\seqactions{\polhorizon}) \, | \, \action_{i} \in \actionspace, \forall i \in [1, \polhorizon]\}\),
        \item $\mathcal{X}$ is either the state space $\statespace$, the observation space $\obsspace$, the policy space $\policyspace$, or possibly others, used as the domain of the decision rule next,
        \item \(\decision: \mathcal{X} \rightarrow  \actionspace\) is a decision rule that outputs an action \(\action \in \actionspace\) given a certain element of the space \(\mathcal{X}\),
        \item \(\genmodel\) is a generative model that approximates the dynamics of the environment, and comes in the form of a POMDP given in~\cref{def:pomdp},
        \item \(\varprob{\cdot}\) is the \emph{variational} distribution that approximates components of the generative model (see \cref{ssec:aif-bayesian-inf,ssec:optim-fe-obj}).
    \end{itemize}
\end{definition}

In the next few sections, we will spell out in some detail what the generative model \(\genmodel\) and the distribution \(\varprob{\cdot}\) involve, and what role they play in an active inference agent.
We will show that, from a collection of environmental observations, it is possible to characterise perception, action, and learning of an embodied active inference agent as particular computational operations with the generative model and the variational distribution to minimise a single objective, i.e., variational free energy (\emph{cf}., third premise).
By doing so, an active inference agent is able to bring about its preferred probability distribution over states/observations.

\section{Sequential Decision-Making with Approximate Bayesian Inference}\label{sec:approx-bi}

\subsection{The Generative Model}\label{ssec:aif-genmodel}
As explained in the previous section, an active inference agent interacts with an environment, described in terms of a POMDP, to move towards a preferred set of states (and/or corresponding observations), as encoded by the probability distribution \(\pprob{\statevar}\) (or \(\pprob{\obsvar})\).
To do so, the agent can only rely on observations received from the environment and its current generative model, \(\genmodel\).
This can be considered as a more or less accurate ``replica'' of the POMDP describing the environment \parencite{DaCosta2020b, Parr2022b} and is defined as follows:

\begin{definition}[Generative model in active inference]\label{def:aif-genmodel}

The generative model \(\genmodel\) of an active inference agent is a POMDP in the sense of \cref{def:pomdp}.
We specify it in more detail using a joint probability distribution over a sequence of state and observation random variables, a policy random variable for sequences of actions, and parameters stored in matrix \(\obsmap\) (for the emission map) and tensor \(\transmap\) (for the transition map), that is, a joint that factors as:



\begin{equation}\label{eq:aif-genmodel-fact}
    \begin{split}
        P(\seqv{\obsvar}{1}{\ntime}, \seqv{\statevar}{1}{\ntime}, \policy, \obsmap, \transmap) = \prob{\policy} \prob{\obsmap} \prob{\transmap} \prob{\statevar_{1}} \prod_{t=2}^{\ntime} \tprobaif \prod_{t=1}^{\ntime} \eprobaif.
    \end{split}
\end{equation}
\end{definition}

The matrix \(\obsmap\) and the tensor \(\transmap\) (one matrix per action) store the parameters for the transition and emission probabilities.
Specifically, \(\obsmap \in \mathbb{R}^{n \times m}\) encodes the probabilities of state-observation mappings at every single time step.
The second dimension of the matrix (number of columns), \(m\), is the number of possible realisations (or values) \(\state_{t} \in \statespace\) of every state random variable \(\statevar_{t}\), \(\forall t \in [1, T]\).
The index of each column can be thought of as picking one of these realisations, i.e., one among the state values \(\state^{1}, \dots, \state^{m}\).
The first dimension of the matrix, \(n\), is the number of possible realizations of an observation random variable \(\obsvar_{t}\), \(\forall t \in [1, T]\).
Similarly, the index of each row picks one of those realisations, i.e., one among the observation values \(\obs^{1}, \dots, \obs^{n}\).
Thus, the \(j\)th column of \(\obsmap\), represented by \(\obsmap_{:,j}\), stores the parameters \(\obsparams_{t}\) of the categorical distribution followed by \(\obsvar_{t}\) conditioned on \(\statevar_{t} = \state^{j}\), i.e., \(\obsvar_{t} \sim P(\cdot|\state^{j}; \obsparams_{t})\) or, equivalently, \(\obsvar_{t} \sim \text{Cat}(\obsparams_{t}|\state^{j})\) (where in both expressions we made explicit the conditioning value of \(\statevar_{t}\) and the parameters).


The tensor \(\transmap \in \mathbb{R}^{\card{\actionspace} \times m \times m}\) stores all the state-transitions probabilities, depending on the action under consideration (indicated by the value of the tensor's first dimension).
Specifically, the matrices \(\transmap^{\action_{1}}, \dots, \transmap^{\action_\mathtt{d}}\), with \(\mathtt{d} = \card{\actionspace}\) specify the most likely distributions over states conditioned on a specific state value and the execution of a specific action (indicated by the superscript).
For each matrix, the row and column dimensions represent the number of possible realisations of a state random variable \(S_{t}\), again meaning that each column and row index identifies a state value among \(s^{1}, \dotsc, s^{m}\).
Thus, the \(j\)th column of a matrix \(\transmap^{x}\), represented by \(\transmap^{x}_{:,j}\), stores the parameters \(\stateparams_{t}\) of the categorical distribution followed by \(\statevar_{t}\) conditioned on \(\statevar_{t-1} = \state^{j}\), i.e., \(\statevar_{t} \sim P(\cdot|\state^{j}_{t-1}, x; \stateparams_{t-1})\) or, equivalently, \(\statevar_{t} \sim \text{Cat}(\stateparams_{t}|\state^{j}_{t-1}, x)\).
In both expressions we made explicit that we are conditioning on a value \(j\) of the state random variable at \(t-1\), i.e., \(\state^{j}_{t-1}\), and on an action \(x \in [\action_{1}, \dots, \action_\mathtt{d}] = \actionspace\).

Note that each column of \(\obsmap\) and \(\transmap^{x}\) can be seen as an output of an approximation (learned by the active inference agent) of the emission map \(\Emission\) and the transition map \(\Transition\), respectively, given a certain input value \(\state^{j}\) for the former and a certain state-action input pair \(\state^{j}, x\) for the latter.
Both maps are assumed to be \emph{time-independent}: the probability that \(s^{j}\) will produce a certain observation and the probability that \(s^{j}\) will lead to a certain state does not change depending on the particular time step indexing the state random variable \(\statevar_{t}\).
Also, note that \(\state_{t}\) and \(\obs_{t}\) stand for one among the values \(\state^{1}, \dots, \state^{m}\) and \(\obs^{1}, \dots, \obs^{n}\), respectively,  and we will use the notation without superscript to refer generically to one of the values of \(\statevar_{t}\) and \(\obsvar_{t}\), when it is superfluous to indicate explicitly that we are working with state and observation values that correspond to particular columns/rows of the matrices just described.

\subsection{Bayesian Inference}\label{ssec:aif-bayesian-inf}
The generative model provides the basis for the following operations:

\begin{enumerate}
    \item determining the most likely past and/or future states, \(\seqv{s}{1}{T} \coloneqq \seqstates{\ntime}\), given a sequence of observations up to the present time step \(t\), \(\seqv{o}{1}{T} \coloneqq \seqobs{t}\), with \(t = \ntime\) when we consider only past states (e.g., at the end of an episode or trajectory)
    \item predicting the most likely next states following the execution of certain actions and given the most probable current state,
    \item determining the most appropriate next action,
    \item updating key parameters to reflect more closely the actual POMDP describing the environment, especially when step 1--3 alone do not allow the agent to reach its goal.
\end{enumerate}

From a computational point of view, these four steps characterise the cognitive life of an active inference agent.
The first one is usually called \emph{perceptual inference}, the second one amounts to planning or \emph{policy inference}, the third one corresponds to the decision-making or \emph{action-selection} stage, and the last corresponds to the \emph{learning} phase.

The ultimate goal of the agent is to perform actions that result in desired observations and/or environmental states.
Observations represent evidence or feedback from the environment for the agent that indicate whether the generative model captures the environmental dynamics well enough to yield accurate predictions and goal-conducive actions.
If not, that evidence can be used to update the generative model to reflect more precisely what would happen in a certain environment.
More precisely, it is formally postulated that an agent is trying to solve an inference problem, corresponding to inference of the most likely (1) hidden states generating an observation, inference of the most likely (2) policy and (3) action given some preferred states, and inference of the most likely (4) parameters of the generative model to make more accurate prediction in the environment.

A principled way of performing inference involves Bayes' rule, which in the POMDP setting under consideration can be spelled out as follows:

\begin{equation}\label{eq:aif-bayes-pomdp}
    \condprob{\seqv{\statevar}{1}{\ntime}, \policy, \obsmap, \transmap}{\seqv{\obsvar}{1}{\ntime}} = \frac{\condprob{\seqv{\obsvar}{1}{\ntime}}{\seqv{\statevar}{1}{\ntime}, \policy, \obsmap, \transmap} \prob{\seqv{\statevar}{1}{\ntime}, \policy, \obsmap, \transmap}}{\prob{\seqv{\obsvar}{1}{\ntime}}},
\end{equation}

where the generative models \(\genmodel\) appears in the numerator, factorised as the product between two probability distributions: (1) the probability of a sequence of observations, conditioned on a sequence of states, the policy random variable, and certain parameters (explained below), and (2) the (prior) probability of the sequence of states, the policy random variable, and the same parameters.
Importantly, the inference problem represented by Bayes' rule in \cref{eq:aif-bayes-pomdp} involves probability distributions over the parameters of \emph{other} probability distributions.
To see this, we can factorise the prior probability distribution \(\prob{\seqv{\statevar}{1}{\ntime}, \policy, \obsmap, \transmap}\) as follows:

\begin{equation}
    \prob{\seqv{\statevar}{1}{\ntime}, \policy, \obsmap, \transmap} = \prob{\seqv{\statevar}{1}{\ntime}}\prob{\policy}\prob{\obsmap}\prob{\transmap^{a_{1}}} \cdots \prob{\transmap^{a_\mathtt{d}}},
\end{equation}

to show explicitly that it involves joint probability distributions over the (vectors of the) matrices \(\obsmap\) and \(\transmap^{a_{1}}, \dots, \transmap^{a_\mathtt{d}}\) (where $\mathtt{d}$ is the number of actions), implying that Bayesian inference will update the parameters of those distributions as well.
In fact, each column of the above matrices should be viewed as a random vector following a Dirichlet probability distribution.
A realization of one of these random vectors forms the set of parameters for \emph{another} distribution, i.e., one of the categorical distributions that specify the state-observation mapping or the action-dependent state transitions.

Formally, \(\prob{\obsmap}\) is a more compact way of writing the joint over random vectors represented by the columns \(\obsmap_{:,i}\) \(\forall i \in [1, m]\) of the matrix, that is: \(\prob{\obsmap} \coloneq \prob{\obsmap_{:,1}, \dots, \obsmap_{:,m}} = \prob{\obsmap_{:,1}}\cdots \prob{\obsmap_{:,m}}\), with \(\obsmap_{:,i} \sim P(\obsmap_{:,i})\), \( \forall i \in [1, m]\).
Further, the latter is defined as a Dirichlet probability distribution, \(P(\obsmap_{:,i}) \coloneq \text{Dir}(\boldsymbol{\alpha}_{i})\), where \(\boldsymbol{\alpha}_{i}\) is a column vector (of the same length as \(\mathbf{A}_{:,i}\)) storing its parameters (note that these should be kept distinct from the elements of \(\obsmap\) which are parameters of categorical distributions instead or, when doing Bayesian inference, are seen as random vectors, whose realizations determine the categorical parameters).
The same analysis applies for the matrices \(\transmap^{a_{1}}, \dots, \transmap^{a_\mathtt{d}}\).
In a nutshell, Bayesian inference consists in updating the Dirichlet parameters \(\boldsymbol{\alpha}_{i}\) and \(\boldsymbol{\beta_{i}}\) for each matrix above, so that new categorical parameters can be sampled from the Dirichlet distributions, replacing the existing elements of the observation mapping and state transition matrices.


By means of a generative model specified as above and a sequence of observations \(\seqv{\obs}{1}{\ntime}\), Bayes' rule in \cref{eq:aif-bayes-pomdp} allows one to derive an approximate posterior distribution over the state random variables, the policy random variable, and the model's parameters, i.e., the probabilities stored in \(\obsmap, \transmap\).
Deriving this posterior distribution is the inference problem the active inference agent has to solve.
Ultimately, this amounts to an update of the probabilistic \emph{beliefs} encoded by the generative model following the acquisition of observational evidence.
However, since finding an analytic solution to \cref{eq:aif-bayes-pomdp} is often intractable, active inference proposes to implement an approximate Bayesian inference scheme revolving around the minimisation of variational free energy.
This quantity is defined in relation to a given generative model, so in this case it can be written as follows (see also~\cref{ssecx:deriv-fe-obj} for a standard derivation):

\begin{equation}\label{eq:aif-fe-pomdp}
    \fe \bigl[ \varprob{\statevar_{1:\ntime}, \policy, \obsmap, \transmap} \bigr] \coloneqq \mathbb{E}_{Q} \Bigl[\log \varprob{\statevar_{1:\ntime}, \policy, \obsmap, \transmap} - \log \prob{ \obsvar_{1:\ntime}, \statevar_{1:\ntime}, \policy, \obsmap, \transmap} \Bigr],
\end{equation}

where \(\varprob{\statevar_{1:\ntime}, \policy, \obsmap, \transmap}\) is known as the \emph{variational posterior}, a probability distribution introduced to approximate the posterior distribution, \(\condprob{\seqv{\statevar}{1}{\ntime}, \policy, \obsmap, \transmap}{\seqv{\obsvar}{1}{\ntime}}\), in \cref{eq:aif-bayes-pomdp} (the outcome of Bayesian inference).

\subsection{Optimization of the Free Energy Objective}\label{ssec:optim-fe-obj}
To minimize the free energy defined in \cref{eq:aif-fe-pomdp}, we make some assumptions about the variational posterior so that the optimization procedure described above becomes more tractable.
If we simply assumed that \(\varprob{\statevar_{1:\ntime}, \policy, \obsmap, \transmap}\) had exactly the same form as \(\condprob{\statevar_{1:\ntime}, \policy, \obsmap, \transmap}{\obsvar_{1:\ntime}}\), making the variational posterior a replica of the actual posterior, one would incur again in issues of computational intractability, similarly to the original problem of determining an analytic solution to \cref{eq:aif-bayes-pomdp}.
In discrete-time active inference, it is thus common to adopt a \emph{mean-field} approximation \parencite{DaCosta2020b}, meaning that the variational posterior is factorised as follows:

\begin{equation}\label{eq:variational-approx}
    \varprob{\statevar_{1:\ntime}, \policy, \obsmap, \transmap} = \varprob{\obsmap} \varprob{\transmap} \varprob{\policy} \prod_{t=1}^{\ntime} \varprob{\statevar_{t}|\policy}.
\end{equation}

By substituting this expression in \cref{eq:aif-fe-pomdp} for the variational posterior, and by considering the factorization of the generative model, we can rewrite the free energy as follows (\emph{cf}.,~\cite{DaCosta2020b}):

\begin{equation}\label{eq:fe-unpacked}
    \begin{split}
        \fe \bigl[\varprob{\statevar_{1:\ntime}, \policy, \obsmap, \transmap} \bigr] = & \infdiv[\Big]{\varprob{\obsmap}}{\prob{\obsmap}} + \infdiv[\Big]{\varprob{\transmap}}{\prob{\transmap}} + \infdiv[\Big]{\varprob{\policy}}{\prob{\policy}} \\
        & + \mathbb{E}_{\varprob{\policy_{k}}} \Biggl[
        \sum_{t=1}^{\ntime} \mathbb{E}_{\varprob{\statevar_{t}|\policy_{k}}} \Bigl[\log \varprob{\statevar_{t}|\policy_{k}} \Bigr] - \sum_{t=1}^{\tau} \mathbb{E}_{\varprob{\statevar_{t}|\policy_{k}} \varprob{\obsmap}} \Bigl[ \log \condprob{\obs_{t}}{\statevar_{t}, \obsmap} \Bigr] \\
        & - \mathbb{E}_{\varprob{\statevar_{1}|\policy_{k}}} \Bigl[ \log \prob{\statevar_{1}} \Bigr] - \sum_{t=2}^{\ntime} \mathbb{E}_{\varprob{\statevar_{t}|\policy_{k}} \varprob{\statevar_{t-1}|\policy_{k}}} \Bigl[\log \condprob{\statevar_{t}}{\statevar_{t-1}, \policy_{k}} \Bigr] \Biggr],
    \end{split}
\end{equation}

where we have singled out the KL divergences between the posterior probability distributions from the variational approximation and the prior probability distributions from the generative model (first three terms), and grouped together all the terms involving one of the variational posteriors \(\varprob{\statevar_{1:\ntime} | \policy_{k}}\), \(k \in [1, \numpolicies]\) where $\numpolicies$ is the number of policies (see~\cref{tab:summary-notation}), inside the expectation \(\mathbb{E}_{\varprob{\policy_{k}}}[\dots]\) (last term), which computes an average with respect to all policies.

Technically, the free energy \(\fe\) is a \emph{functional} (a term from the calculus of variations), i.e., a mapping from a space of functions to (in this case) the real numbers.
Finding its minimum thus consists of looking for particular functions over given variables as opposed to particular values of given variables for a function, as in more traditional optimization problems.
In this case, the functions we are looking for are probability distributions, i.e., the variational posteriors of~\cref{eq:variational-approx}.
Given the assumptions in~\cref{ssec:aif-genmodel}, finding these functions amounts to tweaking the sets of parameters of the variational distribution, until we find those that result in a distribution that minimises the free energy.
Since we are working with discrete probability distributions, there are analytical solutions which can be found by simply setting the gradient of the free energy with respect to each set of parameters to zero, i.e., $\nabla_{\stateparams_{t}} \fe[\varprob{\statevar_{t}|\policy_{k}}] = 0, \nabla_{\boldsymbol{\alpha}} \fe[\obsmap] = 0, \dots$, one set for each probability distribution in question, and solve for the corresponding parameters. In~\cref{secx:percep-plan-act}, we describe in detail some of these solutions.

When the expression in~\cref{eq:fe-unpacked} is optimised with respect to the policy-conditioned variational distributions, \(\varprob{\statevar_{t}|\policy_{k}} \,\forall k \in [1, \numpolicies]\), we can simply focus on the argument of \(\mathbb{E}_{\varprob{\policy_{k}}}[\dots]\) to compute the associated gradient (since that is the only term that contributes to the gradient and by noting that ignoring the expectation does not change the solution of \(\nabla_{\stateparams_{t}} \fe[\varprob{\statevar_{t}|\policy_{k}}] = 0\)).
That argument defines a policy-conditioned free energy:

\begin{equation}\label{eq:policy-cond-fe-annotated}
    \begin{split}
        \fe_{\policy_{k}} \bigl[ \varprob{\statevar_{1:T} | \policy_{k}} \bigr] \coloneq
        & \sum_{t=1}^{\ntime} \mathbb{E}_{\varprob{\statevar_{t}|\policy_{k}}} \Bigl[ \underbrace{\log \varprob{\statevar_{t}|\policy_{k}}}_{\textbf{state log-probabilities}} \Bigr] - \sum_{t=1}^{\tau} \mathbb{E}_{\varprob{\statevar_{t}|\policy_{k}} \varprob{\obsmap}} \Bigl[ \underbrace{\log \condprob{\obs_{t}}{\statevar_{t}, \obsmap}}_{\textbf{observation log-likelihoods}} \Bigr] - \\
        & - \mathbb{E}_{\varprob{\statevar_{1}|\policy_{k}}} \Bigl[ \underbrace{\log \prob{\statevar_{1}}}_{\textbf{state log-probabilities}} \Bigr] - \sum_{t=2}^{\ntime} \mathbb{E}_{\varprob{\statevar_{t}|\policy_{k}} \varprob{\statevar_{t-1}|\policy_{k}}} \Bigl[ \underbrace{\log \condprob{\statevar_{t}}{\statevar_{t-1}, \policy_{k}}}_{\textbf{transition log-likelihoods}}  \Bigr].
    \end{split}
\end{equation}

The update rules for \(\varprob{\statevar_{t}|\policy_{k}} \,\forall k \in [1, \numpolicies]\), derived by taking the corresponding gradient of the expression in \cref{eq:policy-cond-fe-annotated}, define an optimization/inference scheme called \emph{variational message passing} which makes use of past, present and future information to update, in this case, variational probability distributions at different time points along a trajectory.
Following standard treatments in the literature of stochastic processes and (Bayesian) estimation, it is an example of smoothing, to be contrasted with inference (which uses present information only) and filtering (which relies on past and present information), and  prediction (which uses the past only) \cite{Jazwinski1970a, Sarkka2013a}.

From~\cref{eq:fe-unpacked}, one can derive an update rule for the probability distribution over policies, \(\varprob{\policy}\), which guides the agent in the selection of what to do next (its next action).
This update rule is somewhat tweaked in such a way that the agent will sample actions from a policy that both minimise the policy-conditioned and the \emph{expected} free energy (see~\cref{ssecx:planning-efe} for the design choice that introduces expected free energy).
Expected free energy for policy \(\policy_{k}\) and for a single future time step \(t\) can be defined as follow:

\begin{equation}
    \label{eq:efe-unpacked}
    \begin{split}
        \efe_{t}(\policy_{k}) \coloneq
        & \underbrace{\mathbb{E}_{\varprob{\statevar_{t}|\policy_{k}}} \Bigl[\entropy \bigl[\condprob{\obsvar_{t}}{\statevar_{t}} \bigr] \Bigr]}_{\textsc{Ambiguity}} - \underbrace{\mathbb{E}_{\condprob{\obsvar_{t}}{\statevar_{t}} \varprob{\statevar_{t}|\policy_{k}}} \Bigl[\kldiv \bigl[\varprob{\obsmap|\obs_{t}, \state_{t}}| \varprob{\obsmap} \bigr]\Bigr]}_{\textsc{A-Novelty}} \\
        & + \underbrace{\kldiv \bigl[ \varprob{\statevar_{t}|\policy_{k}} | \pprob{\statevar_{t}} \bigr]}_{\textsc{Risk}} - \underbrace{\mathbb{E}_{\varprob{\statevar_{t+1}|\policy_{k}} \varprob{\statevar_{t}|\policy_{k}}} \Bigl[\kldiv \bigl[\varprob{\transmap|\state_{t+1}, \state_{t}}|\varprob{\transmap}\bigr]\Bigr]}_{\textsc{B-novelty}},
    \end{split}
\end{equation}

where the risk term quantifies the divergence between the predicted and preferred state distribution, the ambiguity terms quantifies the uncertainty related to the observation map, and the two novelty terms are expected information gains for the parameters of the observation and the transition maps, thus indicating parts of the generative model that are still inaccurate.
Therefore, we can associate risk with the \emph{instrumental} or \emph{extrinsic} value of a policy, i.e., the extent to which it enables an agent to reach its preferred states, whereas ambiguity and novelty with its \emph{epistemic} or \emph{intrinsic} value, i.e., the extent to which it drives the agent to acquire informative observations (low ambiguity) and visit states that provide new information about the environment (hight novelty).
A policy that minimises expected free energy does so by balancing the pursuit of these different targets, i.e., addressing the exploitation vs.\ exploration dilemma: it makes sure the agent reaches its goals while at the same time exploring sufficiently enough to acquire relevant and useful information about the environment.
Formal details about the minimisation of variational and expected free energies under variational message passing are covered in~\cref{secx:percep-plan-act} for reference.

The components of the free energy in~\cref{eq:fe-unpacked} and~\cref{eq:policy-cond-fe-annotated} involve terms of the variational approximation and of the generative model.
It is important to note, however, that while the agent's generative model (\cref{def:aif-agent}) and generative process (\cref{def:pomdp}) are both POMDPs, they are in general different, they are not ``synchronised''.
To see why, consider when an agent is first put in contact with a new environment: the agent receives observations from a new environment and is trying to make sense of the structure that generates such sensory input, at the beginning its states and parameters are likely not very helpful, but over time they can be optimised so the agent's generative model aligns, or synchronises, with the generative process.
To do so, at every free energy minimisation stage, the agent uses the observations received so far to update the model's parameters, aided by the variational approximation (to overcome the burden of Bayesian inference).
As learning progresses, the generative model will reflect the observation and transition dynamics of the POMDP more accurately (which, recall, is used to describe a particular environment).

\subsection{Action-aware vs.\ action-unaware agents}\label{ssec:unaware-aware-agents}
The policy-conditioned free energy in \cref{eq:policy-cond-fe-annotated} is treated differently depending on whether the agent knows what actions were performed in the past.
This choice has several repercussions for various aspects of active inference, mainly on the notion of policy and on what it means to condition on a policy.

Action-aware agents use a known sequence of actions they performed in the past \((\seqv{a}{1}{\tau - 1})\).
This mean that, in~\cref{eq:policy-cond-fe-annotated}, policy-conditioned variational distributions for past and present time steps, \(\varprob{\statevar_{1} \mid \policy_{k}}, \dots, \varprob{\statevar_{\tau} \mid \policy_{k}}\) for all policies $k \in [1, \numpolicies]$ of length $\ntime - 1$~\footnote{If we are considering an episodic task and \(\ntime\) is the length of an episode, then a policy consists of \(\ntime - 1\) actions because the agent does not execute any action at the last time step.}, are identical, because all policies share the same sequence of actions $\seqv{a}{1}{\tau - 1}$, i.e., the actions that were executed by the agent, but they differ with respect to future actions, $\seqv{a}{\tau}{\ntime-1}$~\footnote{Note that the action the agent takes at the present time step \(\tau\) is part of the future sequence because it is executed by the agent after the perceptual inference and planning stages are over. However, the variational distribution at \(\tau\), i.e., \(\varprob{\statevar_{\tau} \mid \policy_{k}}\), is the same for all policies because it depends on the action taken at \(\tau - 1\).}.
On this view, given a sequence of actions already executed and shared by all policies, perceptual inference corresponds to inferring the divergent future trajectories in state-space afforded by the various policies, as represented by \(\varprob{\statevar_{\tau} | \policy_{k}}, \dots, \varprob{\statevar_{\ntime} | \policy_{k}}\) for all policies \(k \in [1, \numpolicies]\) of length \(\ntime - 1\), while policy inference relies on inferred variational beliefs to score each policy based on expected free energy.
The main implication here is that policy inference for action-aware agents involves updating the probability over policies by differentiating them \emph{only} with respect to their future consequences because all policies share the same past.
Effectively, this means that the number of policies to evaluate shrinks over time, as more knowledge about executed actions is accumulated that removes action sequences that were never performed.
Equivalently, one can also conclude that an agent simply executes a single (known) policy from \(1\) to \(\tau - 1\), $\seqv{a}{1}{\tau - 1}$, and that different policies $\seqv{a}{\tau}{\ntime-1}$ need to be evaluated for future time steps, see also~\cref{sec:discussion}).

In contrast, action-unaware agents must infer the unknown sequence of actions they performed in the past, \((\seqv{a}{1}{\tau - 1})\), before they can successfully plan for the future.
More precisely, for this class of agents, each policy is a distinct sequence of past, present and future actions and therefore it is no longer the case that all policies share the same sequence of past actions.
During perceptual inference, an action-unaware agent will use the policy-conditioned variational distributions, \(\varprob{\statevar_{1} \mid \policy_{k}}, \dots, \varprob{\statevar_{\tau} \mid \policy_{k}}\) for all policies $k \in [1, \numpolicies]$ of length $\ntime - 1$, to represent the likelihood that the hidden sequence of actions it executed, and that generated the sequence of past and present observations, \((\seqv{o}{1}{\tau})\), comes from a policy $\policy_{k}$.
Policy-conditioned free energies will thus grow for policies that do not explain observations collected up to the present and that most likely have not been pursued.
Policy inference on the other hand involves combining the evidence for each policy with the expected free energy to derive an update of the policy probabilities, guiding the selection of what action to perform next.
Thus, policy inference for action-unaware agents involves updating the probability over policies by taking into account their past, present and future consequences (observations) because each policy represents a distinct trajectory over the length \(\ntime\) of an episode (as opposed to a distinct trajectory for the remaining, future \(\ntime-\tau\) time steps of an episode). 


Further algorithmic details on integrating the variational message passing scheme (introduced in \cref{ssec:optim-fe-obj}) and the above perspectives on policies can be found in algorithm~\ref{pscode:aif-action-unaware} for action-unaware agents and algorithm~\ref{pscode:aif-action-aware} for action-aware ones.
In the next sections, we will report findings from simulations of the two types of agents in a T-maze and a Y-maze with episodes characterised by a fixed duration, i.e., finite and fixed horizon episodes.

\section{Experiments}
\label{sec:experiments}

\subsection{Experiment 1: Learning in a T-maze}\label{sec:exp1-Tmaze4}
\begin{figure}[ht]
  \centering
  \includegraphics[width=0.60\textwidth]{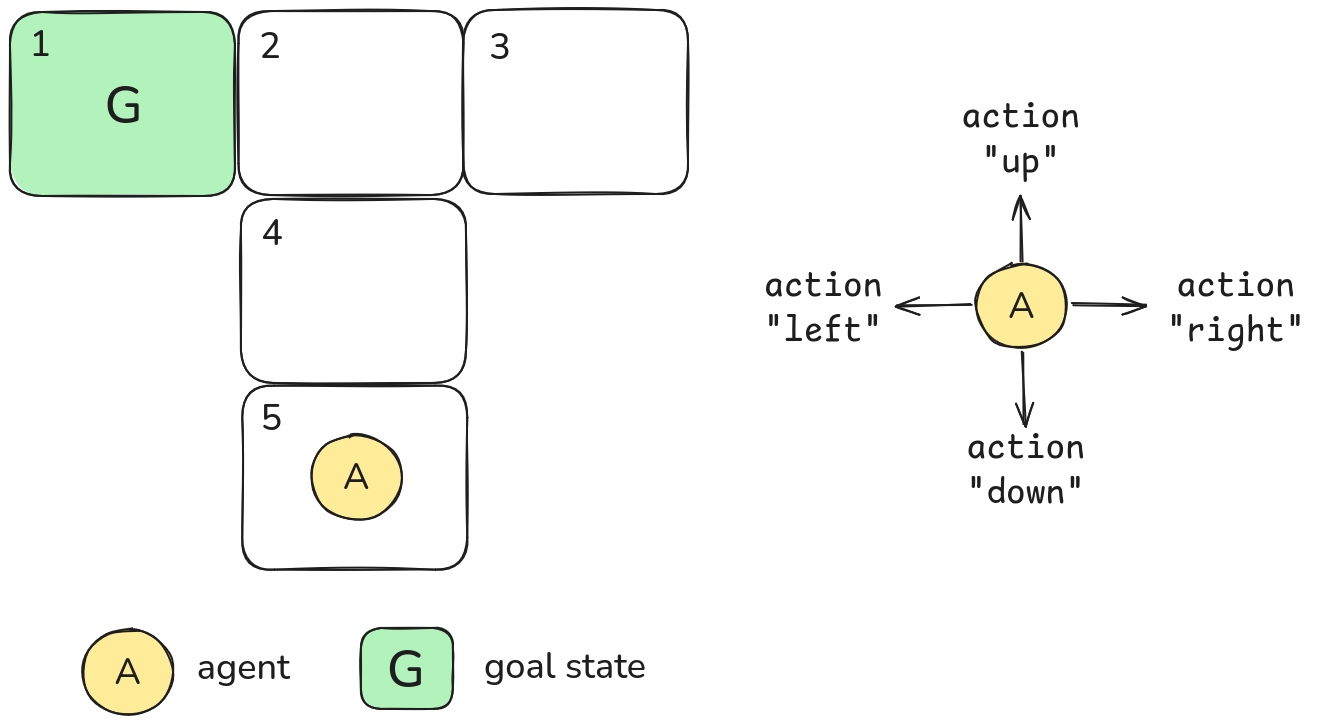}
\caption{Graphical representation of the 4-step T-maze.}\label{fig:tmaze4-env}
\end{figure}

In the first experiment, the agent moves inside the T-maze drawn in~\cref{fig:tmaze4-env}, starting from tile 5 and with a preference to reach the goal state in the left arm, i.e., tile 1.
We simplify the problem structure to be a fully observable MDP (technically, the matrix \(\obsmap\) is not an identity, but it is diagonal and known to the agent), with deterministic but unknown state transitions \(\transmap\).
We trained 10 agents of each type, with and without knowledge of past actions (i.e., action-aware and action-unaware), for 100 episodes (of 4 steps each, with a policy horizon \(\polhorizon = 3\), giving us at most 64 policies to evaluate) in the environment represented in~\cref{fig:tmaze4-env}.

\begin{figure}
  \centering
  \begin{subfigure}{0.45\textwidth}
    \centering
    \includegraphics[width=\textwidth]{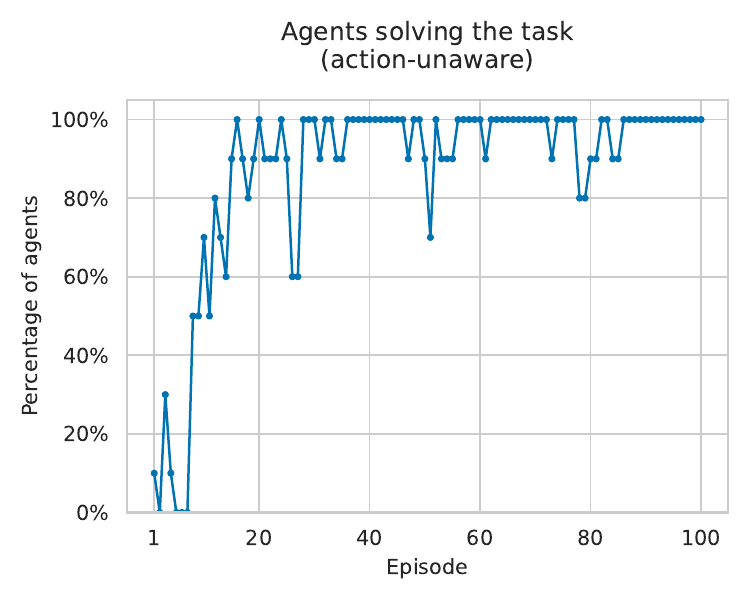}\label{fig:tmaze4-aif-paths-agents-goal}
  \end{subfigure}
  \begin{subfigure}{0.45\textwidth}
    \centering
    \includegraphics[width=\textwidth]{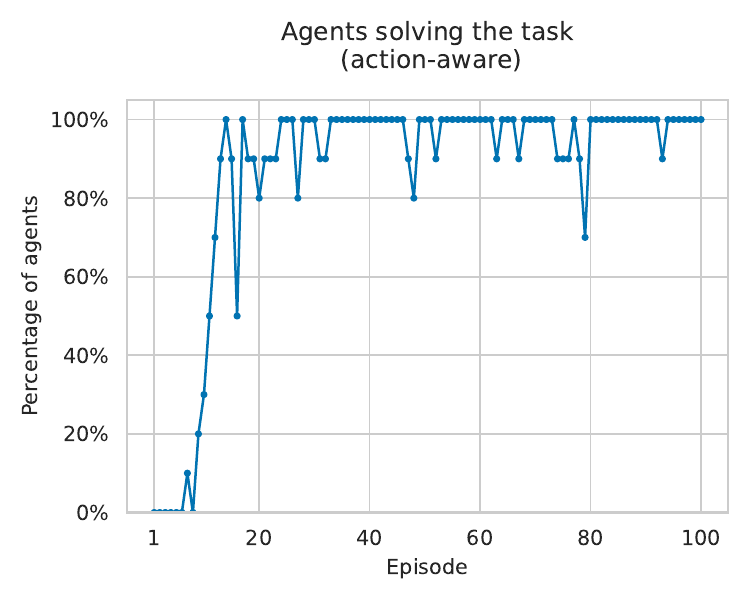}\label{fig:tmaze4-aif-plans-agents-goal}
  \end{subfigure}
  \caption{Percentage of agents reaching the goal state in each episode in the 4-step T-maze (10 total agents).}\label{fig:tmaze4-agents-goal}
\end{figure}

\subsubsection{Results}\label{ssec:tmaze4-results}
In~\cref{fig:tmaze4-agents-goal}, we report the percentage of agents solving the task across episodes.
Both kinds of agents are able to find the optimal policy within the first 20 episodes.
The main differences are their learning speed and pattern.
All action-aware agents fail in the fist 6 episodes, start finding their way to the goal afterwards, and succeed consistently from episode 33 onwards, with a \(100\%\) success rate until the end of the experiment, except for some drops in performance in a handful of episodes.
Despite not having access to past actions, action-unaware agents can also find the optimal policy relatively quickly, with a \(100\%\) success rate from episode 36 onwards but, overall, make a few more mistakes than their action-aware counterparts.
These results indicate that both types of agents were able to learn relevant aspects of the transition model, i.e., the action-dependent transition matrices encoding the (deterministic) effects of performing specific actions in specific states (see~\cref{sssecx:tmaze4-learned-trans-maps} for the learned transition matrices, nearly identical in both types of agents, and compare them with the ground truth ones in \cref{sssecx:tmaze4-gtruth-trans-maps}).
To investigate further whether there other major differences between the two kinds of agents, we examine and compare free energy and expected free energy for the two groups throughout the experiment, the former shedding light on the \emph{perceptual} side of the agents that takes into account its past trajectory, and the latter exploring more in detail its \emph{decision making} side that involves its potential future trajectories.

\paragraph{Perceptual inference}
\begin{figure}[ht]
  \centering
  \begin{subfigure}{0.45\textwidth}\label{fig:tmaze4-aif-paths-marginal-fe-step4}
    \centering
    \includegraphics[width=\textwidth]{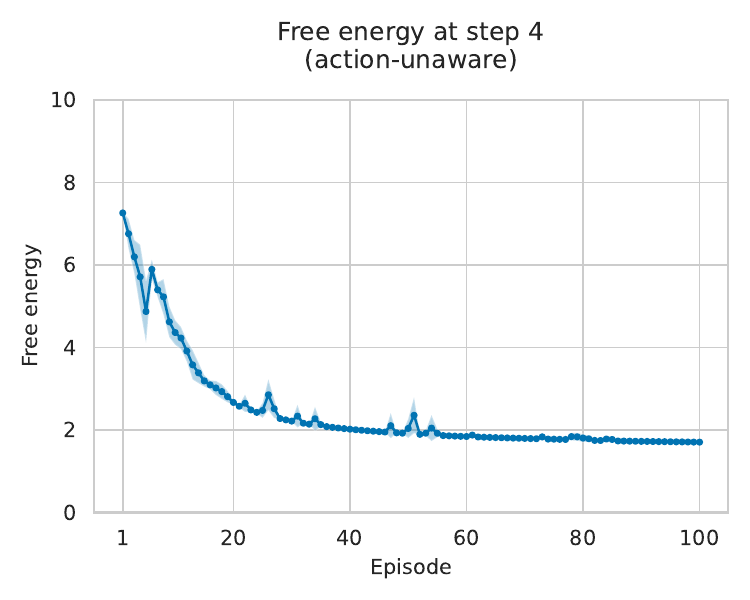}
  \end{subfigure}
  \begin{subfigure}{0.45\textwidth}\label{fig:tmaze4-aif-plans-marginal-fe-step4}
    \centering
    \includegraphics[width=\textwidth]{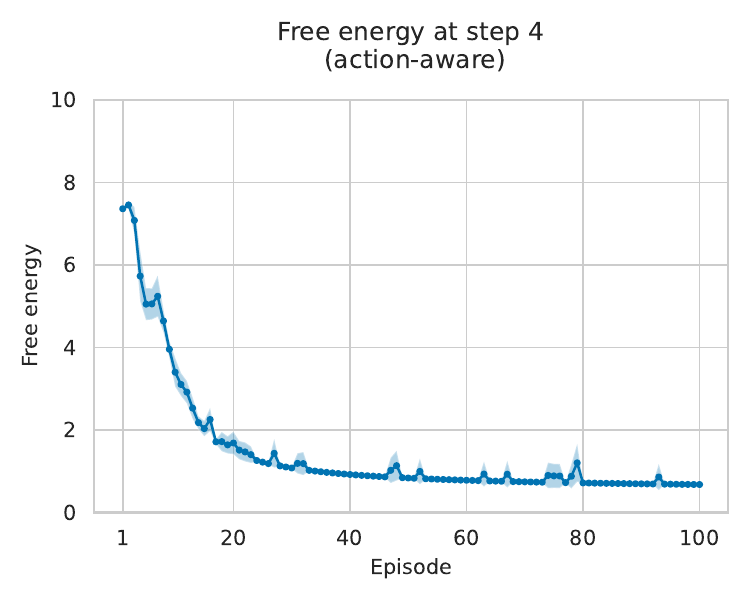}
  \end{subfigure}
  \caption{Free energy across episodes (showing average of 10 agents).}\label{fig:tmaze4-marginal-fe-step4}
\end{figure}

\Cref{fig:tmaze4-marginal-fe-step4} shows the free energy defined in~\cref{eq:fe-unpacked}, which needs to be minimised, at the last step (4th) of each episode for action-aware and action-unaware agents.
We picked the last step because it involves the entire past of an agent within an episode, i.e., the full, episodic trajectory of observations, allowing for a quantification of the agent's uncertainty over the entire time interval a policy covers, and also for the inclusion in the expression for free energy (\cref{eq:fe-unpacked}) of the KL divergence between prior and posterior \(\transmap\) whose parameters are updated at the end of each episode (steps 1, 2 and 3 can be found in~\cref{sssecx:fes-tmaze4}).
The free energy for both agents decreases smoothly but converges at a slightly lower value for action-aware agents than for action-unaware.


\begin{figure}[ht]
  \centering
  \begin{subfigure}{0.45\textwidth}\label{fig:tmaze4-aif-paths-policies-fe-step4}
    \centering
    \includegraphics[width=\textwidth]{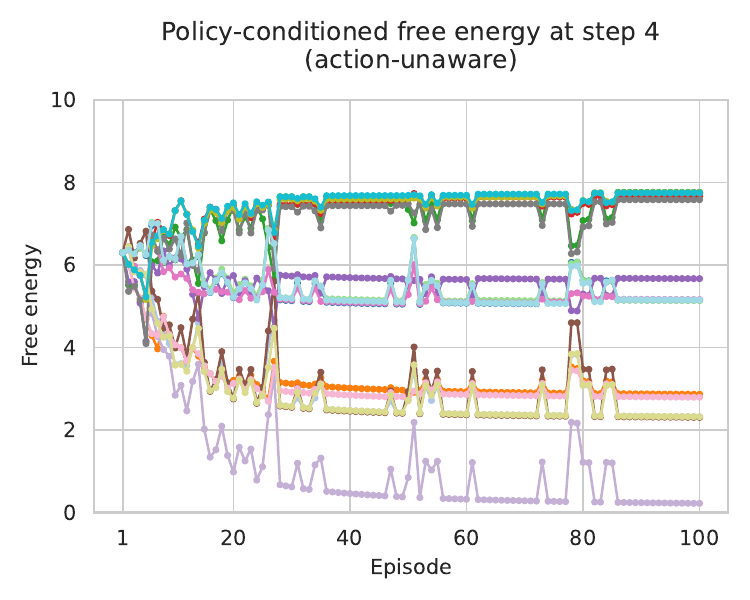}
  \end{subfigure}
  \begin{subfigure}{0.45\textwidth}\label{fig:tmaze4-aif-plans-policies-fe-step4}
    \centering
    \includegraphics[width=\textwidth]{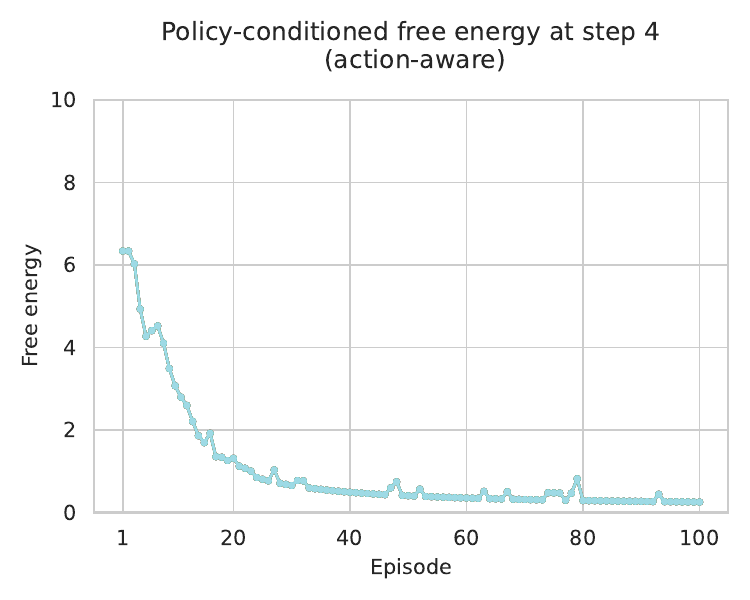}
  \end{subfigure}
  \begin{subfigure}{0.65\textwidth}
    \centering
    \includegraphics[width=\textwidth]{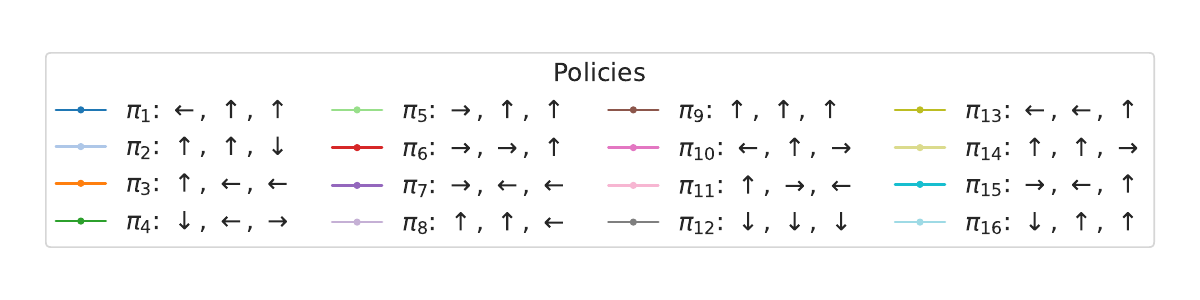}
  \end{subfigure}
  \caption{Policy-conditioned free energies across episodes (showing average of 10 agents).}\label{fig:tmaze4-policies-fe-step4}
\end{figure}

\Cref{fig:tmaze4-policies-fe-step4} shows instead the evolution of the policy-conditioned free energies (see~\cref{eq:policy-cond-fe-annotated}) at step 4 for a subset of all the 64 policies (including the optimal one), for both types of agents (again, figures with the other steps can be found in~\cref{sssecx:tmaze4-pc-fe}).

Starting with action-unaware agents in~\cref{fig:tmaze4-policies-fe-step4} (left), the figure reveals information hidden in the average reported earlier: for the most part, the policy-conditioned free energy that is minimised the most is the one conditioned on the optimal policy.
This makes sense since most action-unaware agents learn to pursue \(\policy_{8}\) from the very few first episodes therefore the associated collection of observation minimises the free energy conditioned on that policy.
However, this free energy is also characterised by several spikes, especially towards the end of the experiment, indicating episodes when the collected evidence is no longer consistent with the actions of \(\policy_{8}\): those are episodes in which the agent has picked a sub-optimal policy, making the associated free energy drop instead.

For action-aware agents instead we note that all the lines essentially overlap on the right side of~\cref{fig:tmaze4-policies-fe-step4}, so that the downward trend captured by the (unconditioned) free energy in~\cref{fig:tmaze4-marginal-fe-step4} is representative of the way the policy-conditioned ones evolve.
More precisely, since the free energy is computed as an average of all the policy-conditioned free energies, the above findings reveal that the values of the latter are all identical.
This should not be surprising because we are considering the last (4th) step: at this point, in action-aware agents, all the “policy-conditioned” free energies are computed by considering the same sequence of actions, the one that produced the state-observation trajectory of a particular episode, and there is no longer a divergent future represented by the future actions of each policy (see how the differences among policy-conditioned free energy decrease across time steps in~\cref{sssecx:tmaze4-pc-fe}).

Overall, this means that observations collected by action-unaware agents correctly minimise the policy-conditioned free energy associated with the policy that was executed, whereas observations collected by action-aware agents simply minimise the variational free energy for the sequence of actions that characterise an entire episode trajectory.
To see if other difference emerges between the two types of agents, we next examine expected free energy and other metrics connected with the planning and action selection mechanisms.

\paragraph{Planning and learning}
\begin{figure}[ht]
  \centering
  \begin{subfigure}{0.45\textwidth}
    \centering
    \includegraphics[width=\textwidth]{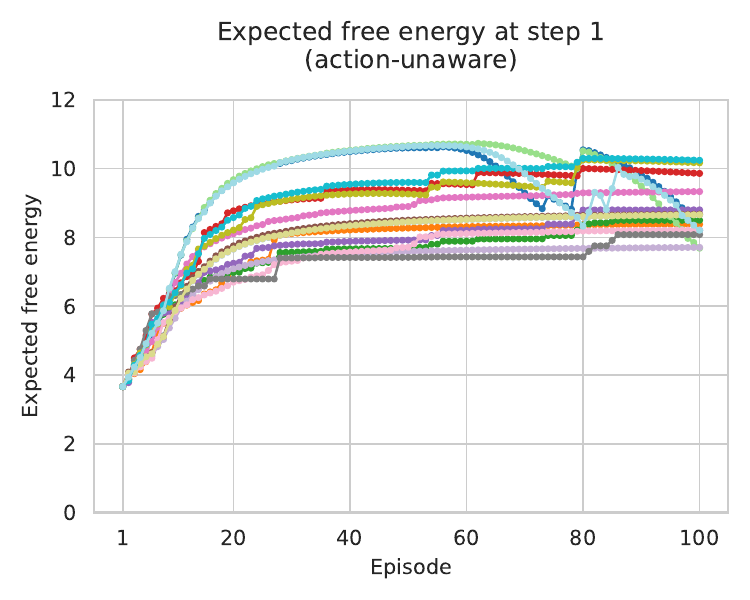}
  \end{subfigure}
  \begin{subfigure}{0.45\textwidth}\label{fig:tmaze4-aif-plans-efe-step0}
    \centering
    \includegraphics[width=\textwidth]{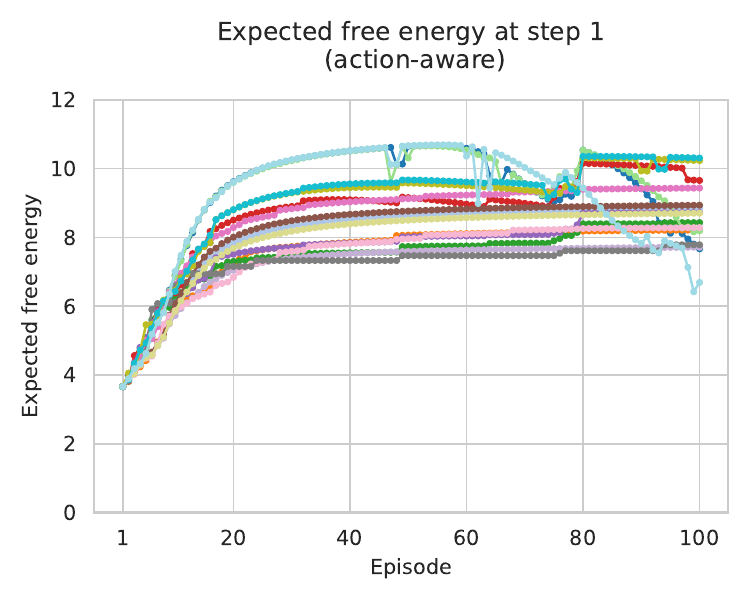}
  \end{subfigure}
  \begin{subfigure}{0.65\textwidth}
    \centering
    \includegraphics[width=\textwidth]{tmaze4_aif_policies_legend}
  \end{subfigure}
  \caption{Expected free energy for each policy across episodes (showing average of 10 agents).
  Notice that we only draw 16 expected free energies, representative of the possible 64.}\label{fig:tmaze4-efe-step0}
\end{figure}

\Cref{fig:tmaze4-efe-step0} shows the total expected free energy for each selected policy across episodes for both types of agents at time step 1.
We chose this step because it involves the sum of all the expected free energy in the future: from time step 1 until the end of the episode (see~\cref{ssecx:planning-efe}), characterising in terms of risk and \(\transmap\)-novelty the entire trajectory afforded by a policy (for completeness, time steps 2 and 3 can be found in~\cref{sssecx:tmaze4-efe-steps}).

The first thing to notice is that expected free energy increases in the first 30--40 episodes for both kinds of agents.
This is surprising because agents ought to minimise it, but can be explained by the fact that, in our experiments, the transition model of an agent (representing the unknown ground truth transitions to be learnt) is randomly initialised at the beginning of the experiment and updated only at the end of each episode.
Since at an early stage the transition model is not a good reflection of ground truth transitions, an agent cannot accurately predict what will happen if a policy is executed.
More precisely, variational beliefs are uniformly initialised at the beginning of each episode, and need to be updated through perceptual inference by using the transition model.
However, if the latter has yet to align with ground truth transitions, the agent will not be able to form accurate beliefs about the locations visited by a certain policy.
As a result, since expected free energy is computed based on those beliefs, its values at step 1 of each episode will not accurately estimate uncertainty/desirability of any sequence of actions for the first few episodes.
Thus, while agents are still learning the transition model, expected free energy increases for each policy until it converges to a value that scores policies more precisely in the current environment, depending on the agent's preferences and the accuracy of its variational beliefs.
To see how different components of this quantity evolve over time, in our simplified setup with no ambiguity and constant A-novelty, see~\cref{sssecx:tmaze4-efe-breakdown}.

\begin{figure}[ht!]
  \centering
  \begin{subfigure}{0.45\textwidth}
    \centering
    \includegraphics[width=\textwidth]{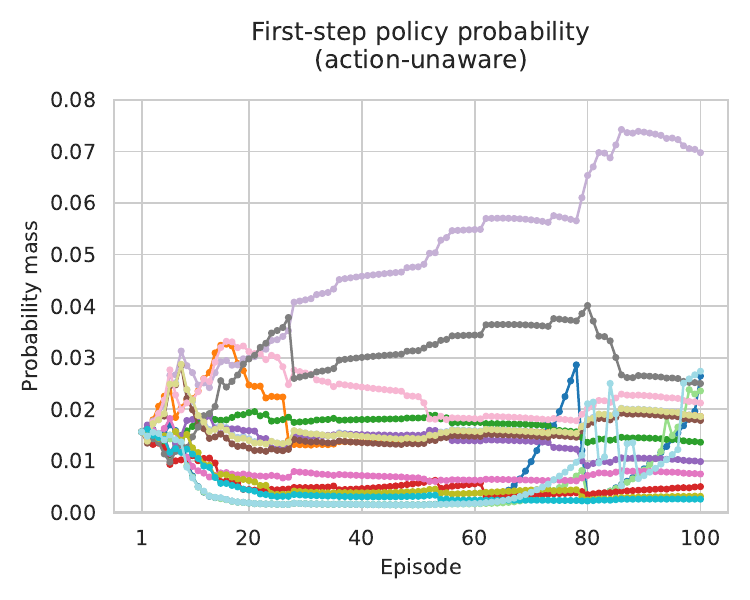}
  \end{subfigure}
  \begin{subfigure}{0.45\textwidth}
    \centering
    \includegraphics[width=\textwidth]{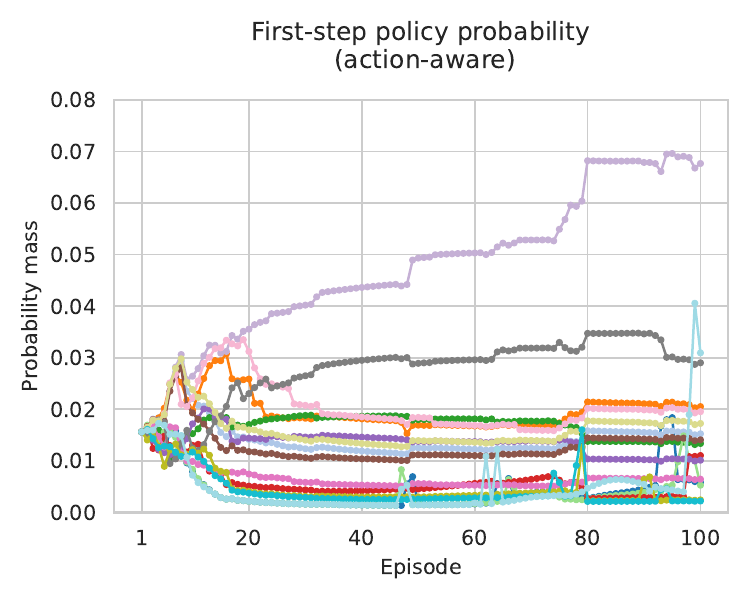}
  \end{subfigure}
  \begin{subfigure}{0.65\textwidth}
    \centering
    \includegraphics[width=\textwidth]{tmaze4_aif_policies_legend}
  \end{subfigure}
  \caption{Policies probabilities at step 1 of each episode (showing average of 10 agents).
  Notice we only draw 16 representative policies out of the possible 64.}\label{fig:tmaze4-first-pi-probs}
\end{figure}

Expected free energy also plays a significant role in the update of the probabilities over policies at each step, which are obtained as a softmax of the negative sum of expected free energies and policy-conditioned free energies (see~\cref{ssecx:planning-efe} for more on expected free energy, and~\cref{eq:fe-policy-update-norm} for the softmax part specifically).
To see the contributions of expected free energies, and their balance against policy-conditioned free energies, we next look at~\cref{fig:tmaze4-first-pi-probs}, showing the first-step policy probabilities for a subset of all the available policies, including the optimal one, for each type of agent.
For both agents, we observe that the optimal policy, \(\policy_{8}\) in the figure, is correctly selected and start to becomes more probable than the others after approximately 20 episodes.
Some sub-optimal policies also become increasingly probable over time, though never enough to surpass the optimal one (e.g., consider the spikes of probability mass for the blue and cyan policies in~\cref{fig:tmaze4-first-pi-probs}).
Further investigations into the underlying reasons for this pattern are left for future work.


\begin{figure}[H]
\centering
\begin{subfigure}{0.45\textwidth}
    \centering
    \includegraphics[width=\textwidth]{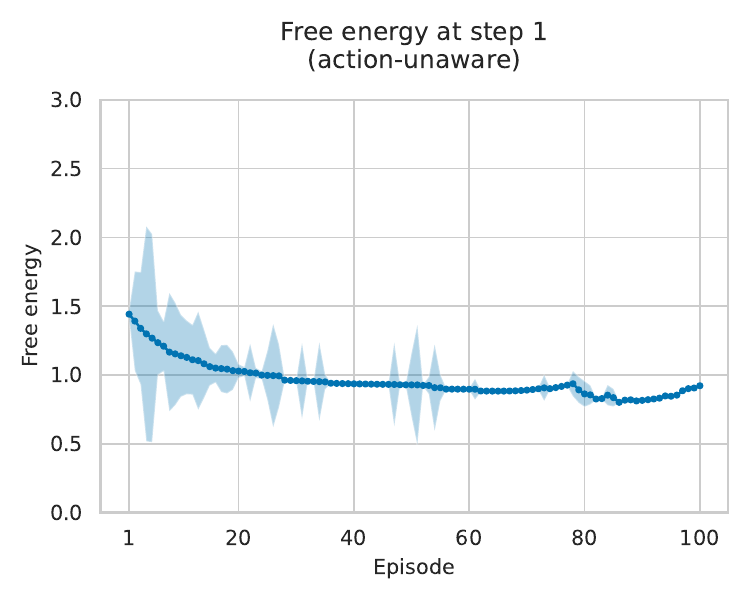}
    \caption{Free energy.}\label{fig:tmaze4-aif-paths-marginal-fe-step1}
  \end{subfigure}
\begin{subfigure}{0.45\textwidth}
    \centering
    \includegraphics[width=\textwidth]{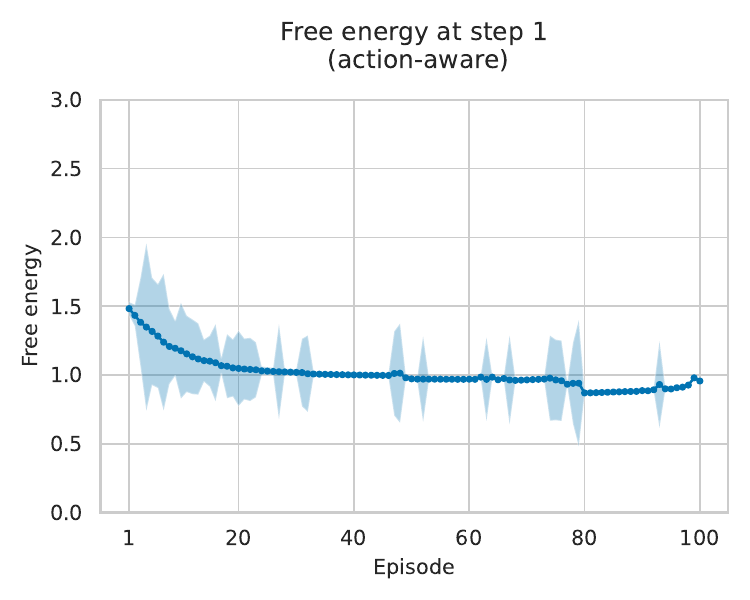}
    \caption{Free energy.}\label{fig:tmaze4-aif-plans-marginal-fe-step1}
  \end{subfigure}
\caption{Free energy at step 1 across episodes (showing average of 10 agents).}\label{fig:tmaze4-marginal-fe-step1}
\end{figure}

Overall, when we consider expected free energies and policy probabilities at the first step of an episode, there is no significant difference between the two types of agent.
This is to be expected because at the beginning of an episode both agents perform perceptual inference, planning, and the update of policy probabilities on the same footing, i.e., there are no significant differences between the respective policy-conditioned free energies.
We have also observed that at a later stage in the experiment both agents become less accurate in predicting the consequences of certain future action sequences, with this phenomenon appearing more marked in action-aware agents (see plots in~\cref{fig:tmaze4-marginal-fe-step1}).


\subsection{Experiment 2: Learning in a grid world}\label{ssec:exp2-gridw9}

\begin{figure}[ht!]
  \centering
  \includegraphics[width=0.60\textwidth]{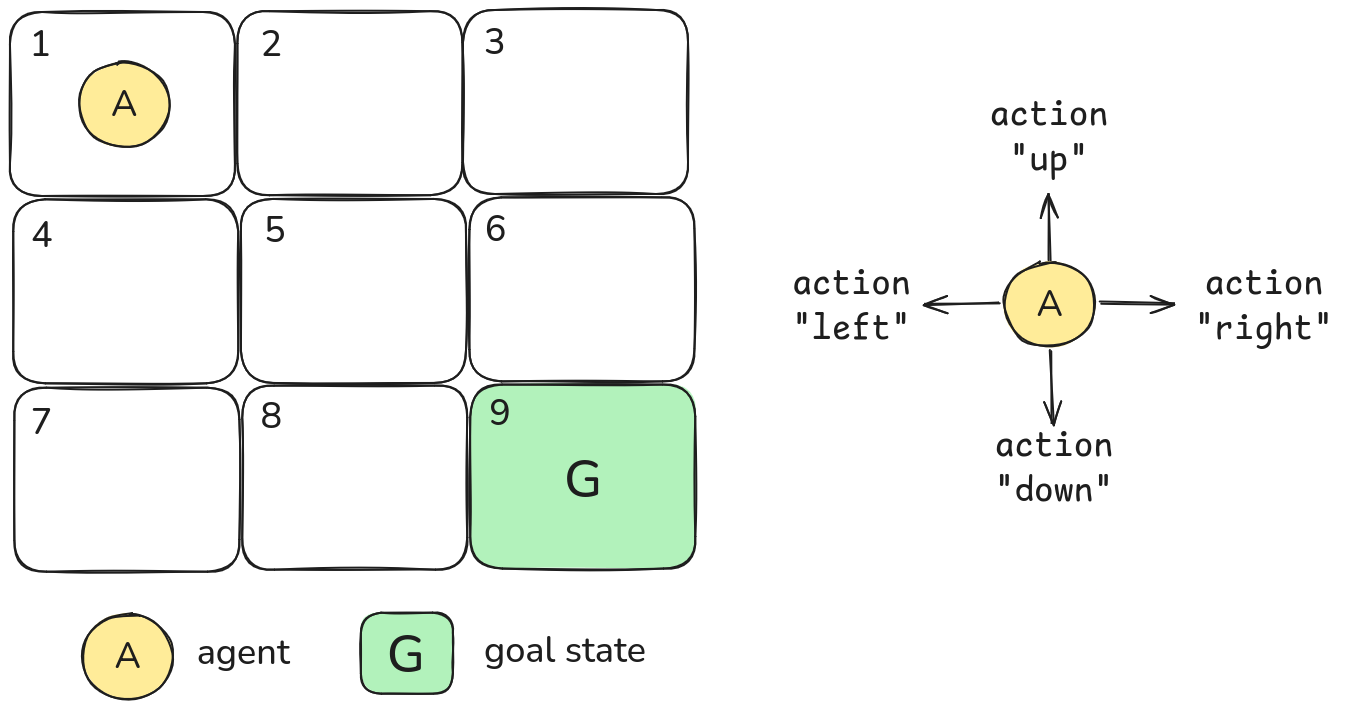}
\caption{Graphical representation of the 5-step grid world.}\label{fig:gridw9-env}
\end{figure}

In the second experiment, we consider an environment with a larger state-space, a \(3 \times 3\) grid world, as depicted in~\cref{fig:gridw9-env}.
While only slightly bigger in terms of states, the policy space in this environment is much larger and includes multiple optimal policies, which could in principle affect our active inference agents.
The agent starts in tile 1 and its goal, tile 9 in the bottom right corner, is encoded as the most preferred state (target location).
Once again, the problem is simplified to be a fully observable MDP (with \(\obsmap\) diagonal and known to the agent), with deterministic but unknown state transitions \(\transmap\).
Here too, we trained 10 agents of each kind, action-unaware and action-aware, for 180 rather than 100 episodes (of 5 steps each, with a policy horizon \(\polhorizon = 4\), giving us at most 256 policies to evaluate) to allow our metrics to converge.

\subsubsection{Results}\label{ssec:exp2-results}
\begin{figure}[ht]
  \centering
  \begin{subfigure}{0.45\textwidth}
    \centering
    \includegraphics[width=\textwidth]{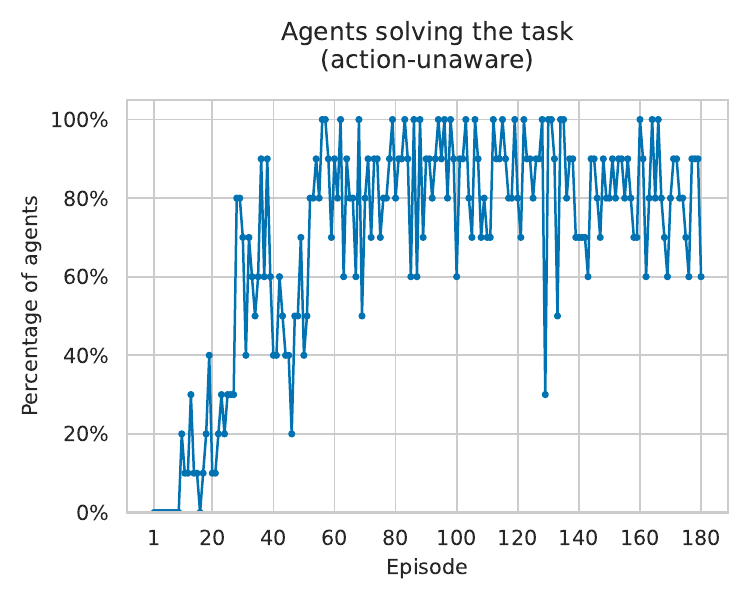}
  \end{subfigure}
  \begin{subfigure}{0.45\textwidth}
    \centering
    \includegraphics[width=\textwidth]{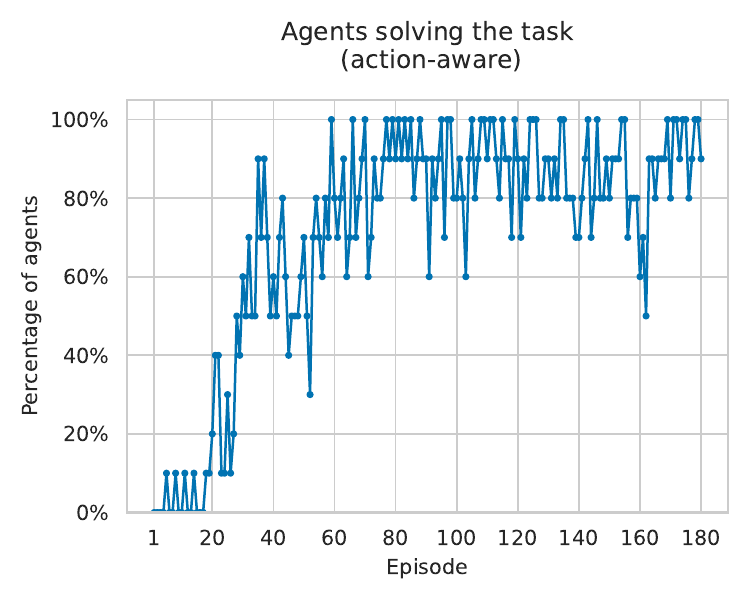}
  \end{subfigure}
  \caption{Percentage of agents reaching the goal state in each episode in the 5-step grid world (10 total agents).}\label{fig:gridw9-agents-goal}
\end{figure}

Similarly to \cref{ssec:tmaze4-results}, we start by comparing the percentage of agent solving the task across episodes in \cref{fig:gridw9-agents-goal}.
Again, we note that both kinds of agent display a similar learning pattern, with agents taking longer to find one of the optimal policies (there are 6 in total this time) due to the larger state-space and number of available policies.
The percentage of successful agents grows until episode 38 and 37, when a \(100\%\) success rate is hit in action-unaware and action-aware agents, respectively, and then drops afterwards to values below \(50\%\) in some episodes, with action-unaware agents registering the more dramatic dips.
Both kinds of agent quickly recover and the success rate remains above \(60\%\) for the most part from around episode 60 until the end of the experiment, with more drops in performance (i.e., to and below \(60\%\)) for both kinds of agent in a handful of episodes.
These results again indicate that both types of agents successfully learned relevant aspects of the transition matrices (see~\cref{sssecx:gridw9-learned-trans-maps} for the learned transition matrices, nearly identical in both types of agents, and compare them with the ground truth ones in~\cref{sssecx:gridw9-gtruth-trans-maps})

\begin{figure}[ht!]
  \centering
  \begin{subfigure}{0.45\textwidth}
    \centering
    \includegraphics[width=\textwidth]{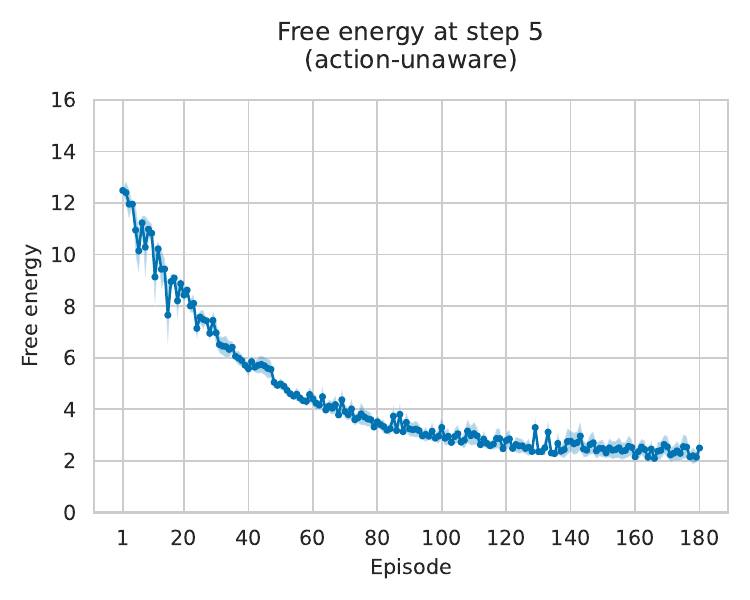}
  \end{subfigure}
  \begin{subfigure}{0.45\textwidth}
    \centering
    \includegraphics[width=\textwidth]{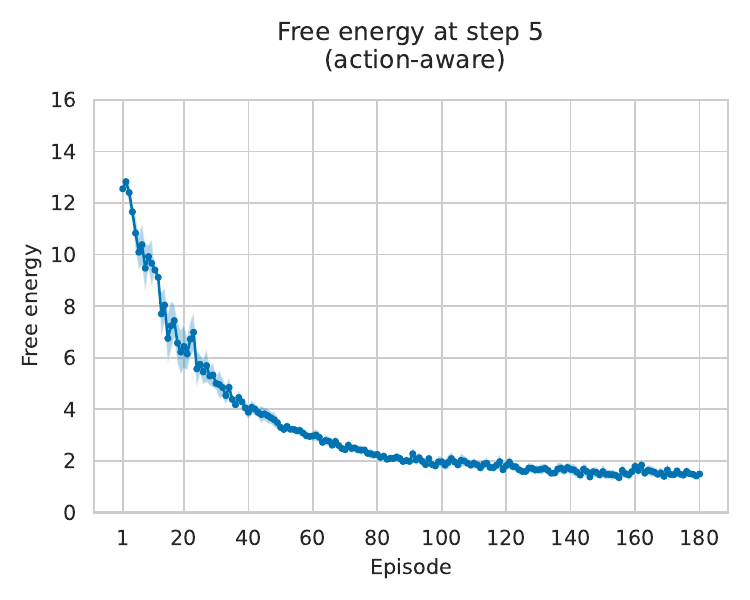}
  \end{subfigure}
  \caption{Free energy across episodes (showing average of 10 agents).}\label{fig:gridw9-marginal-fe-step5}
\end{figure}

\paragraph{Perceptual inference}
The average free energies at the step 5 (last step), see~\cref{fig:gridw9-marginal-fe-step5}, are predictably minimised, but once again hide some relevant information that can be unpacked by showing policy-conditioned free energies (steps 1, 2, 3, and 4 can be found in~\cref{sssecx:fes-gridw9}).

\begin{figure}[ht!]
  \centering
  \begin{subfigure}{0.45\textwidth}\label{fig:gridw9-aif-paths-policies-fe-step5}
    \centering
    \includegraphics[width=\textwidth]{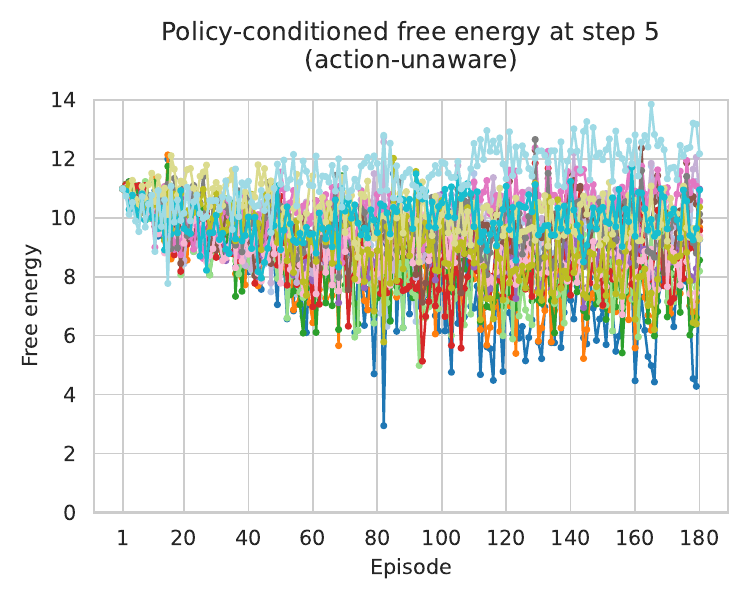}
  \end{subfigure}
  \begin{subfigure}{0.45\textwidth}\label{fig:gridw9-aif-plans-policies-fe-step5}
    \centering
    \includegraphics[width=\textwidth]{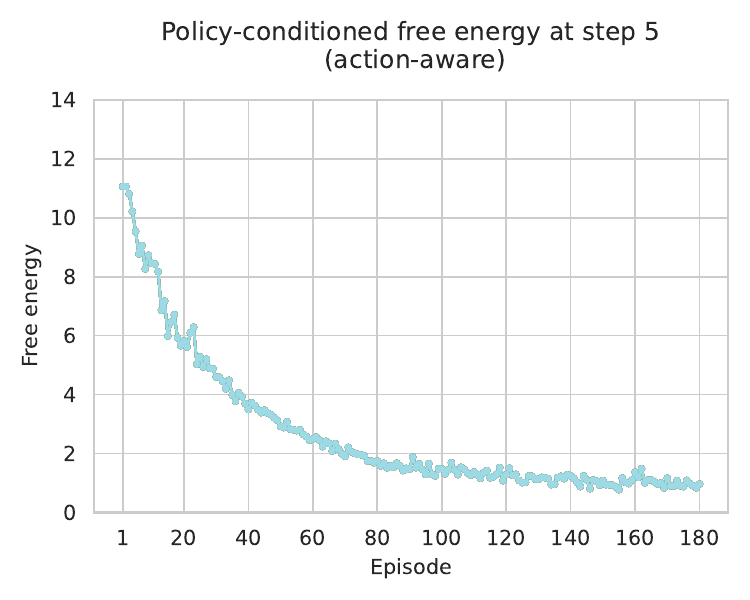}
  \end{subfigure}
  \begin{subfigure}{0.65\textwidth}
    \centering
    \includegraphics[width=\textwidth]{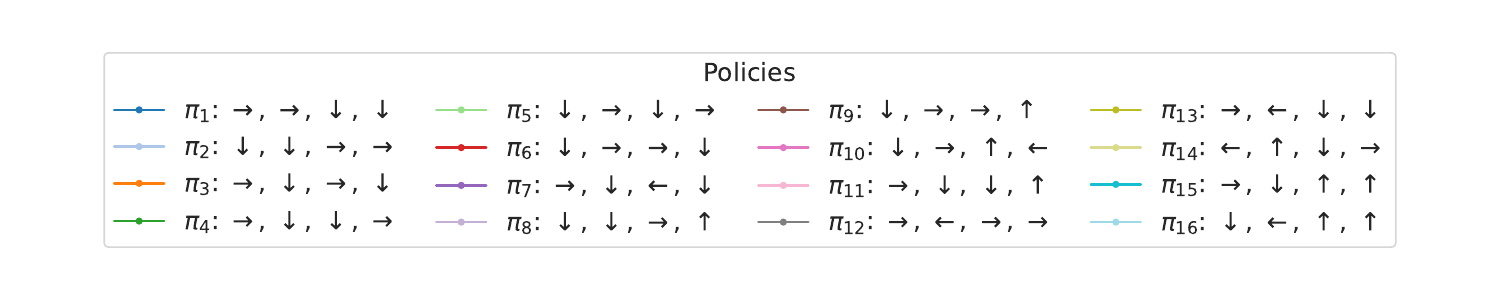}
  \end{subfigure}
  \caption{Policy-conditioned free energies across episodes (showing average of 10 agents).}\label{fig:gridw9-policies-fe-step5}
\end{figure}

\begin{figure}[ht!]
  \centering
  \begin{subfigure}{0.45\textwidth}\label{fig:gridw9-aif-paths-first-pi-probs-offset0}
    \centering
    \includegraphics[width=\textwidth]{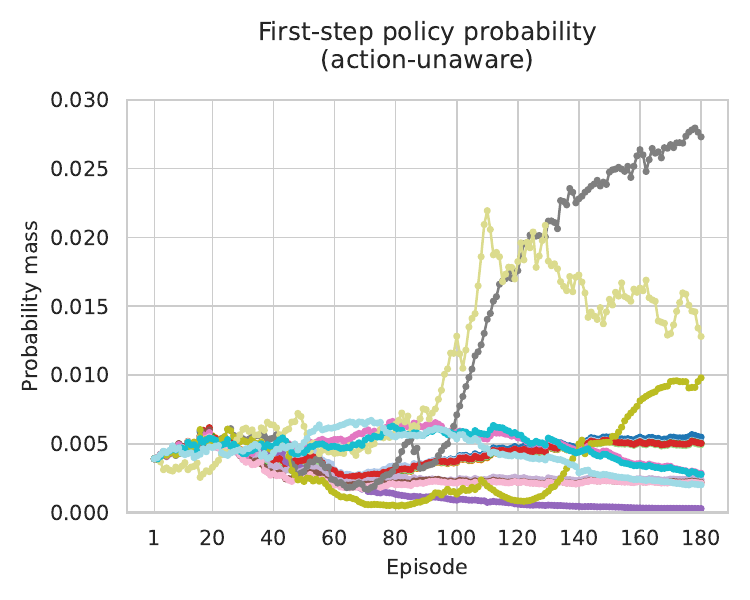}
  \end{subfigure}
  \begin{subfigure}{0.45\textwidth}\label{fig:gridw9-aif-plans-first-pi-probs-offset0}
    \centering
    \includegraphics[width=\textwidth]{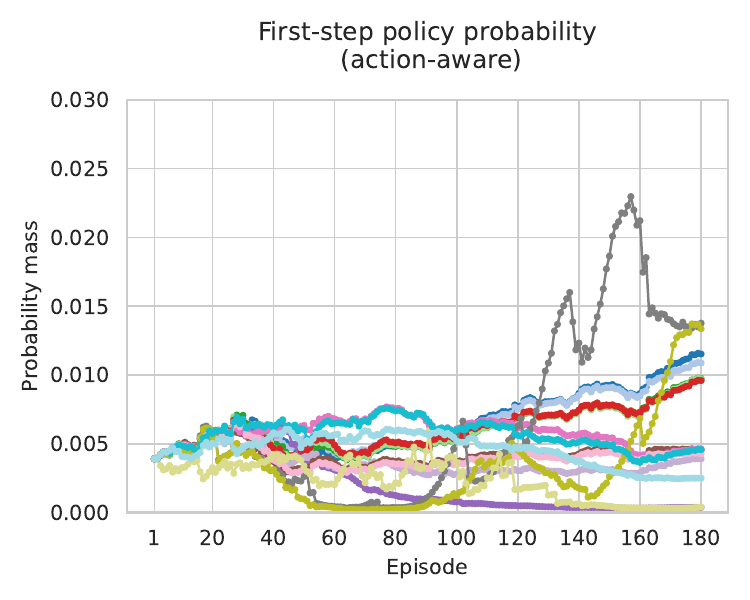}
  \end{subfigure}
  \begin{subfigure}{0.65\textwidth}
    \centering
    \includegraphics[width=\textwidth]{gridw9_aif_policies_legend}
  \end{subfigure}
  \caption{Policies probabilities at the first step of each episode (showing average of 10 agents).
  Notice we only draw 16 representative policies out of the possible 256.}\label{fig:gridw9-first-pi-probs}
\end{figure}

For the policy-conditioned free energies at step 5,~\cref{fig:gridw9-policies-fe-step5}, we selected 16 policies (among the 256) including the 6 optimal policies that lead to the goal state (again, figures with the other steps can be found in~\cref{sssecx:gridw9-pc-fe}).
As seen in the T-maze experiment, in action-aware agents the downward trend of the (unconditioned) free energy in~\cref{fig:gridw9-marginal-fe-step5} is representative of the way the policy-conditioned ones evolve in the right plot of~\cref{fig:gridw9-policies-fe-step5} (i.e., all the policy-conditioned free energies for the selected policies overlap).
In contrast, all the visualised policy-conditioned free energy of action-unaware agents, in the left plot of~\cref{fig:gridw9-policies-fe-step5}, fluctuate considerably throughout the experiment, with none of the optimal policies attaining a consistent decrease of the associated free energy.
The reason for that is that action-unaware agents have discovered all the optimal policies, each offering an alternative route to reach the goal state, and assigned them equal probability mass (see~\cref{fig:gridw9-first-pi-probs}).
Therefore, at the beginning of an episode, agents can randomly choose among alternative paths to the goal, resulting in the minimisation of different policy-conditioned free energies at the end of each episode (recall that for action-unaware agents, a policy-conditioned free energy at the end of an episode is minimised when the observations the agent has received are consistent with having followed the policy in question).




\begin{figure}[ht!]
  \centering
  \begin{subfigure}{0.45\textwidth}\label{fig:gridw9-aif-paths-efe-step0}
    \centering
    \includegraphics[width=\textwidth]{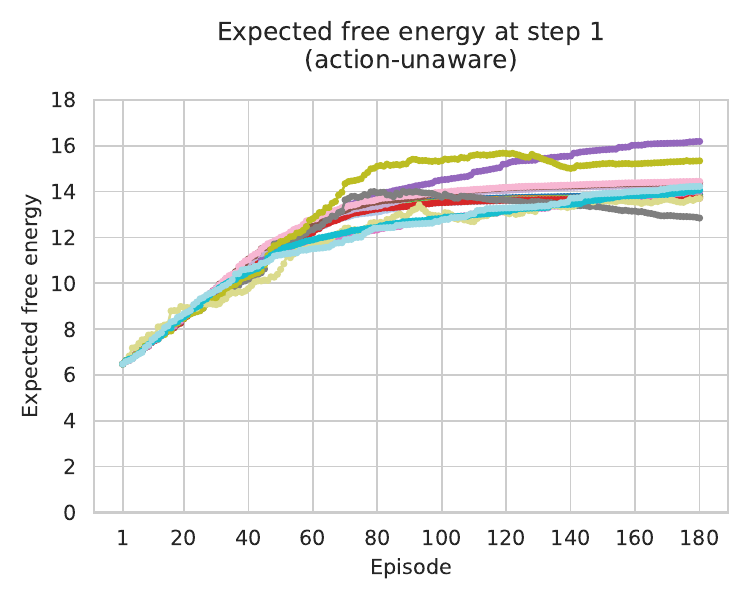}
  \end{subfigure}
  \begin{subfigure}{0.45\textwidth}\label{fig:gridw9-aif-plans-efe-step0}
    \centering
    \includegraphics[width=\textwidth]{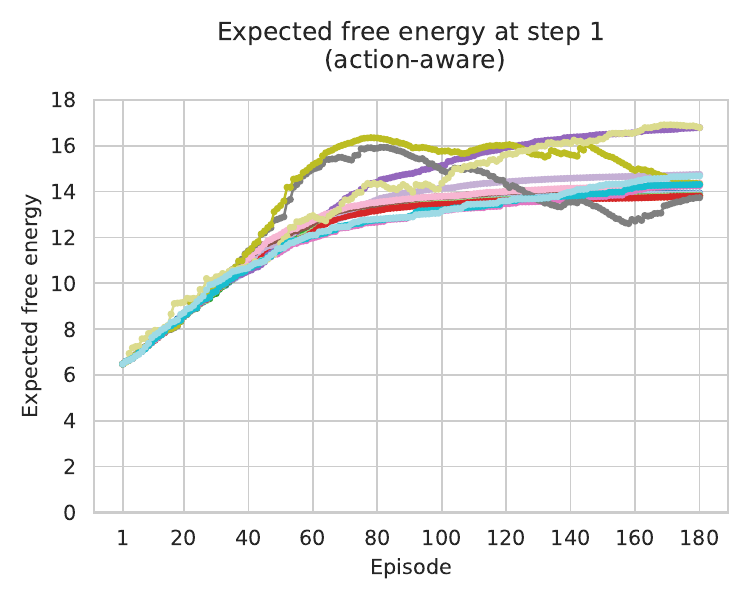}
  \end{subfigure}
  \begin{subfigure}{0.65\textwidth}
    \centering
    \includegraphics[width=\textwidth]{gridw9_aif_policies_legend}
  \end{subfigure}
  \caption{Expected free energy for each policy across episodes (showing average of 10 agents).
  Notice that we only draw 16 expected free energies, representative of the possible 256.}\label{fig:gridw9-efe-step0}
\end{figure}

\begin{figure}[H]
  \centering
  \begin{subfigure}{0.45\textwidth}
    \centering
    \includegraphics[width=\textwidth]{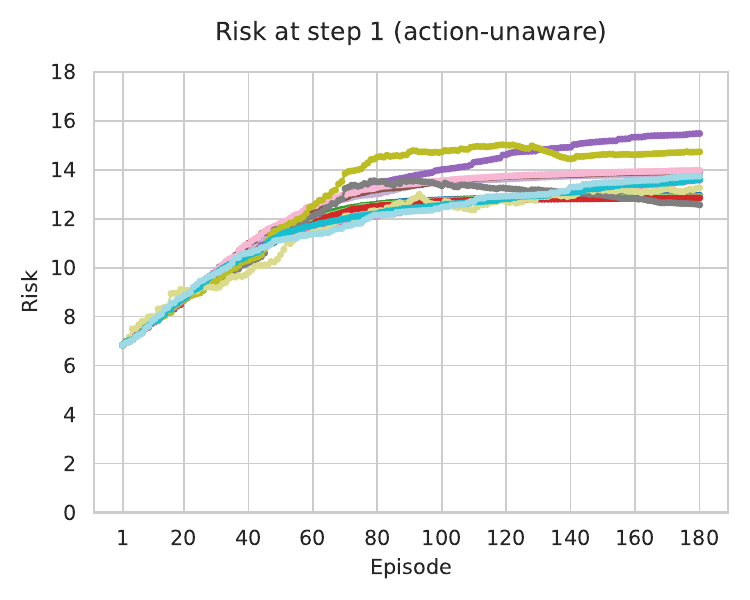}
    \caption{Risk}\label{fig:gridw9-aif-paths-efe-risk}
  \end{subfigure}
  \begin{subfigure}{0.45\textwidth}
    \centering
    \includegraphics[width=\textwidth]{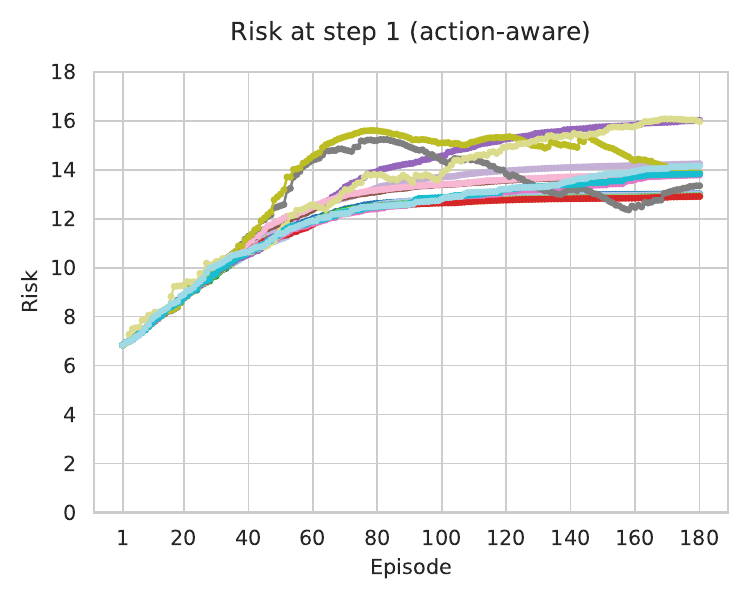}
    \caption{Risk}\label{fig:gridw9-aif-plans-efe-risk}
  \end{subfigure}
    \begin{subfigure}{0.65\textwidth}
    \centering
    \includegraphics[width=\textwidth]{gridw9_aif_policies_legend}
  \end{subfigure}
  \caption{Risk (expected free energy term) for each policy across episodes (showing average of 10 agents).}\label{fig:gridw9-efe-risk}
\end{figure}

\begin{figure}[H]
  \centering
  \begin{subfigure}{0.45\textwidth}
    \centering
    \includegraphics[width=\textwidth]{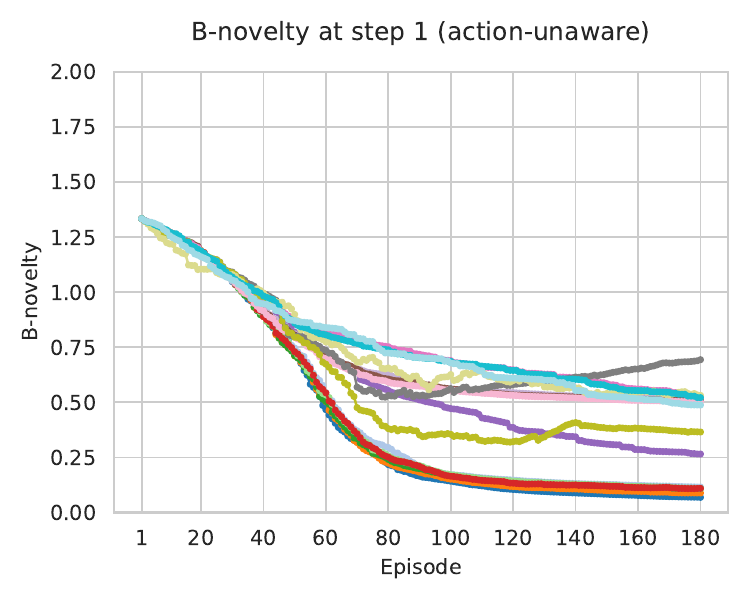}
    \caption{Risk}\label{fig:gridw9-aif-paths-efe-bnov}
  \end{subfigure}
  \begin{subfigure}{0.45\textwidth}
    \centering
    \includegraphics[width=\textwidth]{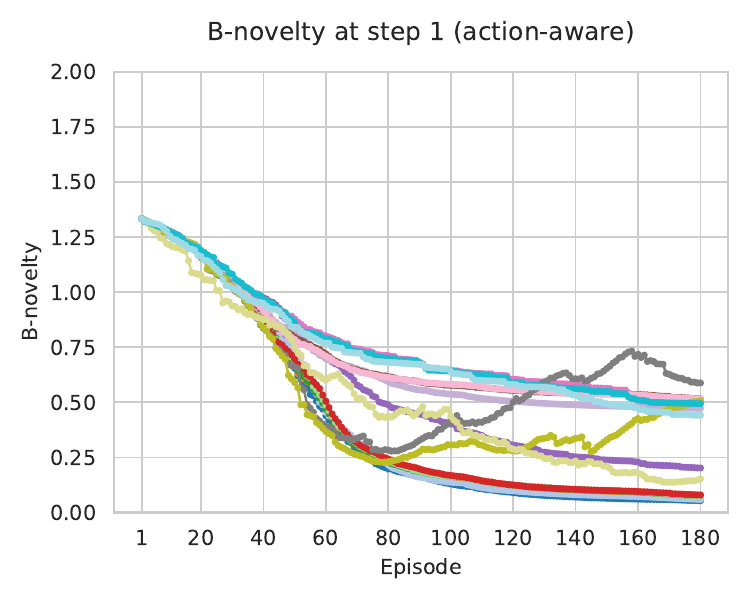}
    \caption{Risk}\label{fig:gridw9-aif-plans-efe-bnov}
  \end{subfigure}
    \begin{subfigure}{0.65\textwidth}
    \centering
    \includegraphics[width=\textwidth]{gridw9_aif_policies_legend}
  \end{subfigure}
  \caption{\(\transmap\)-novelty (expected free energy term) for each policy across episodes (showing average of 10 agents).}\label{fig:gridw9-efe-bnov}
\end{figure}

\paragraph{Planning and learning}
The expected free energies at step 1 in~\cref{fig:gridw9-efe-step0}, again for the same subset of policies considered in~\cref{fig:gridw9-policies-fe-step5}, evolve similarly in both kinds of agent: there is no clear distinction between optimal vs.\ sub-optimal policies, not even at convergence, as a few sub-optimal policies attain expected free energy values comparable to those of the optimal ones (expected free energies at the other steps can be found in~\cref{sssecx:gridw9-efe-steps}).
As seen in the T-maze experiment, risk is much larger than \(\transmap\)-novelty in the composition of expected free energy to the point that the trend of the latter does not differ substantially from that of risk (compare the expected free energy and risk figures,~\cref{fig:gridw9-efe-step0} and~\cref{fig:gridw9-efe-risk}, respectively, and see~\cref{fig:gridw9-efe-bnov} for \(\transmap\)-novelty).
Furthermore, there are again sub-optimal policies for which risk (hence the expected free energy) drops sharply to levels lower than, or comparable to, those of the optimal policies.

\Cref{fig:gridw9-first-pi-probs} shows the policy probabilities across episodes, revealing key differences between the two kinds of agents and between this and the previous experiment.
In action-unaware agents, the probabilities of the six optimal policies share the same upward trend from around episode 60 onwards with their curves almost perfectly overlapping (only the red and blue are visible in the figure, the rest being hidden beneath), and a clear gap emerges between them and those of most sub-optimal policies from around episode 150 (see below for exceptions).
By the end of the experiment, all optimal policies have been recognised and assigned roughly the same probability mass (see left plot in~\cref{fig:gridw9-first-pi-probs}).
In action-aware agents, there is a less perfect overlap between the probabilities of the optimal policies, and the optimal vs.\ suboptimal gap begins somewhat earlier, around episode 120, and is wider by the end of the experiment (again with some exceptions; see right plot in~\cref{fig:gridw9-first-pi-probs} and next).
As in the T-maze experiment, however, some sub-optimal policies also become increasingly probable over time, narrowing the gap with optimal policies in both kinds of agents.
Unlike in the previous experiment, we now find that some of these policies become more probable than optimal ones, even in later episodes, when agents have had ample opportunities to refine the transition model.

\begin{figure}[H]
  \centering
  \begin{subfigure}{0.45\textwidth}\label{fig:gridw9-aif-paths-policies-fe-step1}
    \centering
    \includegraphics[width=\textwidth]{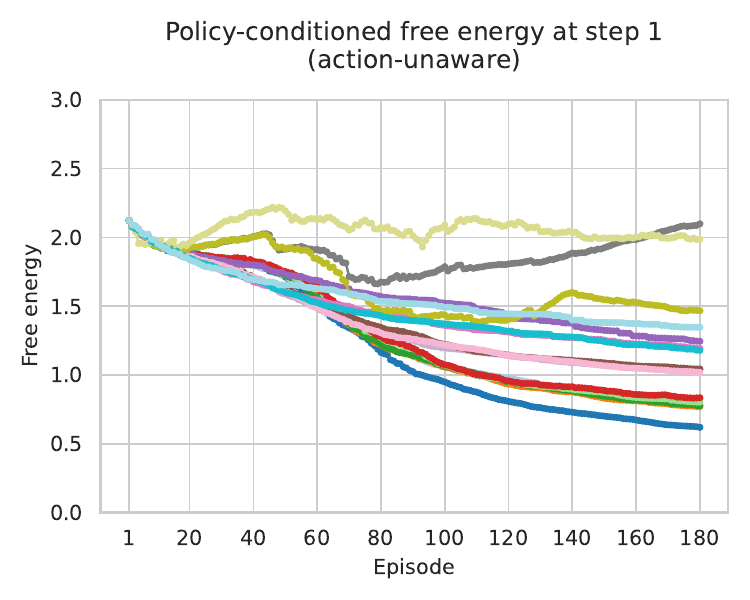}
  \end{subfigure}
  \begin{subfigure}{0.45\textwidth}\label{fig:gridw9-aif-plans-policies-fe-step1}
    \centering
    \includegraphics[width=\textwidth]{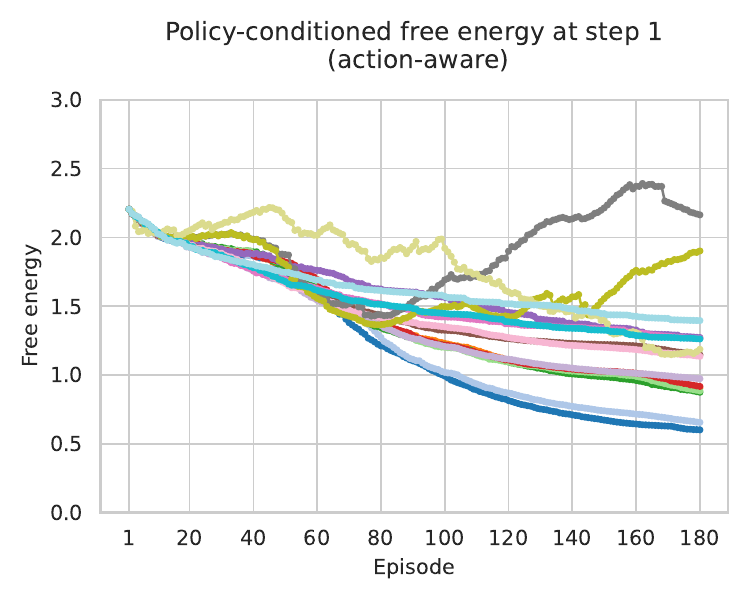}
  \end{subfigure}
  \begin{subfigure}{0.65\textwidth}
    \centering
    \includegraphics[width=\textwidth]{gridw9_aif_policies_legend}
  \end{subfigure}
  \caption{Policy-conditioned free energies at step 1 across episodes (showing average of 10 agents).}\label{fig:gridw9-policies-fes-step1}
\end{figure}

As noted, the evolution of expected free energy is similar in both agents and is not particularly informative about which policies are to be preferred.
Yet, agents can infer a probability distribution over policies that is mostly accurate as it singles out the six optimal policies (despite the significant probability mass acquired by some suboptimal policies).
This can be explained by the relatively low values, achieved by the optimal policies, of the other key quantity used to compute policy probabilities, i.e., the policy-conditioned free energy (at step 1, since we are considering policy probabilities at that step; see~\cref{fig:gridw9-policies-fes-step1}).
When a policy-conditioned free energy is minimised at step 1, it means the agent is more certain about the future consequences of following the corresponding policy for the rest of the episode.
Therefore, more informative policy-conditioned free energies can compensate for less informative expected free energies: given two policies with similar expected free energy, the agent will assign more probability to the one associated with more accurate predictions at the perceptual inference stage (i.e., with the lowest policy-conditioned free energy).

In the case of the sub-optimal policies mentioned above, which at some point surpass the optimal ones in probability, both the expected free energy and the policy-conditioned free energy (at step 1) are informative, but in opposing ways: the expected free energy increases the likelihood of these policies, whereas the policy-conditioned free energy decreases it (compare~\cref{fig:gridw9-efe-step0} and~\cref{fig:gridw9-policies-fes-step1})~\footnote{This is due to the use of the softmax to compute policy probabilities based on the sum of the negative expected free energy and negative policy-conditioned free energy (see~\cref{ssecx:planning-efe})}.
For both kinds of agent, the net effect in this case is that sub-optimal policies gain more probability than optimal ones, indicating that the expected free energy had a greater influence on policy probabilities than the policy-conditioned free energy (see again plots in~\cref{fig:gridw9-first-pi-probs}).
Further investigations into the opposing contributions of these quantities to the policy probabilities, as well as into the reasons why agents' performance does not deteriorate more substantially despite the increased probability of sub-optimal policies, are again left for future work.


\section{Discussion}\label{sec:discussion}

An important difference between reinforcement learning and most active inference works is the particular meaning attributed to the word ‘policy’.
In the former, a policy is often defined as a probability function \(\policy: \statespace \times \actionspace \rightarrow [0, 1]\), and usually written as \(\policy (\action \mid \state)\) to indicate that a policy returns the probability of performing a certain action (at a certain time step) given a state.
In standard active inference algorithms (considered here), a policy is just a sequence of actions indexed by time (recent active inference works have proposed slightly different algorithms in which the notion of policy is much closer to that used in reinforcement learning, see, e.g., \cite{Friston2021b, DaCosta2023a}).


Furthermore, how an active inference agent computes and selects among its policies at each time step, in the more traditional active inference sense of actions indexed by time, is also subject to different interpretations.
On the one hand, a policy can be intended as a motor plan covering a complete trajectory of actions in the past, present and future of an agent's experience.
For instance, a policy could cover a complete trajectory of \(\polhorizon = \ntime - 1\) actions, \(\policy \coloneq [\action_{1},\dots, \action_{\polhorizon}]\), from the beginning to the end of an episode~\cite{DaCosta2020b}.
In this case, at a given time step \(\tau\) (the present), an agent 
assigns a probability to each policy based on how likely it is that its past \(\tau - 1\) actions have generated its past and present observations, and on how likely it is that the policy will lead to the agent's goal in the remaining \(\polhorizon - \tau + 1\) (or \(\ntime - \tau\)) future actions.
On the other hand, a policy can be seen as a motor plan of future actions only.
In this sense, policies correspond to sequences of \(\polhorizon\) actions from the current time step \(t=\tau\) to a future time step \(t=\polhorizon + \tau\), i.e., the planning horizon of each policy such that \(\policy \coloneq [\action_{\tau},\dots, \action_{\polhorizon}]\), see for instance~\parencite{Gottwald2020a,Heins2022a,DaCosta2023a,Friston2021b}~\footnote{Note that in the case of episodes with a fixed number of \(\ntime\) time steps (as those of the experiments described in this work), for action-aware agent we would have that \(\polhorizon \leq \ntime\), if we use \(\polhorizon\) to represent the length of policies intended as sequences of actions from the current time step onwards.}.

In this work, we have taken this difference to characterise agents that are aware and agents that are not aware of the actions that they executed in the past, i.e., action-aware vs.\ action-unaware agents.
Action-aware agents plan to infer the most likely sequences of actions to be followed from the current time step onwards, i.e., their policies contain exclusively future actions.
On the other hand, action-unaware agents plan to infer the most likely sequences of actions that should be \emph{continued} in the future given beliefs about what sequences of actions they performed in the past, 
since they don't have access to explicit past knowledge of their own actions, i.e., their policies include both past and future actions.
Action-aware agents encode variational beliefs that need not be conditional on different past action sequences for past state variables, \(\statevar_{1:\tau}\), because these agents know which ones were executed.
On the other hand, if an agent does not know what its past actions were, then the same beliefs need to be conditioned on the \emph{permissible} sequences of actions that can account for past and present observations properly, i.e., sequence of actions compatible with the agent's experience.

This distinction has some implications for how the agent evaluates future action sequences.
Having access to past actions, an agent can use them to more accurately infer its present location (state) and from there consider different future action sequences, i.e., policies as future plans, based on their most likely and desired consequences.
This lends itself to a separation between free energy minimization of past states/observations and of future ones, potentially relying on two distinct generative models (see \parencite{Parr2019b,Parr2019c,Friston2021b} for examples of this sort of separation, and \parencite{Baltieri2018b} for connection to the separation principle of control theory).
Without access to past actions, an agent's variational beliefs for past, present and future states are conditional on all possible policies.

This discrepancy builds on an established literature in active inference that assumes that agents do not have explicit knowledge of (or access to) the actions they take, either in the past or the present~\parencite{Friston2011b, Adams2013a,Parr2022b}, and constitutes one of its main departures from ``control as inference'' approaches, which instead do~\parencite{Levine2018a}.
Concretely, this means that actions executed by an agent, ground truth actions, are not part of its generative model, but only of the generative process of the environment~\parencite{Friston2009b, Friston2010b, Friston2012b}. 
A generative model contains instead a policy random variable that stands for sequences of actions, whose likelihood needs to be inferred from observations by the agent.
These sequences are not simple copies of ground truth actions~\parencite{Baltieri2018b} but represent all the possible motor paths an agent could have initiated in the past and could be completing in the future.
Agents without such knowledge have to infer the consequences of their own hidden actions (indirectly, as part of the effects a policy has) and the environment's hidden states at the same time, from the same given observations, and this puts a heavier burden on their ability to plan for the future, since they are effectively operating without the classical efference copy mechanism~\parencite{Crapse2008a}.
This is however compatible with variations of the ``equilibrium point hypothesis'' and ``referent control''~\parencite{Feldman2010a,Feldman2009a,Feldman2016a}, which contrast proposals of forward and inverse models based on linear quadratic Gaussian control and the separation principle from control theory, see~\parencite{Baltieri2019b,Baltieri2019c,Baltieri2020c,Baltieri2019f} for a more comprehensive perspective, and constitute the basis for continuous-time formulations of active inference minimising variational free energy 
~\parencite{Friston2010b, Friston2012b, Parr2022b}.
While seemingly disadvantageous when considering the same active inference architecture (knowing one's actions would clearly help), it is often claimed that this constitutes an overall improvement over the standard use of inverse models, see~\parencite{Friston2011b}, replacing complex (forward) model inversions with proprioceptive predictions in a low dimensional latent space and pre-programmed reflexes translating those predictions into actions, which in turn ought to provide a more biologically plausible account of motor control in humans among others~\parencite{Friston2011b, Adams2013a}.

In this light, action-aware agents, following~\parencite{Heins2022a}, deviate from the classic active inference literature just illustrated because, at each time step, they have access to a copy of the ground truth past actions.
Despite the fact that this occurs within a Bayesian framework that no longer distinguishes between forward and inverse models (\parencite[see][Ch. 4]{Clark2016a}), it is closer to a reinstatement of the notion of efference copy mechanism.
One could object that in this active inference framework policies and actions still correspond to proprioceptive predictions, and therefore they should not be confused with the standard notion of efference copy.
However, action-aware agents are required to store copies of ground truth proprioceptive predictions, an operation that is not part of the standard active inference formulation (to the best of our knowledge) and that, again, brings the notion of proprioceptive prediction closer to that of efference copy.
In contrast, action-unaware agents, as formulated in~\parencite{DaCosta2020b}, can be seen as the discrete-time counterpart, operating at slower time scales and at a higher abstraction level, to standard continuous-time active inference, usually focused on low-level motion generation skills~\parencite{Parr2018a}, matching its architecture inspired by referent control, where action/motor commands as proprioceptive predictions are inferred and conditioned upon sensory observations.


Our work provides a Python implementation that relaxes the strong assumption of action-aware agents in~\cite{Heins2022a}, more closely follows standard formulations of active inference and its proposal of more biological plausible models without the traditional mechanism of efference copy, and shows evidence from simulations that action-unaware agents can match the performances of their action-aware counterparts which have explicit knowledge of their own actions.

While action-unaware agents constitute, according to active inference, a more biologically plausible implementation of active inference agents, this comes at a cost.
As showcased by algorithms~\cref{pscode:aif-action-unaware} and~\cref{pscode:aif-action-aware}, the alternative ways of viewing policies that characterise action-aware and action-unaware agents have important consequences on the corresponding implementations.
In particular, they affect the computations that go into perceptual inference and, in turn, its time complexity.
At each step, action-aware agents need to update only one collection of variational distributions over past state variables: those conditioned on the sequence of actions that was executed (this explains why in these agents there is only a single variational free energy for past states, as remarked earlier and as we saw in~\cref{fig:tmaze4-policies-fe-step4,fig:gridw9-policies-fe-step5}).
In contrast, at each time step, action-unaware agents have to update as many collections of variational distribution as the number of policies, to compute the policy-conditioned free energies that quantify the extent to which a policy is consistent with the collected observations.
In other words, the time complexity of the perceptual stage is \(O(n)\) in action-aware agent and \(O(nm)\) in action-unaware, where \(n\) is the number of past state variables and \(m\) is the number of policies.
This makes the algorithm for action-aware agents clearly more efficient than the latter by assuming that the agent has access to more information.

This may suggest that action-unaware agents are more tailored for finite-horizon tasks in which episodes have a \emph{fixed} duration, corresponding to the number of actions in each policy plus 1, i.e., \(\ntime = \polhorizon + 1\) time steps, because there is no action at the last time step.
In this learning setting, action-unaware agents keeps track of how many time steps have passed from the beginning of an episode to evaluate policies based on the remaining actions only.
Conversely, action-aware agents appear to be more congenial to finite-horizon tasks in which episodes have an \emph{indefinite} duration because of the separation between past and future sequences of actions which predisposes them to consider at each time step a certain fixed number of \(\polhorizon\) actions into the future.

\section{Concluding Remarks}\label{sec:ccnclusion}
Active inference has gained traction in computational neuroscience as a modelling framework to study adaptive decision making in a variety of context.
In this work, we introduced the essential aspects of the framework in detail, showing how free energy minimisation can be used as a guiding principle to understand perception, planning, action-selection, and learning in an adaptive agent moving in a simple grid-world environment.


We investigated active inference in two different regimes, studying the typical behaviour of agents that are not aware of their past actions (action-unaware) and of agents that are (action-aware).
The former follows more strictly the tradition of active inference frameworks inspired by the ``equilibrium point hypothesis'' and ``referent control''~\parencite{Feldman2010a,Feldman2009a}, claiming that humans, among other biological agents, do not possess or even need the ability to discount the effects of their actions from their observations~\parencite{Bridgeman2007a,Feldman2016a, Latash2021a}.
The latter assumes that knowledge of past action sequences is available to an agent, which can thus simply discount the effects of known executed actions from its recollection of past observations and from current ones so to more easily plan for the future.

Our simulations in two toy environments, a T-maze and a 3x3 grid world, showed that, while in principle at a severe disadvantage, action-unaware agents can overall match the performances of action-aware ones.
While impressive, this comes at a heavy computational cost, which currently prevents action-unaware agents from being fully scalable to larger simulations, since there is a combinatorial explosion of possible action sequences to be checked that depends not only on present and future time steps and their associated actions, but also on past ones.
At this stage, we speculate that mechanisms such as weight-based sampling of action sequences may provide an affordable implementation in high-dimensional action-sequence spaces, but we leave this and other investigations to future work.

\phantomsection
\addcontentsline{toc}{section}{Acknowledgments} 
\section*{Acknowledgments}
F.T., R.K. and M.B. were supported by JST, Moonshot R\&D, Grant Number JPMJMS2012. F.T. and K.S. were partially supported by Tokyo Electron. K.S. was supported by JST, CREST Grant Number JPMJCR21P4, Japan.

\printbibliography%

@string{ICML = "International Conference on Machine Learning"}

@article{Adams2013a,
  title = {Predictions Not Commands: Active Inference in the Motor System},
  author = {Adams, Rick A. and Shipp, Stewart and Friston, Karl J.},
  date = {2013},
  journaltitle = {Brain Structure and Function},
  shortjournal = {Brain Struct. Funct.},
  volume = {218},
  number = {3},
  pages = {611--643},
  issn = {18632653},
  doi = {10.1007/s00429-012-0475-5},
  langid = {english}
}

@inproceedings{Baltieri2018b,
  title = {The Modularity of Action and Perception Revisited Using Control Theory and Active Inference},
  booktitle = {Artificial {{Life Conference Proceedings}} 30},
  author = {Baltieri, Manuel and Buckley, Christopher L.},
  date = {2018},
  pages = {121--128},
  publisher = {MIT Press One Rogers Street, Cambridge, MA 02142-1209, USA journals-info …},
  copyright = {All rights reserved}
}

@thesis{Baltieri2019b,
  type = {phdthesis},
  title = {Active Inference: {{Building}} a New Bridge between Control Theory and Embodied Cognitive Science.},
  shorttitle = {Active Inference},
  author = {Baltieri, Manuel},
  date = {2019},
  institution = {University of Sussex},
  copyright = {All rights reserved}
}

@inproceedings{Baltieri2019c,
  title = {Active Inference: {{Computational}} Models of Motor Control without Efference Copy},
  shorttitle = {Active Inference},
  booktitle = {Proceedings of the 2019 {{Conference}} on {{Cognitive Computational Neuroscience}}},
  author = {Baltieri, Manuel and Buckley, Christopher L.},
  date = {2019},
  publisher = {ACM, New York},
  copyright = {All rights reserved}
}

@inproceedings{Baltieri2019e,
  title = {Nonmodular Architectures of Cognitive Systems Based on Active Inference},
  booktitle = {2019 {{International Joint Conference}} on {{Neural Networks}} ({{IJCNN}})},
  author = {Baltieri, Manuel and Buckley, Christopher L.},
  date = {2019},
  pages = {1--8},
  publisher = {IEEE},
  copyright = {All rights reserved}
}

@article{Baltieri2019f,
  title = {{{PID}} Control as a Process of Active Inference with Linear Generative Models},
  author = {Baltieri, Manuel and Buckley, Christopher L.},
  date = {2019},
  journaltitle = {Entropy. An International and Interdisciplinary Journal of Entropy and Information Studies},
  shortjournal = {Entropy},
  volume = {21},
  number = {3},
  pages = {257},
  publisher = {MDPI},
  copyright = {All rights reserved}
}

@inproceedings{Baltieri2020c,
  title = {A {{Bayesian}} Perspective on Classical Control},
  booktitle = {2020 {{International Joint Conference}} on {{Neural Networks}} ({{IJCNN}})},
  author = {Baltieri, Manuel},
  date = {2020},
  pages = {1--8},
  publisher = {IEEE},
  copyright = {All rights reserved}
}

@article{Bridgeman2007a,
  title = {Efference Copy and Its Limitations},
  author = {Bridgeman, Bruce},
  date = {2007-07-01},
  journaltitle = {Computers in Biology and Medicine},
  shortjournal = {Computers in Biology and Medicine},
  series = {Vision and {{Movement}} in {{Man}} and {{Machines}}},
  volume = {37},
  number = {7},
  pages = {924--929},
  issn = {0010-4825},
  doi = {10.1016/j.compbiomed.2006.07.001}
}

@incollection{Bruineberg2017a,
  title = {Active Inference and the Primacy of the ‘{{I}} Can’},
  booktitle = {Philosophy and {{Predictive Processing}}},
  author = {Bruineberg, Jelle},
  editor = {Metzinger, Thomas and Wiese, Wanja},
  date = {2017},
  pages = {1--18},
  publisher = {MIND Group},
  location = {Frankfurt am Main, Germany},
  url = {https://predictive-mind.net/DOI?isbn=9783958573062},
  isbn = {978-3-95857-306-2},
  langid = {english}
}

@article{Buckley2017a,
  title = {The Free Energy Principle for Action and Perception: A Mathematical Review},
  author = {Buckley, Christopher L. and Kim, Chang Sub and McGregor, Simon and Seth, Anil K.},
  date = {2017},
  journaltitle = {Journal of Mathematical Psychology},
  shortjournal = {J. Math. Psychol.},
  volume = {81},
  pages = {55--79},
  publisher = {Elsevier Inc.},
  issn = {10960880},
  doi = {10.1016/j.jmp.2017.09.004},
  langid = {english}
}

@article{Catal2020a,
  title = {Learning Generative State Space Models for Active Inference},
  author = {Çatal, Ozan and Wauthier, Samuel and De Boom, Cedric and Verbelen, Tim and Dhoedt, Bart},
  date = {2020},
  journaltitle = {Frontiers in Computational Neuroscience},
  shortjournal = {Frontiers in Computational Neuroscience},
  volume = {14},
  pages = {103},
  issn = {1662-5188},
  doi = {10.3389/fncom.2020.574372},
  langid = {english}
}

@article{Clark2013b,
  title = {Whatever next? {{Predictive}} Brains, Situated Agents, and the Future of Cognitive Science},
  author = {Clark, Andy},
  date = {2013},
  journaltitle = {Behavioral and Brain Sciences},
  shortjournal = {Behav. Brain Sci.},
  volume = {36},
  number = {3},
  eprint = {23663408},
  eprinttype = {pubmed},
  pages = {181--204},
  issn = {14691825},
  doi = {10.1017/S0140525X12000477},
  langid = {english}
}

@article{Clark2015a,
  title = {Radical Predictive Processing},
  author = {Clark, Andy},
  date = {2015},
  journaltitle = {The Southern Journal of Philosophy},
  shortjournal = {South. J. Philos.},
  volume = {53},
  number = {S1},
  pages = {3--27},
  issn = {00384283},
  doi = {10.1111/sjp.12120},
  langid = {english}
}

@book{Clark2016a,
  title = {Surfing Uncertainty: Prediction, Action, and the Embodied Mind},
  author = {Clark, Andy},
  date = {2016},
  publisher = {Oxford University Press},
  location = {Oxford, UK},
  isbn = {978-0-19-021703-7},
  langid = {english}
}

@article{Crapse2008a,
  title = {Corollary Discharge across the Animal Kingdom},
  author = {Crapse, Trinity B. and Sommer, Marc A.},
  date = {2008-08},
  journaltitle = {Nature Reviews Neuroscience},
  shortjournal = {Nat Rev Neurosci},
  volume = {9},
  number = {8},
  pages = {587--600},
  publisher = {Nature Publishing Group},
  issn = {1471-0048},
  doi = {10.1038/nrn2457},
  langid = {english}
}

@article{DaCosta2020b,
  title = {Active Inference on Discrete State-Spaces: {{A}} Synthesis},
  shorttitle = {Active Inference on Discrete State-Spaces},
  author = {Da Costa, Lancelot and Parr, Thomas and Sajid, Noor and Veselic, Sebastijan and Neacsu, Victorita and Friston, Karl},
  date = {2020-12-01},
  journaltitle = {Journal of Mathematical Psychology},
  shortjournal = {Journal of Mathematical Psychology},
  volume = {99},
  pages = {102447},
  issn = {0022-2496},
  doi = {10.1016/j.jmp.2020.102447}
}

@article{DaCosta2023a,
  title = {Reward {{Maximization Through Discrete Active Inference}}},
  author = {Da Costa, Lancelot and Sajid, Noor and Parr, Thomas and Friston, Karl and Smith, Ryan},
  date = {2023-04-18},
  journaltitle = {Neural Computation},
  shortjournal = {Neural Computation},
  volume = {35},
  number = {5},
  pages = {807--852},
  issn = {0899-7667},
  doi = {10.1162/neco_a_01574}
}

@book{Deisenroth2020a,
  title = {Mathematics for Machine Learning},
  author = {Deisenroth, Marc Peter and Faisal, A. Aldo and Ong, Cheng Soon},
  date = {2020-04},
  publisher = {Cambridge University Press},
  isbn = {978-1-108-45514-5}
}

@article{Feldman2009a,
  title = {New Insights into Action–Perception Coupling},
  author = {Feldman, Anatol G.},
  date = {2009-03-01},
  journaltitle = {Experimental Brain Research},
  shortjournal = {Exp Brain Res},
  volume = {194},
  number = {1},
  pages = {39--58},
  issn = {1432-1106},
  doi = {10.1007/s00221-008-1667-3},
  langid = {english}
}

@article{Feldman2010a,
  title = {Attention, Uncertainty, and Free-Energy},
  author = {Feldman, Harriet and Friston, Karl J.},
  date = {2010},
  journaltitle = {Frontiers in Human Neuroscience},
  shortjournal = {Front. Hum. Neurosci.},
  volume = {4},
  pages = {1--23},
  issn = {16625161},
  doi = {10.3389/fnhum.2010.00215},
  issue = {December}
}

@article{Feldman2016a,
  title = {Active Sensing without Efference Copy: Referent Control of Perception},
  shorttitle = {Active Sensing without Efference Copy},
  author = {Feldman, Anatol G.},
  date = {2016-09},
  journaltitle = {Journal of Neurophysiology},
  shortjournal = {J. Neurophysiol.},
  volume = {116},
  number = {3},
  pages = {960--976},
  publisher = {American Physiological Society},
  issn = {0022-3077},
  doi = {10.1152/jn.00016.2016}
}

@article{Friston2005a,
  title = {A Theory of Cortical Responses},
  author = {Friston, Karl},
  date = {2005},
  journaltitle = {Philosophical Transactions of the Royal Society B: Biological Sciences},
  shortjournal = {Philos. Trans. R. Soc. B Biol. Sci.},
  volume = {360},
  number = {1456},
  pages = {815--836},
  issn = {0962-8436},
  doi = {10.1098/rstb.2005.1622},
  langid = {english}
}

@article{Friston2008a,
  title = {Hierarchical Models in the Brain},
  author = {Friston, Karl},
  date = {2008},
  journaltitle = {PLoS Computational Biology},
  shortjournal = {PLoS Comput. Biol.},
  volume = {4},
  number = {11},
  eprint = {18989391},
  eprinttype = {pubmed},
  pages = {1--24},
  issn = {1553734X},
  doi = {10.1371/journal.pcbi.1000211},
  langid = {english}
}

@article{Friston2009b,
  title = {Reinforcement Learning or Active Inference?},
  author = {Friston, Karl J. and Daunizeau, Jean and Kiebel, Stefan J.},
  date = {2009},
  journaltitle = {PLoS ONE},
  volume = {4},
  number = {7},
  eprint = {19641614},
  eprinttype = {pubmed},
  issn = {19326203},
  doi = {10.1371/journal.pone.0006421}
}

@article{Friston2009c,
  title = {The Free-Energy Principle: A Rough Guide to the Brain?},
  author = {Friston, Karl},
  date = {2009},
  journaltitle = {Trends in Cognitive Sciences},
  shortjournal = {Trends Cogn. Sci.},
  volume = {13},
  number = {7},
  eprint = {19559644},
  eprinttype = {pubmed},
  pages = {293--301},
  issn = {13646613},
  doi = {10.1016/j.tics.2009.04.005},
  langid = {english}
}

@article{Friston2010b,
  title = {Action and Behavior: A Free-Energy Formulation},
  author = {Friston, Karl J. and Daunizeau, Jean and Kilner, James and Kiebel, Stefan J.},
  date = {2010},
  journaltitle = {Biological Cybernetics},
  shortjournal = {Biol. Cybern.},
  volume = {102},
  number = {3},
  eprint = {20148260},
  eprinttype = {pubmed},
  pages = {227--260},
  issn = {03401200},
  doi = {10.1007/s00422-010-0364-z},
  langid = {english}
}

@article{Friston2011b,
  title = {What {{Is Optimal}} about {{Motor Control}}?},
  author = {Friston, Karl},
  date = {2011-11-03},
  journaltitle = {Neuron},
  shortjournal = {Neuron},
  volume = {72},
  number = {3},
  pages = {488--498},
  issn = {0896-6273},
  doi = {10.1016/j.neuron.2011.10.018}
}

@article{Friston2012b,
  title = {Active Inference and Agency: Optimal Control without Cost Functions},
  author = {Friston, Karl and Samothrakis, Spyridon and Montague, Read},
  date = {2012},
  journaltitle = {Biological Cybernetics},
  shortjournal = {Biol. Cybern.},
  volume = {106},
  number = {8--9},
  eprint = {22864468},
  eprinttype = {pubmed},
  pages = {523--541},
  issn = {03401200},
  doi = {10.1007/s00422-012-0512-8},
  langid = {english}
}

@article{Friston2012g,
  title = {Prediction, Perception and Agency},
  author = {Friston, Karl},
  date = {2012-02},
  journaltitle = {International Journal of Psychophysiology},
  shortjournal = {Int. J. Psychophysiol.},
  volume = {83},
  number = {2},
  eprint = {22178504},
  eprinttype = {pubmed},
  pages = {248--252},
  issn = {01678760},
  doi = {10.1016/j.ijpsycho.2011.11.014}
}

@article{Friston2016a,
  title = {Active Inference and Learning},
  author = {Friston, Karl and FitzGerald, Thomas and Rigoli, Francesco and Schwartenbeck, Philipp and O'Doherty, John and Pezzulo, Giovanni},
  date = {2016},
  journaltitle = {Neuroscience and Biobehavioral Reviews},
  shortjournal = {Neurosci. Biobehav. Rev.},
  volume = {68},
  eprint = {27375276},
  eprinttype = {pubmed},
  pages = {862--879},
  publisher = {Elsevier Ltd},
  issn = {18737528},
  doi = {10.1016/j.neubiorev.2016.06.022},
  langid = {english}
}

@article{Friston2017a,
  title = {The Graphical Brain: Belief Propagation and Active Inference},
  author = {Friston, Karl J. and Parr, Thomas and family=Vries, given=Bert, prefix=de, useprefix=true},
  date = {2017},
  journaltitle = {Network Neuroscience},
  shortjournal = {Netw. Neurosci.},
  volume = {1},
  number = {4},
  pages = {381--414},
  issn = {2472-1751},
  doi = {10.1162/NETN_a_00018},
  langid = {english}
}

@article{Friston2017c,
  title = {Active Inference: A Process Theory},
  author = {Friston, Karl and FitzGerald, Thomas and Rigoli, Francesco and Schwartenbeck, Philipp and Pezzulo, Giovanni},
  date = {2017},
  journaltitle = {Neural Computation},
  shortjournal = {Neural Comput.},
  volume = {29},
  number = {1},
  pages = {1--49},
  doi = {10.1162/NECO_a_00912},
  langid = {english}
}

@article{Friston2018a,
  title = {Deep Temporal Models and Active Inference},
  author = {Friston, Karl J. and Rosch, Richard and Parr, Thomas and Price, Cathy and Bowman, Howard},
  date = {2018},
  journaltitle = {Neuroscience \& Biobehavioral Reviews},
  shortjournal = {Neurosci. Biobehav. Rev.},
  volume = {90},
  pages = {486--501},
  issn = {0149-7634},
  doi = {10.1016/j.neubiorev.2018.04.004},
  langid = {english}
}

@article{Friston2019b,
  title = {A Free Energy Principle for a Particular Physics},
  author = {Friston, Karl},
  date = {2019},
  pages = {1--140},
  url = {http://arxiv.org/abs/1906.10184},
  langid = {english}
}

@article{Friston2021b,
  title = {Sophisticated Inference},
  author = {Friston, Karl and Da Costa, Lancelot and Hafner, Danijar and Hesp, Casper and Parr, Thomas},
  date = {2021-02-24},
  journaltitle = {Neural Computation},
  shortjournal = {Neural Comput.},
  volume = {33},
  number = {3},
  pages = {713--763},
  publisher = {MIT Press},
  issn = {0899-7667},
  doi = {10.1162/neco_a_01351}
}

@article{Gottwald2020a,
  title = {The Two Kinds of Free Energy and the {{Bayesian}} Revolution},
  author = {Gottwald, Sebastian and Braun, Daniel A.},
  date = {2020},
  journaltitle = {PLOS Computational Biology},
  shortjournal = {PLOS Comput. Biol.},
  volume = {16},
  number = {12},
  pages = {1--32},
  publisher = {Public Library of Science},
  doi = {10.1371/journal.pcbi.1008420},
  langid = {english}
}

@article{Heins2020a,
  title = {Deep Active Inference and Scene Construction},
  author = {Heins, R. Conor and Mirza, M. Berk and Parr, Thomas and Friston, Karl and Kagan, Igor and Pooresmaeili, Arezoo},
  date = {2020},
  journaltitle = {Frontiers in Artificial Intelligence},
  shortjournal = {Front. Artif. Intell.},
  volume = {3},
  pages = {1--23},
  issn = {2624-8212},
  doi = {10.3389/frai.2020.509354},
  langid = {english}
}

@article{Heins2022a,
  title = {Pymdp: {{A Python}} Library for Active Inference in Discrete State Spaces},
  author = {Heins, Conor and Millidge, Beren and Demekas, Daphne and Klein, Brennan and Friston, Karl and Couzin, Iain D. and Tschantz, Alexander},
  date = {2022-05},
  journaltitle = {Journal of Open Source Software},
  shortjournal = {J. Open Source Softw.},
  volume = {7},
  number = {73},
  pages = {4098},
  publisher = {The Open Journal},
  issn = {2475-9066},
  doi = {10.21105/joss.04098}
}

@book{Hohwy2013a,
  title = {The Predictive Mind},
  author = {Hohwy, Jakob},
  date = {2013},
  publisher = {Oxford University Press},
  location = {Oxford, UK},
  isbn = {978-0-19-968673-5},
  langid = {english}
}

@article{Hohwy2020a,
  title = {New Directions in Predictive Processing},
  author = {Hohwy, Jakob},
  date = {2020},
  journaltitle = {Mind and Language},
  shortjournal = {Mind Lang.},
  volume = {35},
  number = {2},
  pages = {209--223},
  issn = {14680017},
  doi = {10.1111/mila.12281},
  langid = {english}
}

@book{Jazwinski1970a,
  title = {Stochastic Processes and Filtering Theory},
  author = {Jazwinski, Andrew H.},
  date = {1970},
  publisher = {Academic Press},
  location = {New York, USA},
  isbn = {978-0-12-381550-7}
}

@article{Kaplan2018a,
  title = {Planning and Navigation as Active Inference},
  author = {Kaplan, Raphael and Friston, Karl J.},
  date = {2018},
  journaltitle = {Biological Cybernetics},
  shortjournal = {Biol. Cybern.},
  volume = {112},
  number = {4},
  pages = {323--343},
  publisher = {Springer Berlin Heidelberg},
  issn = {14320770},
  doi = {10.1007/s00422-018-0753-2}
}

@article{Kappen2012a,
  title = {Optimal Control as a Graphical Model Inference Problem},
  author = {Kappen, Hilbert J. and Gómez, Vicenç and Opper, Manfred},
  date = {2012-05-01},
  journaltitle = {Machine Learning},
  shortjournal = {Machine Learning},
  volume = {87},
  number = {2},
  pages = {159--182},
  issn = {1573-0565},
  doi = {10.1007/s10994-012-5278-7}
}

@article{Lanillos2021a,
  title = {Active Inference in Robotics and Artificial Agents: {{Survey}} and Challenges},
  author = {Lanillos, Pablo and Meo, Cristian and Pezzato, Corrado and Meera, Ajith Anil and Baioumy, Mohamed and Ohata, Wataru and Tschantz, Alexander and Millidge, Beren and Wisse, Martijn and Buckley, Christopher L. and Tani, Jun},
  date = {2021},
  journaltitle = {ArXiv211201871v1 CsRO},
  eprint = {2112.01871v1},
  eprinttype = {arXiv},
  eprintclass = {cs.RO},
  pages = {1--20},
  url = {https://arxiv.org/abs/2112.01871},
  langid = {english}
}

@article{Latash2021a,
  title = {Efference Copy in Kinesthetic Perception: A Copy of What Is It?},
  shorttitle = {Efference Copy in Kinesthetic Perception},
  author = {Latash, Mark L.},
  date = {2021-04},
  journaltitle = {Journal of Neurophysiology},
  shortjournal = {J. Neurophysiol.},
  volume = {125},
  number = {4},
  pages = {1079--1094},
  publisher = {American Physiological Society},
  issn = {0022-3077},
  doi = {10.1152/jn.00545.2020}
}

@article{Lee2003a,
  title = {Hierarchical {{Bayesian}} Inference in the Visual Cortex},
  author = {Lee, Tai Sing and Mumford, David},
  date = {2003},
  journaltitle = {Journal of the Optical Society of America A},
  shortjournal = {J. Opt. Soc. Am. A},
  volume = {20},
  number = {7},
  eprint = {12868647},
  eprinttype = {pubmed},
  pages = {1434--1448},
  issn = {1084-7529},
  doi = {10.1364/josaa.20.001434},
  langid = {english}
}

@article{Levine2018a,
  title = {Reinforcement Learning and Control as Probabilistic Inference: Tutorial and Review},
  author = {Levine, Sergey},
  date = {2018},
  journaltitle = {ArXiv180500909v3 CsLG},
  eprint = {1805.00909v3},
  eprinttype = {arXiv},
  eprintclass = {cs.LG},
  pages = {1--22},
  url = {https://arxiv.org/abs/1805.00909},
  langid = {english}
}

@article{Mazzaglia2022a,
  title = {The {{Free Energy Principle}} for {{Perception}} and {{Action}}: {{A Deep Learning Perspective}}},
  author = {Mazzaglia, Pietro and Verbelen, Tim and Çatal, Ozan and Dhoedt, Bart},
  date = {2022},
  journaltitle = {Entropy},
  volume = {24},
  number = {2},
  pages = {1--22},
  issn = {1099-4300},
  doi = {10.3390/e24020301},
  langid = {english}
}

@online{Millidge2020d,
  title = {Whence the Expected Free Energy?},
  author = {Millidge, Beren and Tschantz, Alexander and Buckley, Christopher L},
  date = {2020},
  number = {2004.08128},
  eprint = {2004.08128},
  eprinttype = {arXiv},
  eprintclass = {cs.AI},
  doi = {10.48550/arXiv.2004.08128},
  langid = {english},
  pubstate = {prepublished}
}

@online{Millidge2021a,
  title = {Predictive Coding: A Theoretical and Experimental Review},
  author = {Millidge, Beren and Seth, Anil and Buckley, Christopher L},
  date = {2021},
  eprint = {2107.12979v4},
  eprinttype = {arXiv},
  eprintclass = {cs.AI},
  doi = {10.48550/arXiv.2107.12979},
  langid = {english},
  pubstate = {prepublished}
}

@article{Millidge2021c,
  title = {Whence the {{Expected Free Energy}}?},
  author = {Millidge, Beren and Tschantz, Alexander and Buckley, Christopher L.},
  date = {2021-02-01},
  journaltitle = {Neural Computation},
  shortjournal = {Neural Computation},
  volume = {33},
  number = {2},
  pages = {447--482},
  issn = {0899-7667},
  doi = {10.1162/neco_a_01354}
}

@article{Mirza2016a,
  title = {Scene Construction, Visual Foraging, and Active Inference},
  author = {Mirza, M. Berk and Adams, Rick A. and Mathys, Christoph D. and Friston, Karl J.},
  date = {2016},
  journaltitle = {Frontiers in Computational Neuroscience},
  shortjournal = {Front. Comput. Neurosci.},
  volume = {10},
  pages = {1--16},
  issn = {16625188},
  doi = {10.3389/fncom.2016.00056},
  langid = {english}
}

@article{Mirza2019a,
  title = {Introducing a {{Bayesian}} Model of Selective Attention Based on Active Inference},
  author = {Mirza, M. Berk and Adams, Rick A. and Friston, Karl and Parr, Thomas},
  date = {2019},
  journaltitle = {Scientific Reports},
  shortjournal = {Sci. Rep.},
  volume = {9},
  number = {1},
  eprint = {31558746},
  eprinttype = {pubmed},
  pages = {1--22},
  issn = {20452322},
  doi = {10.1038/s41598-019-50138-8},
  langid = {english}
}

@book{Murphy2023a,
  title = {Probabilistic Machine Learning: Advanced Topics},
  author = {Murphy, Kevin P.},
  date = {2023},
  publisher = {The MIT Press},
  location = {Cambridge, Massachusetts},
  url = {https://probml.github.io/pml-book/book2.html},
  langid = {english}
}

@article{Nehrer2025a,
  title = {Introducing {{ActiveInference}}.Jl: {{A Julia Library}} for {{Simulation}} and {{Parameter Estimation}} with {{Active Inference Models}}},
  shorttitle = {Introducing {{ActiveInference}}.Jl},
  author = {Nehrer, Samuel William and Ehrenreich Laursen, Jonathan and Heins, Conor and Friston, Karl and Mathys, Christoph and Thestrup Waade, Peter},
  date = {2025-01},
  journaltitle = {Entropy},
  volume = {27},
  number = {1},
  pages = {62},
  publisher = {Multidisciplinary Digital Publishing Institute},
  issn = {1099-4300},
  doi = {10.3390/e27010062},
  issue = {1},
  langid = {english}
}

@article{Parr2017a,
  title = {Uncertainty, Epistemics and Active Inference},
  author = {Parr, Thomas and Friston, Karl J.},
  date = {2017-11},
  journaltitle = {Journal of The Royal Society Interface},
  shortjournal = {J. R. Soc. Interface},
  volume = {14},
  number = {136},
  eprint = {15944135},
  eprinttype = {pubmed},
  pages = {20170376},
  issn = {1742-5689},
  doi = {10.1098/rsif.2017.0376}
}

@article{Parr2018a,
  title = {The Discrete and Continuous Brain: From Decisions to Movement---and Back Again},
  author = {Parr, Thomas and Friston, Karl J. and Shankar, Tanmay},
  date = {2018},
  journaltitle = {Neural Computation},
  shortjournal = {Neural Comput.},
  volume = {30},
  number = {9},
  pages = {2319--2347},
  doi = {10.1162/neco_a_01102},
  langid = {english}
}

@article{Parr2019b,
  title = {Neuronal Message Passing Using Mean-Field, {{Bethe}}, and Marginal Approximations},
  author = {Parr, Thomas and Markovic, Dimitrije and Kiebel, Stefan J. and Friston, Karl J.},
  date = {2019},
  journaltitle = {Scientific Reports},
  shortjournal = {Sci. Rep.},
  volume = {9},
  number = {1},
  pages = {1--18},
  publisher = {Springer US},
  issn = {20452322},
  doi = {10.1038/s41598-018-38246-3},
  isbn = {4159801838}
}

@article{Parr2019c,
  title = {Generalised Free Energy and Active Inference},
  author = {Parr, Thomas and Friston, Karl J.},
  date = {2019},
  journaltitle = {Biological Cybernetics},
  shortjournal = {Biol. Cybern.},
  volume = {113},
  number = {5--6},
  eprint = {31562544},
  eprinttype = {pubmed},
  pages = {495--513},
  publisher = {Springer Berlin Heidelberg},
  issn = {14320770},
  doi = {10.1007/s00422-019-00805-w},
  isbn = {0123456789}
}

@article{Parr2020a,
  title = {Prefrontal Computation as Active Inference},
  author = {Parr, Thomas and Rikhye, Rajeev Vijay and Halassa, Michael M and Friston, Karl J},
  date = {2020},
  journaltitle = {Cerebral Cortex},
  shortjournal = {Cereb. Cortex},
  volume = {30},
  number = {2},
  pages = {682--695},
  issn = {1047-3211},
  doi = {10.1093/cercor/bhz118},
  langid = {english}
}

@book{Parr2022b,
  title = {Active {{Inference}}: {{The Free Energy Principle}} in {{Mind}}, {{Brain}}, and {{Behavior}}},
  author = {Parr, Thomas and Pezzulo, Giovanni and Friston, Karl J.},
  date = {2022},
  publisher = {The MIT Press},
  url = {https://direct.mit.edu/books/oa-monograph/5299/Active-InferenceThe-Free-Energy-Principle-in-Mind},
  isbn = {978-0-262-04535-3},
  langid = {english}
}

@article{Pezzulo2015a,
  title = {Active Inference, Homeostatic Regulation and Adaptive Behavioural Control},
  author = {Pezzulo, Giovanni and Rigoli, Francesco and Friston, Karl},
  date = {2015},
  journaltitle = {Progress in Neurobiology},
  shortjournal = {Prog. Neurobiol.},
  volume = {134},
  eprint = {26365173},
  eprinttype = {pubmed},
  pages = {17--35},
  publisher = {Elsevier Ltd},
  issn = {18735118},
  doi = {10.1016/j.pneurobio.2015.09.001},
  langid = {english}
}

@article{Pezzulo2018a,
  title = {Hierarchical Active Inference: A Theory of Motivated Control},
  author = {Pezzulo, Giovanni and Rigoli, Francesco and Friston, Karl J.},
  date = {2018},
  journaltitle = {Trends in Cognitive Sciences},
  shortjournal = {Trends Cogn. Sci.},
  volume = {22},
  number = {4},
  eprint = {29475638},
  eprinttype = {pubmed},
  pages = {294--306},
  publisher = {Elsevier Ltd},
  issn = {1879307X},
  doi = {10.1016/j.tics.2018.01.009},
  langid = {english}
}

@article{Pezzulo2024a,
  title = {Active Inference as a Theory of Sentient Behavior},
  author = {Pezzulo, Giovanni and Parr, Thomas and Friston, Karl},
  date = {2024-02-01},
  journaltitle = {Biological Psychology},
  shortjournal = {Biological Psychology},
  volume = {186},
  pages = {108741},
  issn = {0301-0511},
  doi = {10.1016/j.biopsycho.2023.108741}
}

@book{Russell2021a,
  title = {Artificial Intelligence: {{A}} Modern Approach},
  author = {Russell, Stuart and Norvig, Peter},
  date = {2021},
  series = {Pearson Series in {{Artificial Intelligence}}},
  edition = {4},
  publisher = {Pearson},
  isbn = {978-1-292-15396-4},
  langid = {english}
}

@book{Sarkka2013a,
  title = {Bayesian {{Filtering}} and {{Smoothing}}},
  author = {Särkkä, Simo},
  date = {2013},
  % series = {Institute of {{Mathematical Statistics Textbooks}}},
  publisher = {Cambridge University Press},
  location = {Cambridge},
  doi = {10.1017/CBO9781139344203},
  abstract = {Filtering and smoothing methods are used to produce an accurate estimate of the state of a time-varying system based on multiple observational inputs (data). Interest in these methods has exploded in recent years, with numerous applications emerging in fields such as navigation, aerospace engineering, telecommunications and medicine. This compact, informal introduction for graduate students and advanced undergraduates presents the current state-of-the-art filtering and smoothing methods in a unified Bayesian framework. Readers learn what non-linear Kalman filters and particle filters are, how they are related, and their relative advantages and disadvantages. They also discover how state-of-the-art Bayesian parameter estimation methods can be combined with state-of-the-art filtering and smoothing algorithms. The book's practical and algorithmic approach assumes only modest mathematical prerequisites. Examples include Matlab computations, and the numerous end-of-chapter exercises include computational assignments. Matlab code is available for download at www.cambridge.org/sarkka, promoting hands-on work with the methods.},
  isbn = {978-1-107-03065-7},
  file = {/home/filconscious/Library/Bayesian Filtering and Smoothing - Sarkka.pdf}
}

@article{Seth2016a,
  title = {Active Interoceptive Inference and the Emotional Brain},
  author = {Seth, Anil K. and Friston, Karl J.},
  date = {2016},
  journaltitle = {Philosophical Transactions of the Royal Society B: Biological Sciences},
  shortjournal = {Philos. Trans. R. Soc. B Biol. Sci.},
  volume = {371},
  number = {1708},
  pages = {1--10},
  issn = {14712970},
  doi = {10.1098/rstb.2016.0007},
  langid = {english}
}

@article{Smith2022a,
  title = {A Step-by-Step Tutorial on Active Inference and Its Application to Empirical Data},
  author = {Smith, Ryan and Friston, Karl J. and Whyte, Christopher J.},
  date = {2022-04-01},
  journaltitle = {Journal of Mathematical Psychology},
  shortjournal = {Journal of Mathematical Psychology},
  volume = {107},
  pages = {102632},
  issn = {0022-2496},
  doi = {10.1016/j.jmp.2021.102632}
}

@book{Sutton2018a,
  title = {Reinforcement Learning: An Introduction},
  author = {Sutton, Richard S. and Barto, Andrew G.},
  date = {2018},
  series = {Adaptive {{Computation}} and {{Machine Learning}}},
  publisher = {The MIT Press},
  url = {http://incompleteideas.net/book/the-book-2nd.html},
  isbn = {978-0-262-03924-6},
  langid = {english}
}

@inproceedings{Todorov2008a,
  title = {General Duality between Optimal Control and Estimation},
  booktitle = {47th {{IEEE Conference}} on {{Decision}} and {{Control}}},
  author = {Todorov, Emanuel},
  date = {2008},
  pages = {4286--4292},
  doi = {10.1109/CDC.2008.4739438},
  eventtitle = {47th {{IEEE Conference}} on {{Decision}} and {{Control}}},
  langid = {english}
}

@inproceedings{Toussaint2006a,
  title = {Probabilistic Inference for Solving Discrete and Continuous State {{Markov}} Decision Processes},
  booktitle = {Proceedings of the 23rd International Conference on Machine Learning},
  author = {Toussaint, Marc and Storkey, Amos},
  date = {2006},
  series = {{{ICML}} '06},
  pages = {945--952},
  publisher = {Association for Computing Machinery},
  location = {New York, NY, USA},
  doi = {10.1145/1143844.1143963},
  isbn = {1-59593-383-2}
}

@incollection{Wiese2017c,
  title = {Vanilla {{PP}} for Philosophers: A Primer on Predictive Processing},
  booktitle = {Philosophy and {{Predictive Processing}}},
  author = {Wiese, Wanja and Metzinger, Thomas K.},
  editor = {Metzinger, Thomas K. and Wiese, Wanja},
  date = {2017},
  publisher = {MIND Group},
  location = {Frankfurt am Main},
  doi = {10.15502/9783958573024},
  isbn = {978-3-95857-302-4},
  langid = {english}
}
\clearpage
\beginsupplement%

\section{Mathematical Background}\label{secx:math-background}

\subsection{Notation}\label{ssecx-notation}
\begin{table}[H]\label{tab:summary-notation}
    \caption{Summary of notation.}
    \centering
    \resizebox{\textwidth}{!}{%
    \begin{tabular}{lcc}
        \toprule
        Symbol & Meaning\\
        \midrule
        \(t, \tau, \ntime\) & integers, i.e., generic, current, and terminal time index, respectively\\
        \(1:t\), \(1:\ntime\) & sequences of times steps up to \(t\) and \(\ntime\), respectively\\
        \(\polhorizon\) & integer, length of a sequence of actions (i.e., the policy horizon), in general \(H \leq T\) \\
        \(\numpolicies\) & integer, the number of action sequences (policies) the agent considers \\
        \(X_{t}\) & random variable with support in \(\mathcal{X}\), and with \(t \in [1, \ntime]\) \\
        \(\seqv{X}{1}{T}, \seqv{x}{1}{T}\) & sequence of random variables with time index and related values \\
        \(\mathbf{X}_{:,j}\) & \(j\)th column of matrix \(\mathbf{X}\) or random vector associated with that column \\
          \(\prob{X_{t}}, \prob{x_{t}}\) & probability distribution of random variable \(X_{t}\) and probability that \(X_{t} = x_{t}\) (when defined) \\
        \(\entropy[X_{t}]\) & Shannon entropy of random variable \(X_{t}\) \\
        \(\text{Cat}(\mathbf{x}_{t})\) & categorical distribution with vector of parameters \(\mathbf{x}_{t}\) \\
        \(\text{Dir}(\mathbf{x}_{t})\) & Dirichlet distribution with vector of parameters \(\mathbf{x}_{t}\) \\
        \(\statespace\) & finite set of cardinality \(\card{\statespace}\), i.e., the set of states \\
        \(\obsspace\) & finite set of cardinality \(\card{\obsspace}\), i.e., the set of observations \\
        \(\actionspace\) & finite set of cardinality \(\card{\actionspace}\), i.e., the set of actions \\
        \(\actionspace^{\polhorizon}\) & finite set of action tuples (\(\polhorizon\)-fold Cartesian product) \\
        \(\Pi\) & subset of action sequences, i.e.,  \(\Pi \subseteq\actionspace^{\polhorizon}\) \\
        \((\seqactions{\polhorizon})\) & element in \(\actionspace^{\polhorizon}\), shortened as \((\seqv{a}{1}{\polhorizon})\) \\
        \(\statevar_{t}, \obsvar_{t}, \actionvar_{t}, \policy\) & categorical random variables with support in \(\statespace, \obsspace, \actionspace, \policyspace\), respectively, i.e., \(\statevar_{t} \sim \text{Cat}(\mathbf{s}_{t}), \dots\) \\
        \(\state_{t}, \obs_{t}, \action_{t}, \policy_{k}\)  & elements in \(\statespace, \obsspace, \actionspace, \policyspace\), respectively, where \(k \in [1, \numpolicies]\) and \(\numpolicies \in [1, \card{\policyspace}]\) \\
        \(\stateparams_{t}, \obsparams_{t}, \policyparams\), & column vectors of parameters for state, observation, and policy random variables, respectively \\
        \(\stateparams_{t}[i], \obsparams_{t}[i], \policyparams[i]\) & \(i\)th element of the parameter vector for state, observation, and policy random variables, respectively \\
        \(\ohstate_{t}, \ohobs_{t}, \ohaction_{t}\) & one-hot vectors, i.e., for some \(i \in \card{\mathcal{S}}\), \(\ohstate_{t}[i] = 1\), and \(\ohstate_{t}[j] = 0, \forall j \neq i\), similarly for \(\ohobs_{t}, \ohaction_{t}\)\\
        \(\Transition\) & transition map/function \\
        \(\Emission\) & emission map/function \\
        \(\tprob{\state_{t}}\) & transition probability distribution (returned by \(\Transition\)) \\
        \(\eprob{\obs_{t}}\) & emission probability distribution (returned by \(\Emission\)) \\
          \(\pprob{\state_{t}}, \pprob{\obs_{t}}\) & stationary distributions over \(\statespace\) and \(\obsspace\), respectively \\
        \(\decision\) & function that maps an element \(x\) of a state space \(\mathcal{X}\) to an actiont in \(\actionspace\) \\
        \(\genmodel\) & generative model (collection of probability distributions) \\
        \(\obsmap\) & matrix in \(\mathbb{R}^{n \times m}\) storing parameters of \(P(\obsvar_{t} \vert \statevar_{t-1})\) (the same for any \(t\)) \\
        \(\transmap\) & tensor in \(\mathbb{R}^{\card{\actionspace} \times m \times m}\) storing parameters of \(P(\statevar_{t} \vert \statevar_{t-1})\) (the same for any \(t\)) \\
        \(\transmap^{a_{1}}, \dots, \transmap^{a_\mathtt{d}}\) & state-transition matrices in \(\mathbb{R}^{m \times m}\) for each available action, \(d = \card{\actionspace}\) \\
          \(\fe\) & free energy \\
          \(\fe_{a_{\tau-1}}\) & action-conditioned free energy in vanilla active inference \\
          \(\fe_{\policy_{k}}\) & policy-conditioned free energy in variational message passing \\
          \(\efe_{t}\) & single-step expected free energy \\
          \(\totefe\) & total expected free energy, i.e., sum of expected free energies for \(H\) time steps in the future \\
          \(\nabla_{\stateparams_{t}} \fe_{\policy_{k}}\) & gradient of policy-conditioned free energy with respect to vector of parameters \(\stateparams_{t}\) \\
          \(\nabla_{\policyparams} \fe\) & gradient of free energy with respect to vector of policy parameters \(\policyparams\) \\
          \(\boldsymbol{\fe}_{\policy}^{\intercal}\) & row-vector in \(\mathbb{R}^{1 \times \card{\policyspace}}\) of policy-conditioned free energies \\
          \(\boldsymbol{\totefe}^{\intercal}\) & row-vector in \(\mathbb{R}^{1 \times \card{\policyspace}}\) of total expected free energy \\
         & \\
        \bottomrule
    \end{tabular}
  }
\end{table}

\subsection{Categorical random variable}\label{ssecx:categorical-rv}
If \(\statevar_{t}\) follows a categorical distribution with \(m\) categories or values, then the random variable can be realised in \(m\) different ways, each having a corresponding probability \(p\), where \(\sum_{j=1}^{m} p_{j} = 1\) (the probabilities sum to one).
We can then indicate one such value of \(\statevar_{t}\) (the state random variable at \(t\)) as \(\state_{t}\)
and use \(P(S_{t}=s_{t})\) for the probability that the random variable takes that value, i.e., \(P(S_{t}=s_{t}) = p_{j}\) for some \(j \in [1, m]\).
Note that these probabilities are regarded as parameters of the categorical distribution and can be stored in a vector, \(\stateparams_{t} = [p_{1}, \dots, p_{m}]^{\intercal}\), which allowed us to write \(\statevar_{t} \sim \text{Cat}(\stateparams_{t})\) in the main text.

\subsection{Dirichlet distribution as conjugate prior}\label{ssecx:dirichlet-conj-prior}
The choice of Dirichlet distributions to model the parameters of categorical distributions is not arbitrary.
In the context of Bayesian inference, the former are \emph{conjugate priors} for the latter, meaning that using a Dirichlet distribution as a prior distribution in the presence of a categorical likelihood (the other term in the numerator of Bayes' rule) results in a posterior distribution (the outcome of Bayesian inference) with the same form as the prior, i.e., a Dirichlet posterior.
In other words, this simplifies the process of inferring posterior parameters.
So, for instance, if we consider inference on \(\obsmap_{:,i} \sim \text{Dir}(\boldsymbol{\alpha}_{i})\), this would roughly amount to adjusting the parameters \(\boldsymbol{\alpha}_{i}\) based on the acquired observations, resulting in the Dirichlet posterior \(P^{\ast}(\mathbf{A}_{:,i}) \coloneq \text{Dir}(\boldsymbol{\alpha}_{i}^{\ast})\), where we used the asterisk to identify the posterior and the new, revised, set of Dirichlet parameters.
For a more detailed introduction to Bayesian inference with conjugate priors~\parencite[see, e.g.,][Ch. 6]{Deisenroth2020a}.


\subsection{Derivation of the Free Energy Objective}\label{ssecx:deriv-fe-obj}
Given a sequence of observations \(\seqobs{T}\), the goal of performing Bayesian inference using~\cref{eq:aif-bayes-pomdp} is to derive the posterior probability of a certain state trajectory of the POMDP, \(\seqstates{T}\), the most probable policy pursued so far, $\policy$, and the most probable parameters specifying state-observation mapping $\obsmap$, and state transitions $\transmap$.
However, these posterior probabilities cannot be computed analytically using that equation because this would require evaluating the denominator \(P(\obsvar_{1:T}) = \sum \condprob{\seqv{\obsvar}{1}{\ntime}}{\seqv{\statevar}{1}{\ntime}, \policy, \obsmap, \transmap} \prob{\seqv{\statevar}{1}{\ntime}, \policy, \obsmap, \transmap}\) by considering all the possible sequences of observations, \(\seqobs{T}\), in relation to all possible state sequences, \(\seqstates{T}\), all possible policies, and all combinations of matrices' parameters.
This is however usually computationally intractable: with \(T=5\) and \(\obs_{t}, \state_{t} \in [0,8]\), i.e., states and observations taking one of \(9\) possible values, there are \(59049\) observation sequences to evaluate by summing \(59049\) probabilities, each related to one state sequence.
In other words, there would be a total of \(59049^{2}\) values to be computed, and this is omitting the combinations with respect to policies and matrices' parameters.

Variational Bayesian inference is a technique to make the above inference problem more tractable.
It involves the introduction of an approximate posterior distribution, \(\varprob{\cdot}\), also called the variational posterior, that ought to becomes ``as close as possible'' to the true posterior.
This can be achieved by solving a tractable optimisation problem, thereby avoiding the intractable computations described above.

The variational posterior is one of the defining elements of the active inference agent (see~\cref{def:aif-agent}) and it is commonly indicated by \(\varprob{\statevar_{1:\ntime}, \policy, \obsmap, \transmap}\).
To make this approximate posterior ``as close as possible'' to the true one, \(\condprob{\statevar_{1:\ntime}, \policy, \obsmap, \transmap}{\obsvar_{1:\ntime}}\), one usually minimises the Kullback-Leibler (KL) divergence, \(\KL\).

Therefore, we can write (\emph{cf}. Equation \(2\) in~\parencite{DaCosta2020b}):

\begin{subequations}\label{eq:aif-kl}
    \begin{align}
    & \infdiv[\Big]{\varprob{\statevar_{1:\ntime}, \policy, \obsmap, \transmap}}{\condprob{\statevar_{1:\ntime}, \policy, \obsmap, \transmap}{\obsvar_{1:\ntime}}} \label{eq:aif-kl-pomdp-a} \\
    & = \mathbb{E}_{Q} \Bigl[\log \varprob{\statevar_{1:\ntime}, \policy, \obsmap, \transmap} - \log \condprob{\statevar_{1:\ntime}, \policy, \obsmap, \transmap}{\obsvar_{1:\ntime}} \Bigr] \geq 0 \label{eq:aif-kl-pomdp-b} \\
        \begin{split}\label{eq:aif-kl-pomdp-c}
            & = \mathbb{E}_{Q} \Bigl[\log \varprob{\statevar_{1:\ntime}, \policy, \obsmap, \transmap} - \log \prob{\statevar_{1:\ntime}, \policy, \obsmap, \transmap, \obsvar_{1:\ntime}} + \log \prob{\obsvar_{1:\ntime}} \Bigr]
        \end{split} \\
        \begin{split} \label{eq:aif-kl-pomdp-d}
            & = \mathbb{E}_{Q} \Bigl[\log \varprob{\statevar_{1:\ntime}, \policy, \obsmap, \transmap} - \log \prob{ \obsvar_{1:\ntime}, \statevar_{1:\ntime}, \policy, \obsmap, \transmap} \Bigr] + \log \prob{\obsvar_{1:\ntime}},
        \end{split}
    \end{align}
\end{subequations}

where each expectation $\mathbb{E}$ is with respect to \(Q(S_{1:\ntime}, \mathbf{A}, \mathbf{B}, \policy)\) and is shortened as \(\mathbb{E}_{Q}[\dots]\).

We obtain \cref{eq:aif-kl-pomdp-b} using the definition of the KL divergence.
Having noted that the second logarithm corresponds to the posterior probability distribution in \cref{eq:aif-bayes-pomdp} (Bayes' rule), we replaced it with the right-hand side of that equation and obtain \cref{eq:aif-kl-pomdp-c}.
Finally, since \(\log \prob{\obsvar_{1:\ntime}}\) does not involve variables over which the expectation is computed, we can take it out from the expectation and arrive at \cref{eq:aif-kl-pomdp-d}.

By defining the free energy \(\fe\) as the expectation in \cref{eq:aif-kl-pomdp-d}, i.e.:

\begin{equation}\label{eqx:aif-fe-pomdp}
    \fe \bigl[ \varprob{\statevar_{1:\ntime}, \policy, \obsmap, \transmap} \bigr] \coloneqq \mathbb{E}_{Q} \Bigl[\log \varprob{\statevar_{1:\ntime}, \policy, \obsmap, \transmap} - \log \prob{ \obsvar_{1:\ntime}, \statevar_{1:\ntime}, \policy, \obsmap, \transmap} \Bigr],
\end{equation}

and by the non-negativity of the KL divergence, it follows that the free energy is an upper bound on the negative logarithm of the sequence of observations:

\begin{equation}\label{eq:aif-fe-upper-bound}
- \log \prob{\obsvar_{1:\ntime}} \leq \fe \bigl[ \varprob{\statevar_{1:\ntime}, \policy, \obsmap, \transmap} \bigr].
\end{equation}

where the term on the left-hand side of \cref{eq:aif-fe-upper-bound} is known as \emph{surprisal} and measures how unlikely a sequence of observation is.

The closer the surprisal and the free energy are to each other, the closer the KL is to zero.
Thus, if the goal is to reduce the KL divergence between the variational posterior and the true posterior, this can be achieved by optimising the variational distributions' parameters on which \(\fe\) depends so that the free energy upper bound is as tight as possible.

Minimising the free energy is the tractable optimisation problem that provides a solution to the intractable Bayesian inference problem described earlier.
Also, the derivation reveals how performing Bayesian inference via free energy minimisation involves finding ways to increase the likelihood of observations (to reduce surprisal) because \(\fe\) can be seen as a proxy for surprisal.
Then, the notion that an active inference agent exists insofar as it can avoid unexpected states or observations and move towards desired ones can be understood as a consequence of free energy minimisation.

\section{Perception, Planning, and Action Selection via variational message passing}\label{secx:percep-plan-act}

In active inference, perception, planning, action selection, and learning can be regarded as different stages in the process of minimising the expression in~\cref{eq:policy-cond-fe-annotated}.
In the next few sections, we will illustrate the actual update equations used to implement them (\crefrange{ssecx:perception-se}{ssecx:learning-ea}).

\subsection{Perception as State Estimation}\label{ssecx:perception-se}
Since the goal is to minimize~\cref{eq:policy-cond-fe-annotated}, the expectations over the \textbf{state log-probabilities} need to be minimised so to concentrate the probability mass onto one realization of every state random variable \(\statevar_{1}, \dots, \statevar_{\ntime}\), depending on what the \emph{actual} trajectory afforded by the conditioning policy is (assigning equal probabilities to all those realizations would not achieve the minimum of this term and would misrepresent what a policy really achieves).
Thus, the minimisation here consists in updating the parameters of the variational distributions \(\varprob{\statevar_{1} | \policy_{k}}, \dots, \varprob{\statevar_{T} | \policy_{k}}\)  at every time step, with an increasingly longer sequence of collected observations.
This is akin to \textbf{perception} since the object of those operations is to uncover the causes of sensory evidence, i.e., observations, and everything occurs at a fast time scale, i.e., at every time step.
Perception is thus framed as the update of the parameters of the variational probability distributions \(\varprob{\statevar_{1} | \policy_{k}}, \dots, \varprob{\statevar_{T} | \policy_{k}}\), according to the available, collected evidence, with the goal of minimising \( \fe_{\policy_{k}} \bigl[ \varprob{\statevar_{1:T} | \policy_{k}} \bigr]\) (see~\cref{eq:policy-cond-fe-annotated}).


To obtain the update equations for the collection of parameters \(\stateparams_{1:T} \coloneq \stateparams_{1}, \dots, \stateparams_{T}\), where each \(\stateparams_{t}\) is the vector of probabilities defining \(\varprob{\statevar_{t}|\policy_{k}}\), we rewrite \(\fe_{\policy_{k}} [\varprob{\statevar_{1:T}|\policy_{k}}]\) in vectorised form, by substituting the vector of parameters for the various probability distributions, then we compute the gradients with respect to each of those vectors, namely:


\begin{equation}\label{eq:fe-grads-states}
    \nabla_{\stateparams_{t}} \fe_{\policy_{k}} (\stateparams_{1:T}) = \mathbf{1} + \log \stateparams_{t}^{\intercal} -
    \begin{cases}
        \ohobs_{t}^{\intercal} \cdot \mathbf{S}^{\boldsymbol{\alpha}} + \stateparams_{t+1}^{\intercal} \cdot \log \transmap^{\action^{t}} + \log \stateparams_{1}^{\intercal} & \text{if} \quad t = 1 \\
        \ohobs_{t}^{\intercal} \cdot \mathbf{S}^{\boldsymbol{\alpha}} + \stateparams_{t+1}^{\intercal} \cdot \log \transmap^{\action^{t}} + \Bigr[(\log \transmap^{\action^{t-1}}) \cdot \stateparams_{t-1}\Bigl]^{\intercal} & \text{if} \quad 1 < t \leq \tau \\
        \stateparams_{t+1}^{\intercal} \cdot \log \transmap^{\action^{t}} + \Bigr[(\log \transmap^{\action^{t-1}}) \cdot \stateparams_{t-1}\Bigl]^{\intercal} & \text{if} \quad t > \tau
    \end{cases}
\end{equation}

where:
\begin{itemize}
  \item \(\ohobs_{t}^{\intercal}\) is the transposed (one-hot) observation vector, i.e., \(\ohobs_{t}^{\intercal}[i] = 1\) if \(i\) corresponds to the observation category (value) of \(\obsvar_{t}\) observed at \(t\);
  \item \(\transmap^{\action^{t}}\) is the transition matrix for the action \(\action_{j}\), with \(j \in [1, \card{\mathcal{A}}]\), that the policy \(\policy_{k}\) mandates at time step \(t\), note that we indicate such action by \(\action^{t}\), omitting the reference to the policy and the subscript \(j\) to avoid too much notational clutter;
  \item \(\mathbf{S}^{\boldsymbol{\alpha}} \coloneq \psi\Bigr([\boldsymbol{\alpha}_{1:m}]\Bigl) - \psi\Bigr(\mathbf{J}_{n, m} \cdot [\boldsymbol{\alpha}_{1:m}]\Bigl)\), where:
  \begin{itemize}
    \item \(\boldsymbol{\alpha}_{1:m} \coloneq \boldsymbol{\alpha}_{1}, \dots, \boldsymbol{\alpha}_{m}\) are the column vectors of Dirichlet parameters for \(\obsmap\) (one for each column, see~\cref{ssec:aif-bayesian-inf}), and with \([\boldsymbol{\alpha}_{1:m}] \in \mathbf{R}^{n \times m}\) representing the matrix whose columns are those vectors;
    \item \(\psi([\boldsymbol{\alpha}_{1:m}])\) is the digamma function applied element-wise to the matrix of Dirichlet parameters whereas \(\psi(\mathbf{J}_{n, m} \cdot [\boldsymbol{\alpha}_{1:m}])\) is the digamma function applied to the result of the matrix multiplication between the matrix of ones \(\mathbf{J}_{n, m}\) and \([\boldsymbol{\alpha}_{1:m}]\) (note that a column \(j\) of the resulting matrix is filled with the same value, namely, the dot product or sum \(\mathbf{1}^{\intercal} \boldsymbol{\alpha}_{j}\), usually indicated by \(\alpha_{0}\), \emph{cf}.~\parencite{DaCosta2020b})~\footnote{The difference between digamma functions appears because of the term \(\mathbb{E}_{\varprob{\statevar_{t}|\policy_{k}} \varprob{\obsmap}} [ \log \condprob{\obs_{t}}{\statevar_{t}, \obsmap} ]\). If we consider the gradient of the free energy with respect to the element \(\stateparams_{t}[i]\) of the parameter vector \(\stateparams_{t}\), then we obtain the expression \(\mathbb{E}_{\varprob{\obsmap_{:,i}}} [ \log \obsmap_{:,i} ]\), assuming we did not know the value of \(\obsvar_{t}\). That expression is the expectation of the log of a Dirichlet-distributed random vector which is equal to \(\psi (\boldsymbol{\alpha}_{i}) - \psi(\boldsymbol{\alpha}_{0})\) where \(\boldsymbol{\alpha}_{0}\) is the vector whose elements are all equal to \(\mathbf{1}^{\intercal} \boldsymbol{\alpha}_{i}\). Considering all the elements of \(\stateparams_{t}\) gives us the matrix \(\mathbf{S}^{\boldsymbol{\alpha}}\) which is vector-multiplied by \(\ohobs_{t}\) to arrive at the correct expression for the gradient.};
  \end{itemize}
  \item finally, \(\cdot\) stands for the inner product, \(\log\) is applied element-wise to both the elements of vectors and matrices, and the vectors of parameters are now transposed because we consider the \emph{numerator layout} when taking the gradient of a vector-valued function~\footnote{The numerator layout specifies the order in which to compute the partial derivatives of a vector-valued function, resulting in a Jacobian matrix whose numbers of columns and rows correspond to the number of function's inputs and outputs, respectively. Given a function \(\mathbf{f}: \mathbb{R}^{n} \rightarrow \mathbb{R}^{m}\) that maps a vector \(\mathbf{x} \in \mathbb{R}^{n}\) to a vector \(\mathbf{y} \in \mathbb{R}^{m}\), the \(m\) elements in \(\mathbf{y}\) define the rows of the Jacobian whereas the \(n\) elements in \(\mathbf{x}\) define its columns, i.e., \(\mathbf{J} \in \mathbb{R}^{m \times n}\). For instance, if we have the function \(f: \mathbb{R}^{n} \rightarrow \mathbb{R}^{1}\), then the Jacobian is a row vector, i.e., \(\nabla_{\mathbf{x}} f = [ \nicefrac{\partial f(\mathbf{x})}{\partial x_{1}}, \dots, \nicefrac{\partial f(\mathbf{x})}{\partial x_{n}} ]\), where \(x_{1}, \dots, x_{n}\) are the elements of \(\mathbf{x}\). In the main text, when the free energy is considered with respect to one of its vectors of parameters, e.g.~\(\stateparams_{t}\), it can be seen precisely as a vector-valued function, \(\fe_{\policy_{k}}(\stateparams_{t}): \mathbb{R}^{n} \rightarrow \mathbb{R}\), with the Jacobian being a row vector.}.
\end{itemize}

By setting the above gradients to zero, we can derive analytically the new \emph{unnormalised} parameters of the various \(\varprob{\statevar_{t} | \policy_{k}}\) that minimize free energy, that is:

\begin{equation}\label{eq:fe-update-states}
   \log \stateparams_{t}^{\intercal} =
\begin{cases}
\ohobs_{t}^{\intercal} \cdot \mathbf{S}^{\boldsymbol{\alpha}} + \stateparams_{t+1}^{\intercal} \cdot \log \transmap^{\action^{t}} + \log \stateparams_{1}^{\intercal} - \mathbf{1} & \text{if} \quad t = 1 \\
\ohobs_{t}^{\intercal} \cdot \mathbf{S}^{\boldsymbol{\alpha}} + \stateparams_{t+1}^{\intercal} \cdot \log \transmap^{\action^{t}} + \Bigr[(\log \transmap^{\action^{t-1}}) \cdot \stateparams_{t-1}\Bigl]^{\intercal} - \mathbf{1} & \text{if} \quad 1 < t \leq \tau \\
\stateparams_{t+1}^{\intercal} \cdot \log \transmap^{\action^{t}} + \Bigr[(\log \transmap^{\action^{t-1}}) \cdot \stateparams_{t-1}\Bigl]^{\intercal} - \mathbf{1} & \text{if} \quad t > \tau
\end{cases}
\end{equation}

Since these parameters stand for probabilities defining categorical probability distributions, we need to impose that each set of parameters sums to one.
This is usually done by applying the softmax function \(\softmax{\cdot}\) to the expressions in~\cref{eq:fe-update-states} (an alternative method is to set up the whole problem as one of constrained optimization and use Lagrange multipliers).

In active inference studies that aim to describe neuronal dynamics as a form of gradient descent on free energy, where the gradient can be considered as prediction error, the following update rule is used instead.

\begin{equation}\label{eq:fe-descent}
\stateparams_{t} \coloneq \softmax{\stateparams_{t} - \nabla_{\stateparams_{t}} \fe_{\pi_{i}}},
\end{equation}

where again the softmax function \(\softmax{\cdot}\) is used to make sure the parameters represent legitimate probability distributions.


\subsection{Planning with Expected Free Energy}\label{ssecx:planning-efe}
The minimization of the free energy with respect to the policy random variable, \(\policy_{k}\), also occurs at every time step and can be associated with the cognitive operations of \textbf{planning}, but it requires a separate treatment that considers the expectation over \(\fe_{\policy_{k}} \bigl[ \varprob{\statevar_{1:T} | \policy_{k}} \bigr]\) (last term of~\cref{eq:fe-unpacked}) and the \emph{expected free energy}.

Once the agent has updated its probabilistic beliefs about the past, present, and future states variables, it can proceed to update its probabilistic beliefs over the set of policies.
This is achieved by predicting what is most likely to happen if a certain policy is followed and by scoring each policy depending on whether it would result in a desired sensorimotor trajectory.
An updated probability distribution over the policies can then be paired with a decision rule \(\decision\) (see \cref{def:aif-agent}) to determine what action the agent will perform at the next time step.

For each policy, this process involves computing an \emph{expected free energy}, one for each future time step of the potential trajectory that the policy might realize:

\begin{equation}
    \label{eqx:efe-unpacked}
    \begin{split}
        \efe_{t}(\policy_{k}) =
        & \underbrace{\mathbb{E}_{\varprob{\statevar_{t}|\policy_{k}}} \Bigl[\entropy \bigl[\condprob{\obsvar_{t}}{\statevar_{t}} \bigr] \Bigr]}_{\textsc{Ambiguity}} - \underbrace{\mathbb{E}_{\condprob{\obsvar_{t}}{\statevar_{t}} \varprob{\statevar_{t}|\policy_{k}}} \Bigl[\kldiv \bigl[\varprob{\obsmap|\obs_{t}, \state_{t}}| \varprob{\obsmap} \bigr]\Bigr]}_{\textsc{A-Novelty}} \\
        & + \underbrace{\kldiv \bigl[ \varprob{\statevar_{t}|\policy_{k}} | \pprob{\statevar_{t}} \bigr]}_{\textsc{Risk}} - \underbrace{\mathbb{E}_{\varprob{\statevar_{t+1}|\policy_{k}} \varprob{\statevar_{t}|\policy_{k}}} \Bigl[\kldiv \bigl[\varprob{\transmap| \state_{t+1},\state_{t}}|\varprob{\transmap}\bigr]\Bigr]}_{\textsc{B-novelty}}.
    \end{split}
\end{equation}

In other words, each of these expected free energies can be approximately regarded as the free energy most likely to be registered at a future time step, if the actions of the conditioning policy are performed up to that point (see~\textcite{Millidge2021c} for why, technically, describing the expected free energy in this way is not entirely correct, but nonetheless common in the active inference literature).
The sum of these expected free energies is indicative of how much the considered policy would allow the agent to reduce uncertainty and achieve a preferred distribution over states at some point in the future.

The total expected free energy for policy \(\policy_{k}\),  \(\totefe(\policy_{k})\), is then defined as the sum of expected free energies at future time steps up to the policy horizon, \(\polhorizon\), i.e.:

\begin{equation}
    \totefe(\policy_{k}) = \sum_{t=\tau + 1}^{\polhorizon} \efe_{t}(\policy_{k}).
\end{equation}

The different terms that constitute the expected free energy tend to be associated with distinct \emph{behavioural drives}.

The \textbf{ambiguity} term is the expected value of the entropy, \(\entropy{\cdot}\), of the state-observation mappings, and quantifies the uncertainty about future outcomes given hidden states.
A policy for which this term is low is a policy that drives the agent towards unambiguous parts of the environment.

The \textbf{risk} term introduces another defining component of the active inference agent (\cref{def:aif-agent}), i.e., a probability distribution that specifies its preference(s) or goal(s), expressed in this case over state variables.
In other words, this is a probability distribution with the probability mass concentrated on one or a few states, representing states in the environment to which the agent is moving.
The KL divergence between the expected states in the future and these agent's preferences over states quantifies how much the policy will allow the agent to get there, thereby achieving its goal(s).

The two \textbf{novelty} terms are expected values of KL divergences between the posterior and prior distributions over generative model's parameters, i.e., those of state-observation mappings and transition probabilities.
Since they are subtracted from the expected free energy and the agent is looking for a policy that minimises it, the agent is pushed to pick a policy whose actions may increase these terms, i.e., leading to environmental consequences that were not very well captured by the current generative model.
Therefore, these terms are interpreted as capturing the \emph{exploratory drives} of an active inference agent insofar as they score a policy based on how likely it will disclose as-yet unknown parts of the environment, which may be informative about state-observation mappings and transition probabilities.
In other words, a policy for which these terms are very high will provide new information to the agent, not already encoded in its generative model.

To understand how the computation of expected free energy fit in the process of free energy minimisation, i.e., of minimising the free energy in~\cref{eq:fe-unpacked}, we take the gradient of that expression with respect to the parameters \(\policyparams\) of \(\varprob{\policy}\), obtaining:

\begin{equation}\label{eq:fe-grad-policy}
\nabla_{\policyparams} \fe[\varprob{\policy}] = \ln \policyparams_{Q}^{\intercal} - \ln \policyparams_{P}^{\intercal} + \boldsymbol{\fe}_{\policy}^{\intercal} + \mathbf{1},
\end{equation}

where \(\policyparams_{Q}^{\intercal}\) and \(\policyparams_{P}^{\intercal}\) are the row vectors of parameters of \(\varprob{\policy}\) and \(\prob{\policy}\), respectively, and \(\boldsymbol{\fe}_{\policy}^{\intercal}\) is the row vector of policy-conditioned free energies (one for each policy, i.e., for each value the policy random variable can take, see~\cref{eq:policy-cond-fe-annotated}).

As done earlier, setting the above gradient to zero gives us the new, \emph{unnormalised} vector of parameters for \(\varprob{\policy}\):

\begin{equation}\label{eq:fe-policy-update-unorm}
\ln \policyparams_{Q}^{\intercal} = \ln \policyparams_{P}^{\intercal} - \boldsymbol{\fe}_{\policy}^{\intercal} - \mathbf{1}.
\end{equation}

Again, to obtain a proper, normalised probability distribution, the softmax map is applied.
For this update equation there is a further issue, as we have not clarified what the (column) vector of parameters \(\policyparams_{P}\) represents.
This vector stores the probabilities that define the distribution \(\prob{\policy}\) which gets compared to \(\varprob{\policy}\) in~\cref{eq:fe-unpacked} by means of the KL divergence.
Therefore, it can be regarded as the \emph{prior} probability distribution over the policies provided by the agent's generative model.
Essentially, the above equation says that in order to update the probabilistic beliefs about the policy random variable, the current evidence represented by the free energy values has to be integrated with the agent's prior beliefs (this is in line with the notion of Bayesian inference, and it holds true for the other variational updates as well).

The crucial question is how that prior should be specified.
According to active inference, the answer is provided by expected free energy insofar as it manages to balance instrumental and epistemic values when it comes to policy and action selection (offering a solution to the exploration-exploitation dilemma).
In particular, we have that \(\policyparams_{P} \coloneq \softmax{-\boldsymbol{\totefe}}\), where \(\boldsymbol{\totefe}\) is the vector of expected free energies (one for each policy, i.e., for each value the policy random variable can take), giving us the following update rule:

\begin{equation}\label{eq:fe-policy-update-norm}
\policyparams_{Q}^{\intercal} = \softmax{-\boldsymbol{\totefe}^{\intercal} - \boldsymbol{\fe}_{\policy}^{\intercal}},
\end{equation}

where the constant \(\mathbf{1}\) can be dropped as it does not affect the softmax function (whether using expected free energy represents a principled way of specifying that prior has been a subject of debate, see~\parencite{Millidge2020d})~\footnote{In some active inference works, an additional prior term is included in the softmax, i.e., a probability distribution representing preferences for one or more “habitual” policy/policies, see, e.g.,~\cite[33]{Heins2022a}.}.

The above parameters can also be used to update the policy-independent state probabilities, that is:

\begin{equation}\label{eq:fe-policy-indep-states}
\varprob{\statevar_{\tau}} = \sum_{k=0}^{\card{\policyspace}} \varprob{\statevar_{\tau} | \policy_{k}} \varprob{\policy_{k}},
\end{equation}

where the marginal probabilities over the state random variable at \(\tau\) are computed using the updated policy probabilities.
These marginal probabilities provide an indication of what the agent believes are the most probable state values at a certain point in the training process.
As learning progresses, they should converge to the probabilities representing the agent's preferences (over states).

After obtaining an approximate posterior probability distribution over policies, the agent can proceed to implement an action selection procedure, which will be described next.




\subsection{Action Selection}\label{ssecx:action-selection}
With updated probabilistic beliefs on past and future sensorimotor trajectories, afforded by different policies, and a new probability distribution over them, the agent is equipped to select an action towards the goal state.
The decision rule \(\decision\) of an active inference agent implements the action-selection mechanism which can take many forms.

One strategy is to pick the policy with the highest probability and perform one of its actions, since the objective of the agent is to minimise free energy now and in the future, and the look-ahead operations with expected free energy ultimately scored the different policies based on that requirement.
This could be described as a kind of \emph{greedy} strategy that priorities executing whatever policy was deemed to be more conducive to low free energy states in the future.
In this case, the decision rule would be a function \(\decision: \policyspace \times \lbrace 1, \dots, \ntime \rbrace \rightarrow \actionspace\) (see \cref{def:aif-agent}) that maps from the Cartesian product between the policy space and the set of time indices to the action space, in a such a way that at time step \(\tau\) the agent will pick the following action:

\begin{equation}\label{eq:greedy-action-sel}
\action_{t} = \decision(\policy_{*}, t) = \policy_{*}[t]
\end{equation}

i.e., simply, this decision function returns the action that the policy chosen as input (the most probable in this case) specifies for that time step (recall that each policy is a sequence of time-indexed actions).
Alternatively, the policy that goes into the decision function could be sampled from \(\varprob{\policy}\), adding some randomness into the action selection procedure.

The above strategies narrowly focus on the action specified by a single policy.
That is, once a policy has been picked, the action that the policy dictates is going to be performed regardless of what the other policies suggest.
This may turn out to be a suboptimal way of selecting the next action if in the current learning phase the agent does not have (yet) a highly probable policy.
For instance, if the selected policy is only slightly more probable than the others, and all these agree on the \emph{same} action to perform, then it might be better to perform the action backed by most policies than the action indicated by the most probable one.
In light of the above considerations, in \textcite{DaCosta2020b} and in the following experiments, the agent picks the most likely action under all possible policies.
This specific action is found by computing a \emph{Bayesian model average}, that is:

\begin{equation}\label{eq:bayes-m-avg-policy}
\action_{t} = \decision(\policyspace, t) = \argmax_{\action \in \actionspace} \Biggl(\sum_{\policy_{k} \in \policyspace} \delta_{\action, c^{\policy_{k}}_{t}} \varprob{\policy_{k}} \Biggr),
\end{equation}

where, given a candidate action \(\action\) from the set of possible actions \(\actionspace\) and for every policy \(\policy_{k}\), \(k \in [1, \dots, \numpolicies]\), we sum the products \(\delta_{\action, c^{\policy_{k}}_{t}} \varprob{\policy_{k}}\).
The first factor of those products is the \emph{Kronecker delta} between the candidate action \(\action\) and the action that the considered policy dictates at time step \(t\), indicated by \(c^{\policy_{k}}_{t}\).
If these two actions are equal, i.e., \(\action = c^{\policy_{k}}_{t}\), then \(\delta_{\action, c^{\policy_{k}}_{t}} = 1\), and zero otherwise (this is the definition of the Kronecker delta for two variables).
The second factor is the probability of the given policy.

Thus, for every action we are going to compute a sum of policy probabilities, according to how many policies would suggest to perform that action at the current time step.
The action with the highest sum of probabilities will be selected.
If an action is not dictated by any policy, then the Kronecker delta will make it the worst possible candidate for what to do at time step \(t\).
If the same action is dictated by very likely policies, then that will be a good candidate.
In other words, this procedure recommends to pick the action that most policies agree upon at the current time step as long as those policies or a subset thereof have a high probability.

\subsection{Learning as Evidence Accumulation}\label{ssecx:learning-ea}
The second term in~\cref{eq:policy-cond-fe-annotated}, the negative sum of expectations of the \textbf{observation log-likelihoods}, involves log-probabilities of the observation values collected until the present time step \(\tau\).
Intuitively, the minimization of this term will occur when a high probability value for the current observation is matched by a high probability for that state value that truly generated the observation in the first place.
Concretely, if the agent strongly believes that \(\statevar_{t} = \state_{t}\), then it should be the case that the observation at \(t\) reflects that, i.e., that \(\obsvar_{t} = \obs_{t}\) and \(\obs_{t} = \state_{t}\), meaning that the categorical values of both random variables are equal (recall that for the categorical random variables in question \(\state_{t}\) and \(\obs_{t}\) amount to indices that identify one of state/observation categories).
In contrast, for all the other realizations of the same state random variable, the probability of receiving that observation should be low (unless there are environmental states that admit of \emph{identical} observations).

The last term is the negative sum of expectations of the \textbf{transition log-likelihoods}.
This term captures how well the agent is modelling the transition dynamics of the environment.
That is, if performing the action policy \(\policy_{k}\) dictates at \(t-1\), from state \(\state_{t-1}\), leads to state \(\state_{t}\) at the next time step \(t\), then the corresponding probability that \(\statevar_{t} = \state_{t}\) given \(\statevar_{t-1} = \state_{t-1}\) (and the execution of that action) should be high (in other words, in expectation the agent should assign high probabilities to those state transitions that characterize a certain sequence of action).

The optimization of both log-likelihoods requires updating the parameters stored in the \(\obsmap\)-matrices  and \(\transmap\)-tensors, respectively.
This can also be done at every time step but the impact of the update is in general small because the collected observation will affect mostly a single state-observation mapping and state transitions, i.e., the one involving the present time step.
Therefore, the update of \(\obsmap\)-matrices  and \(\transmap\)-tensors occurs at a slower time scale as the agent acquires experience about the most common state-observation mappings and state transition in the given environment.
For this reason, these computational operations have been regarded as a form of \textbf{learning}, i.e., adaptation that require longer time (concretely, in implementations the update of these parameters is carried out in batch at the end of an episode or trajectory).

After the agent has collected a full trajectory of observations (e.g., at the end of an episode), learning consists in updating the parameters of the emission and transition maps, stored in the observation matrix \(\obsmap\) and in the transition tensor \(\transmap\), respectively (crucial components of the generative model used to model the environment, see \cref{sec:ae-interact}).

To derive the update rules, one starts again with the free energy introduced in~\cref{eq:fe-unpacked} and takes its gradient with respect to each of the Dirichlet distributions \(\prob{\obsmap_{:,1}}, \dots,  \prob{\obsmap_{:,m}}\), defined on the random vectors associated with the corresponding columns of the observation matrix, and each of the Dirichlet distributions \(\prob{\transmap^{a_{i}}_{:,1}}, \dots, \prob{\transmap^{a_{i}}_{:,m}}, \, \forall i \in [1, \dots, \card{\actionspace}]\), defined on the random vectors associated with the corresponding columns of the various matrices forming the transition tensor.
The derivation requires considering the KL divergence between many pairs of Dirichlet distributions.
For instance, the KL term \(\kldiv \bigl[\varprob{\obsmap}|\prob{\obsmap} \bigr]\) in~\cref{eq:fe-unpacked} is a more compact way of writing those KL divergences, that is:

\begin{equation}\label{eq:kl-unpacked}
\kldiv \bigl[\varprob{\obsmap}|\prob{\obsmap} \bigr] = \sum_{i=1}^{m} \kldiv \bigl[\varprob{\obsmap^{Q}_{:,i}} | \prob{\obsmap^{P}_{:,i}} \bigr],
\end{equation}

where \(\obsmap^{Q}_{:,i} \sim \text{Dir}(\boldsymbol{\alpha}^{Q}_{i})\) and \(\obsmap^{P}_{:,i} \sim \text{Dir}(\boldsymbol{\alpha}^{P}_{i})\), and the superscripts \(P\) and \(Q\) are used to indicate that we are dealing with different distributions and parameters, i.e., the prior and the posterior, respectively.

For reference, we state the update rules here as follows:

\begin{gather}
\boldsymbol{\alpha}^{Q}_{:,i} \coloneq \boldsymbol{\alpha}^{P}_{:,i} + \sum_{t=1}^{\ntime} \ohobs_{t} \odot \stateparams_{\tau}[i] \label{eq:update-A} \\
\boldsymbol{\beta}^{Q}_{:,i} \coloneq \boldsymbol{\beta}^{P}_{:,i} + \sum_{t=2}^{\ntime} \sum_{\pi_{k} \in \Pi} \delta_{\action, c^{\policy}_{t}} \varprob{\policy} \bigl( \stateparams_{t}^{\policy} \odot \stateparams_{t-1}^{\policy}[i] \bigr) \label{eq:update-B}
\end{gather}

\(\forall i \in [1, m]\), where recall: \(\stateparams_{t}\) is the vector of parameters for \(\varprob{\statevar_{t}|\policy_{k}}\); \(\ohobs_{t}\) is a one-hot vector indicating the category of observation that has been acquired at that time step; and the symbol \(\odot\) is used to indicate the element-wise (or Hadamard) product between a vector and the \(i\)th element of another vector (for a derivation of the above rules refer to~\parencite{DaCosta2020b}).

To gain some insight on these update rules, first notice that they return the parameters of (approximate) posterior Dirichlet distributions (indicated by superscript \(Q\)) through an adjustment of the prior parameters (indicated by superscript \(P\)).
Crucially, this revision of the prior parameters is made using the observations and the (updated) state-variable beliefs obtained from interacting with the environment for \(T\) time steps.

More specifically, in~\cref{eq:update-A} the value of an element in \(\boldsymbol{\alpha}^{P}_{:,i}\) is increased by the probability represented by \(\stateparams_{t}[i]\).
The particular value in \(\boldsymbol{\alpha}^{P}_{:,i}\) which is updated depends on the location of the \(1\) in the one-hot vector \(\ohobs_{t}\).
This captures the idea that if a certain observation value, say, \(\obs_{t}\), is repeatedly acquired at \(t\), and the probability \(\stateparams_{t}[i]\) associated with the \(i\)th value of \(\statevar_{t}\) is large, then the agent has some reasons to consider that observation as more likely when it is somewhat confident of being in state \(\statevar_{t} = s_{t}\).
In a nutshell, the probability \(\condprob{\obsvar_{t} = \obs_{t}}{\statevar_{t}=\state_{t}}\) should be increased proportionally to the probability that \(\statevar_{t} = \state_{t}\) at \(t\) when \(\obs_{t}\) has been registered from the environment.

Since the Dirichlet parameters \(\boldsymbol{\alpha}^{P}_{:,i}\) are used to sample the values in \(\obsmap_{:,i}\), which are in turn the parameters of the categorical distribution \(\condprob{\obsvar_{t}}{\statevar_{t}=\state_{t}}\), adjusting the Dirichlet parameters using the above rule has the desired learning effect, i.e., that of improving on the current state-observation mapping by capturing what are the most likely observational consequences of being in various states.

A similar reasoning applies to the Dirichlet parameters in \cref{eq:update-B}, where this time the key evidence is represented by the amount of state transitions of a certain type that have been encountered.
For instance, the more state transitions have been observed from state \(\state_{t-1}\) to state \(\state_{t}\) upon performing action \(\action^{t}\), the more the corresponding probabilities \(\condprob{\statevar_{t} = \state_{t}}{\statevar_{t} = \state_{t}}\) in matrix \(\transmap^{\action^{t}}\) should be increased.
One of the subtleties here is that these probability updates should be weighted according to how likely it is that an action was indeed performed to realise that state transition, which is achieved with the Kronecker delta term, \(\delta_{\action, c^{\policy}_{\tau}} \varprob{\policy}\), in the equation, i.e., by considering the probability of those policies suggesting that action at \(t\).

The accumulation of experience throughout an episode of interaction with the environment allows the agent to update its model of what the most probable observations and state transitions are in that environment.
These update rules establish a learning dynamics that is supposed to occur at a much slower temporal scale than perceptual inference, planning, and action selection so they are generally implemented at the end of an episode and/or a sequence of observation, and they have been described as a form of synaptic Hebbian plasticity \parencite[see, e.g,][]{Friston2016a,Friston2017a,DaCosta2020b}.

In summary, during an episode of length \(T\), an active inference agent goes through a phase of perceptual inference, planning, and action selection at each time step whereas at the end of the episode it capitalises on the acquired experience through a learning phase, before a new episode begins.
This active inference cycle is summarised in algorithm~\ref{pscode:aif-action-unaware} and algorithm~\ref{pscode:aif-action-aware} for the action-unaware and action-aware agent, respectively.

\begin{pscode}[label={pscode:aif-action-unaware}]{Action-unaware Active Inference}
\DontPrintSemicolon%
\SetKwInput{KwHyperp}{Hyperparameters:}
\KwHyperp{\(\ntime\), epidode length, \(N\), number of episodes, \(\ntime^{\text{max}} = \ntime \times N\), max number of steps, \(\numpolicies = \card{\policyspace}\), number of policies.}
\KwData{sensory observations, \(\obs_{1}, \dots, \obs_{t}\), and policies, \(\policy_{1} \dots \policy_{\numpolicies}\).}
\KwResult{updated \(Q(\statevar_{1}|\policy_{k}), \dots, Q(\statevar_{\ntime}|\policy_{k})\), \(\forall k \in [1, \numpolicies]\).}
\KwResult{updated \(Q(\policy)\), and next action, \(\action_{\tau}\).}
\KwResult{updated \(Q(\obsmap)\) and \(Q(\transmap)\)}
\BlankLine%
\For{\(t \in [1, \dots, \ntime^{\text{max}}]\)}{
\textcolor{Peach}{1. Perceptual Phase}: \\
\For{\(k \in [1, \dots, \numpolicies]\)}{
    \textcolor{ForestGreen}{a. Update probabilities over states}: \\
	\For{\(t \in [1, \dots , \ntime]\)}
	{\(\nabla_{\stateparams_{t}} \fe_{\policy_{k}} = 0 \Rightarrow \stateparams_{t} \coloneq \softmax{\ln \stateparams_{t}}\), see~\cref{eq:fe-update-states}}
	}
\textcolor{Peach}{2. Planning Phase}: \\
\For{\(k \in [1, \dots, \numpolicies]\)}{
    \textcolor{ForestGreen}{a. Compute total expected free energy}: \\
    \(\totefe(\pi_{k}) = \sum_{t = \tau + 1}^{T} \efe_{t}(\pi_{k})\) \\
    }
    \textcolor{ForestGreen}{b. Update probabilities over policies}: \\
\(\policyparams = \softmax{-\boldsymbol{\totefe}^{\intercal} - \boldsymbol{\fe}_{\policy}^{\intercal}} \) \\
\(Q(\policy) \sim \text{Cat}(\policyparams)\) \\
\textcolor{Peach}{3. Action Selection Phase}: \\
\(\action_{t} = \decision(\policyspace, t) \)
}

\textcolor{Peach}{4. Learning Phase}: \\
\If{\(t \mod T = 0\)}{\For{\(i \in [1, \dotsc, m]\)}{

\(\boldsymbol{\alpha}^{Q}_{:,i} \coloneq \boldsymbol{\alpha}^{P}_{:,i} + \sum_{t=1}^{\ntime} \ohobs_{t} \odot \stateparams_{\tau}[i]\) \\
\(\boldsymbol{\beta}^{Q}_{:,i} \coloneq \boldsymbol{\beta}^{P}_{:,i} + \sum_{t=2}^{\ntime} \sum_{\pi_{k} \in \Pi} \delta_{\action, c^{\policy}_{t}} \varprob{\policy} \bigl( \stateparams_{t}^{\policy} \odot \stateparams_{t-1}^{\policy}[i] \bigr)\)
}
\textcolor{Peach}{5. Reset (before a new episode starts)}: \\
Reset \(Q(\statevar_{1}|\policy_{k}), \dots, Q(\statevar_{\ntime}|\policy_{k})\), \(\forall k \in [1, \numpolicies]\) to uniform probability distributions.
}
\end{pscode}

\begin{pscode}[label={pscode:aif-action-aware}]{Action-aware Active Inference}
\DontPrintSemicolon%
\SetKwInput{KwHyperp}{Hyperparameters:}
\KwHyperp{\(\ntime\), epidode length, \(N\), number of episodes, \(\ntime^{\text{max}} = \ntime \times N\), max number of steps, \(\numpolicies = \card{\policyspace}\), number of policies.}
\KwData{sensory observations, \(\obs_{1}, \dots, \obs_{t}\), and policies, \(\policy_{1} \dots \policy_{\numpolicies}\).}
\KwResult{updated \(Q(\statevar_{t}|\action_{t-1})\), \(\forall t \in [1, \dots, \tau]\) and \(Q(\statevar_{\tau+1}|\policy_{k}), \dots, Q(\statevar_{\ntime}|\policy_{\numpolicies})\), \(\forall k \in [1, \dots, \numpolicies]\).}
\KwResult{updated \(Q(\policy)\), and next action, \(\action_{\tau}\).}
\KwResult{updated \(Q(\obsmap)\) and \(Q(\transmap)\)}
\BlankLine%
\For{\(t \in [1, \dots, \ntime^{\text{max}}]\)}{
\textcolor{Peach}{1. Perceptual Phase}: \\
\textcolor{ForestGreen}{a. Update probabilities over states}: \\
\For{\(t \in [1, \dots , \ntime]\)}{
    \uIf{\(t < \tau\)}{\(\nabla_{\stateparams_{t}} \fe_{(\action_{1:\tau - 1})} = 0 \Rightarrow \stateparams_{t} \coloneq \softmax{\ln \stateparams_{t}}\), see~\cref{eq:fe-update-states}}
    \Else{\For{\(k \in [1, \dots, \numpolicies]\)}{
        \(\nabla_{\stateparams_{t}} \fe_{\policy_{k}} = 0 \Rightarrow \stateparams_{t} \coloneq \softmax{\ln \stateparams_{t}}\), see~\cref{eq:fe-update-states}}}
  }
\textcolor{Peach}{2. Planning Phase}: \\
\For{\(k \in [1, \dots, \numpolicies]\)}{
    \textcolor{ForestGreen}{a. Compute total expected free energy}: \\
    \(\totefe(\pi_{k}) = \sum_{t = \tau + 1}^{T} \efe_{t}(\pi_{k})\) \\
    }
    \textcolor{ForestGreen}{b. Update probabilities over policies}: \\
\(\policyparams = \softmax{-\boldsymbol{\totefe}^{\intercal} - \boldsymbol{\fe}_{(\action_{1:\tau - 1})}^{\intercal}} \) \\
\(Q(\policy) \sim \text{Cat}(\policyparams)\) \\
\textcolor{Peach}{3. Action Selection Phase}: \\
\(\action_{t} = \decision(\policyspace, t) \)
}
\textcolor{Peach}{4. Learning Phase}: \\
\If{\(t \mod T = 0\)}{
\For{\(i \in [1, \dotsc, m]\)}{

\(\boldsymbol{\alpha}^{Q}_{:,i} \coloneq \boldsymbol{\alpha}^{P}_{:,i} + \sum_{t=1}^{\ntime} \ohobs_{t} \odot \stateparams_{\tau}[i]\) \\
\(\boldsymbol{\beta}^{Q}_{:,i} \coloneq \boldsymbol{\beta}^{P}_{:,i} + \sum_{t=2}^{\ntime} \sum_{\pi_{k} \in \Pi} \delta_{\action, c^{\policy}_{t}} \varprob{\policy} \bigl( \stateparams_{t}^{\policy} \odot \stateparams_{t-1}^{\policy}[i] \bigr)\)
}
\textcolor{Peach}{5. Reset (before a new episode starts)}: \\
Reset \(Q(\statevar_{1}|\policy_{k}), \dots, Q(\statevar_{\ntime}|\policy_{k})\), \(\forall k \in [1, \numpolicies]\) to uniform probability distributions.
}
\end{pscode}

\section{Further Information on Experiments}\label{secx:further-info-exps}

\subsection{Experiment 1: 4-step T-maze}\label{ssecx:exp-tmaze4}
\subsubsection{How to Reproduce the Results of the Experiment}\label{sssecx:how-to-reproduce-tmaze4}
The results reported in \cref{sec:exp1-Tmaze4} were obtained by using the following command line arguments.

For the action-unaware agent:

\begin{lstlisting}[breaklines=true,]
main_aif_paths --exp_name aif_paths --gym_id gridworld-v1 --env_layout tmaze4 --num_runs 10 --num_episodes 100 --num_steps 4 --inf_steps 10 --action_selection kd -lB --num_policies 64 --pref_loc all_goal
\end{lstlisting}

For the action-aware agent:

\begin{lstlisting}[breaklines=true,]
main_aif_plans_pi_cutoff --exp_name aif_plans --gym_id gridworld-v1 --env_layout tmaze4 --num_runs 10 --num_episodes 100 --num_steps 4 --inf_steps 10 --action_selection kd -lB --num_policies 64
\end{lstlisting}

The plots were obtained using the following command line instructions:

\begin{lstlisting}[breaklines=true]
vis_aif -gid gridworld-v1 -el tmaze4 -nexp 2 -rdir episodic_e100_pol16_maxinf10_learnB -fpi 0 1 2 3 -i 4 -v 8 -ti 4 -tv 8 -vl 3 -hl 3 -xtes 20 -ph 3 -selrun 0 -selep 24 49 74 99 -npv 16 -sb 4 -ab 0 1 2 3
\end{lstlisting}

With these instructions, one can visualise more metrics than those reported in the main text. We offer a selection next.

\subsubsection{Free energy at steps 1-3}\label{sssecx:fes-tmaze4}

\begin{figure}[H]
\centering
\begin{subfigure}{0.45\textwidth}
    \centering
    \includegraphics[width=\textwidth]{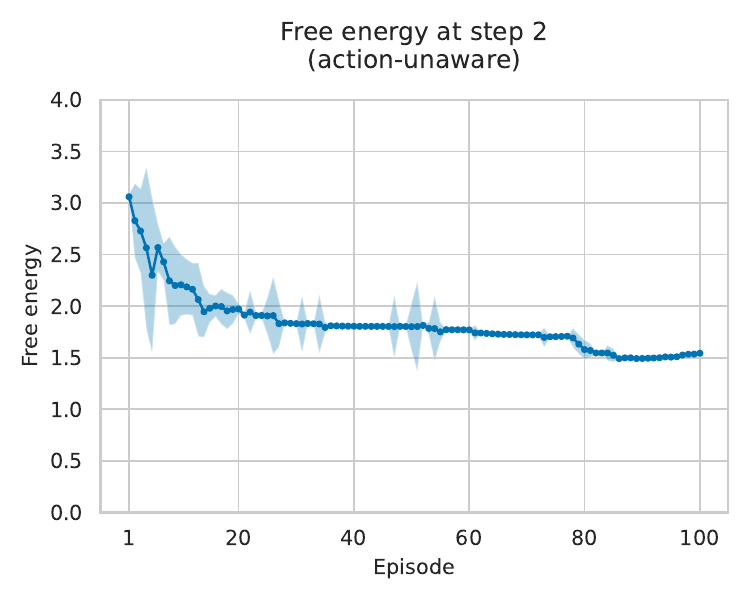}
    \caption{Free energy.}\label{fig:tmaze4-aif-paths-marginal-fe-step2}
  \end{subfigure}
\begin{subfigure}{0.45\textwidth}
    \centering
    \includegraphics[width=\textwidth]{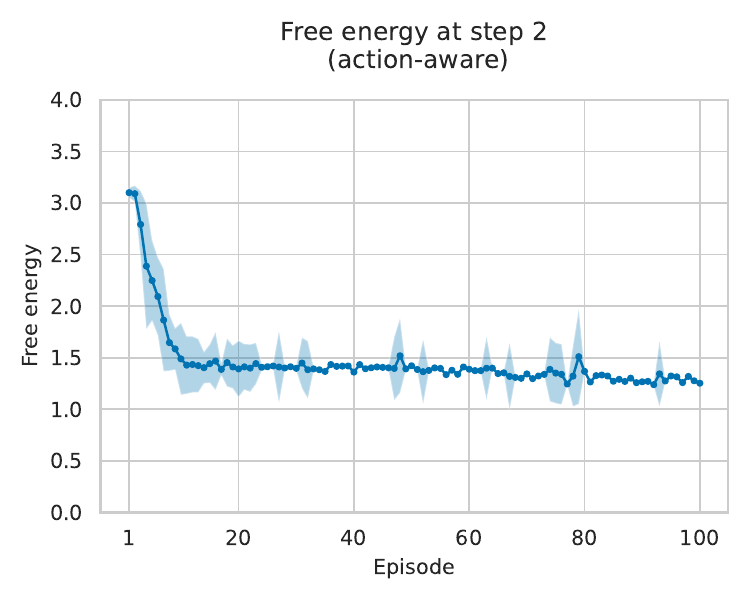}
    \caption{Free energy.}\label{fig:tmaze4-aif-plans-marginal-fe-step2}
  \end{subfigure}
\caption{Free energy at step 2 across episodes (showing average over 10 agents).}\label{fig:tmaze4-marginal-fe-step2}
\end{figure}

\begin{figure}[H]
\centering
\begin{subfigure}{0.45\textwidth}
    \centering
    \includegraphics[width=\textwidth]{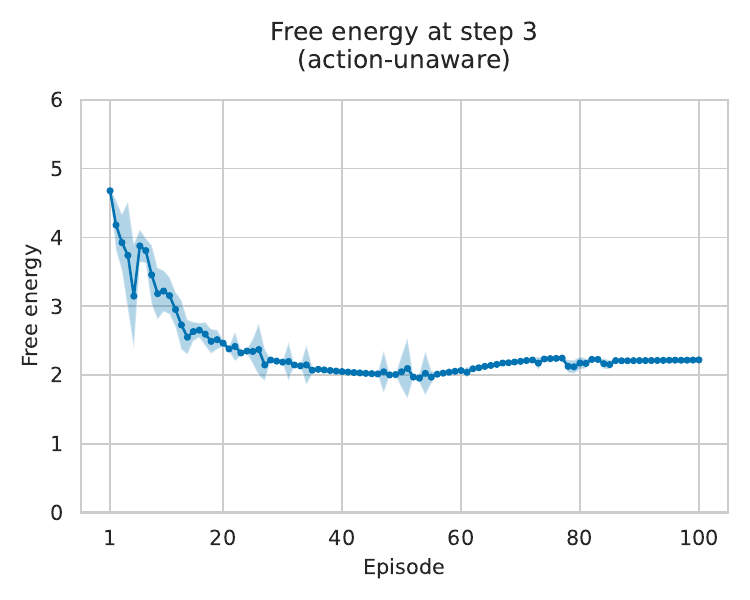}
    \caption{Free energy.}\label{fig:tmaze4-aif-paths-marginal-fe-step3}
  \end{subfigure}
\begin{subfigure}{0.45\textwidth}
    \centering
    \includegraphics[width=\textwidth]{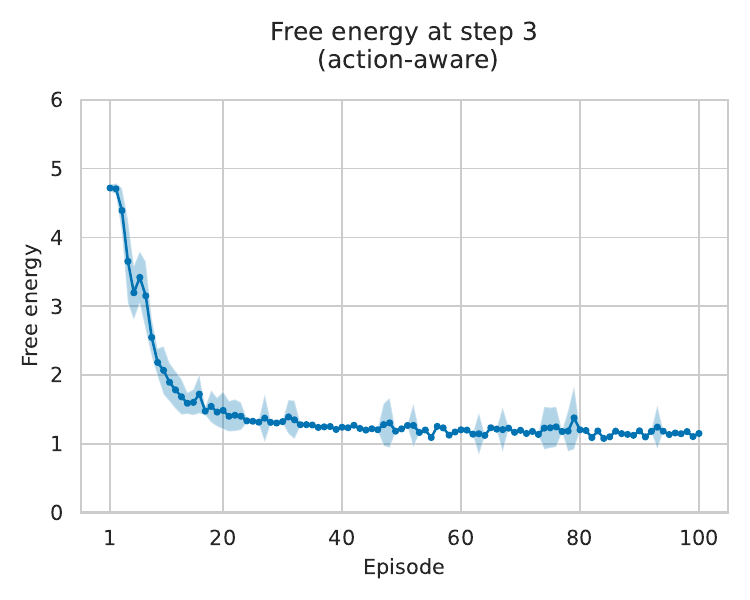}
    \caption{Free energy.}\label{fig:tmaze4-aif-plans-marginal-fe-step3}
  \end{subfigure}
\caption{Free energy at step 3 across episodes (showing average of 10 agents).}\label{fig:tmaze4-marginal-fe-step3}
\end{figure}

\subsubsection{Policy-conditioned free energy at steps 1-3}\label{sssecx:tmaze4-pc-fe}

\begin{figure}[H]
  \centering
  \begin{subfigure}{0.45\textwidth}\label{fig:tmaze4-aif-paths-policies-fe-step1}
    \centering
    \includegraphics[width=\textwidth]{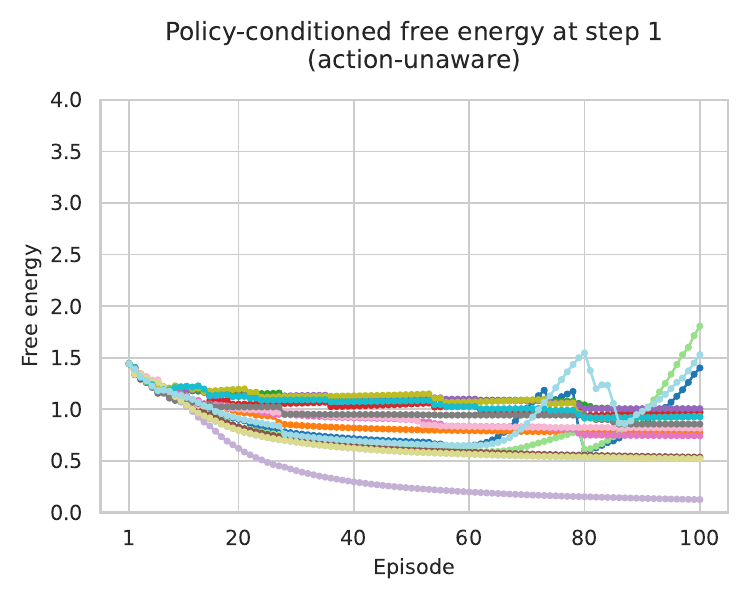}
  \end{subfigure}
  \begin{subfigure}{0.45\textwidth}\label{fig:tmaze4-aif-plans-policies-fe-step1}
    \centering
    \includegraphics[width=\textwidth]{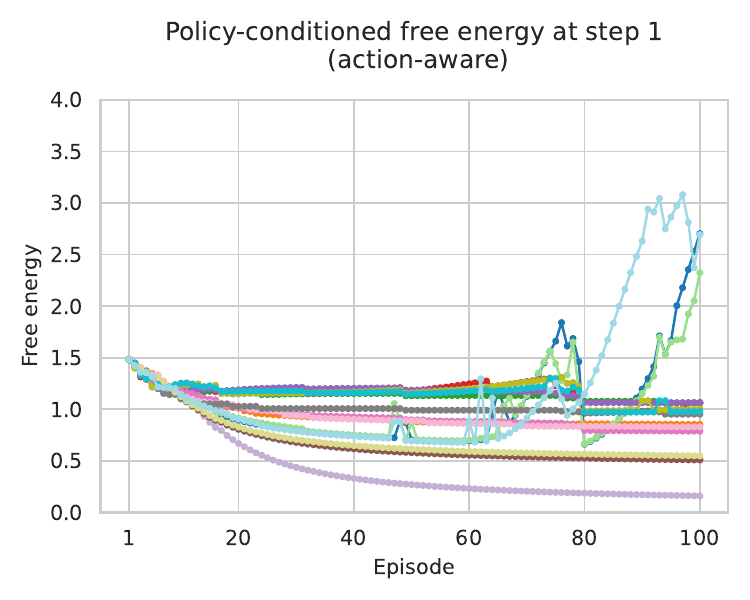}
  \end{subfigure}
  \begin{subfigure}{0.65\textwidth}
    \centering
    \includegraphics[width=\textwidth]{tmaze4_aif_policies_legend}
  \end{subfigure}
  \caption{Policy-conditioned free energies at step 1 across episodes (showing average of 10 agents).}\label{fig:tmaze4-policies-fes-step1}
\end{figure}

\begin{figure}[H]
  \centering
  \begin{subfigure}{0.45\textwidth}\label{fig:tmaze4-aif-paths-policies-fe-step2}
    \centering
    \includegraphics[width=\textwidth]{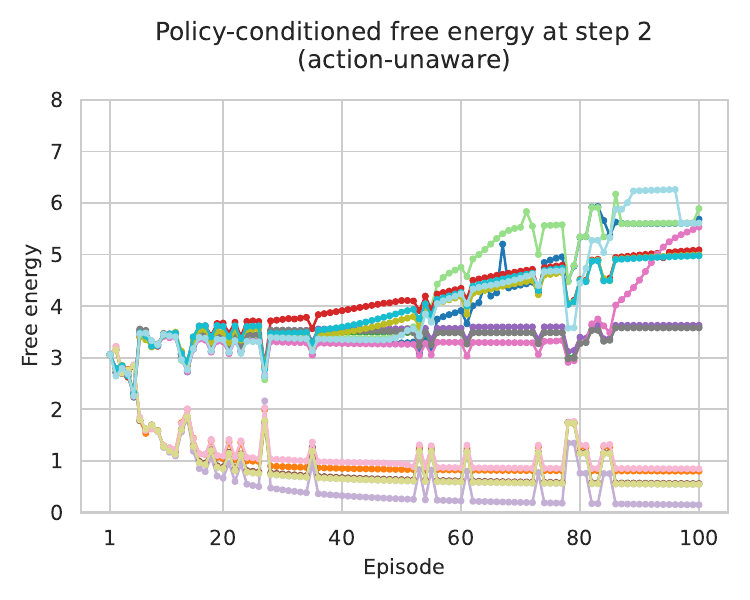}
  \end{subfigure}
  \begin{subfigure}{0.45\textwidth}\label{fig:tmaze4-aif-plans-policies-fe-step2}
    \centering
    \includegraphics[width=\textwidth]{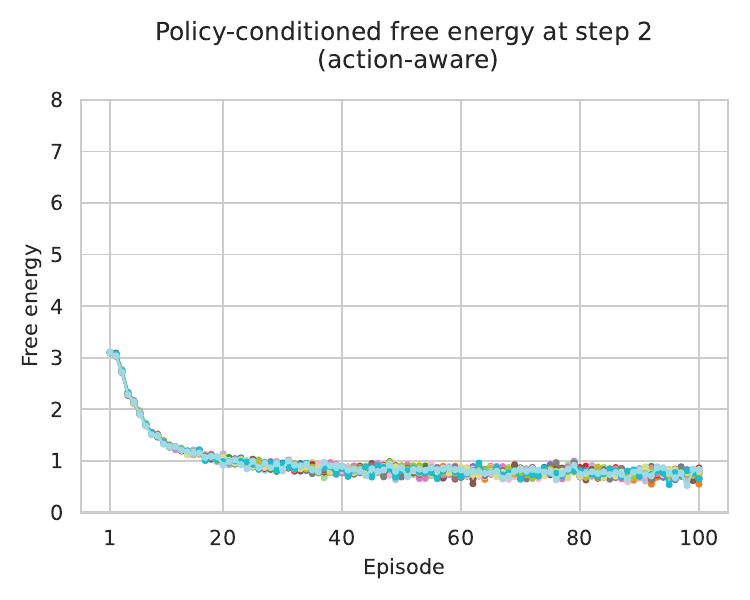}
  \end{subfigure}
    \begin{subfigure}{0.65\textwidth}
    \centering
    \includegraphics[width=\textwidth]{tmaze4_aif_policies_legend}
  \end{subfigure}
  \caption{Policy-conditioned free energies at step 2 across episodes (showing average of 10 agents).}\label{fig:tmaze4-policies-fes-step2}
\end{figure}

\begin{figure}[H]
  \centering
  \begin{subfigure}{0.45\textwidth}\label{fig:tmaze4-aif-paths-policies-fe-step3}
    \centering
    \includegraphics[width=\textwidth]{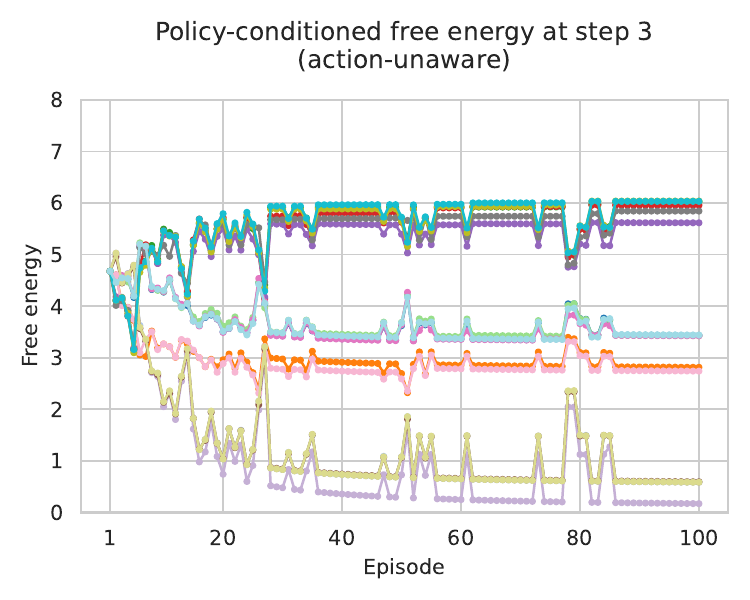}
  \end{subfigure}
  \begin{subfigure}{0.45\textwidth}\label{fig:tmaze4-aif-plans-policies-fe-step3}
    \centering
    \includegraphics[width=\textwidth]{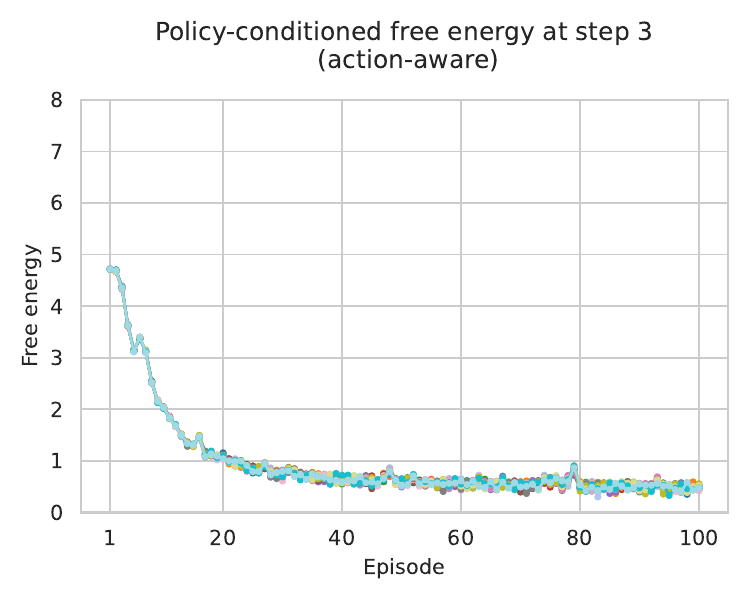}
  \end{subfigure}
    \begin{subfigure}{0.65\textwidth}
    \centering
    \includegraphics[width=\textwidth]{tmaze4_aif_policies_legend}
  \end{subfigure}
  \caption{Policy-conditioned free energies at step 3 across episodes (showing average of 10 agents).}\label{fig:tmaze4-policies-fes-step3}
\end{figure}

\subsubsection{Expected free energy at steps 2--3}\label{sssecx:tmaze4-efe-steps}

\begin{figure}[H]
  \centering
  \begin{subfigure}{0.45\textwidth}\label{fig:tmaze4-aif-paths-efe-step1}
    \centering
    \includegraphics[width=\textwidth]{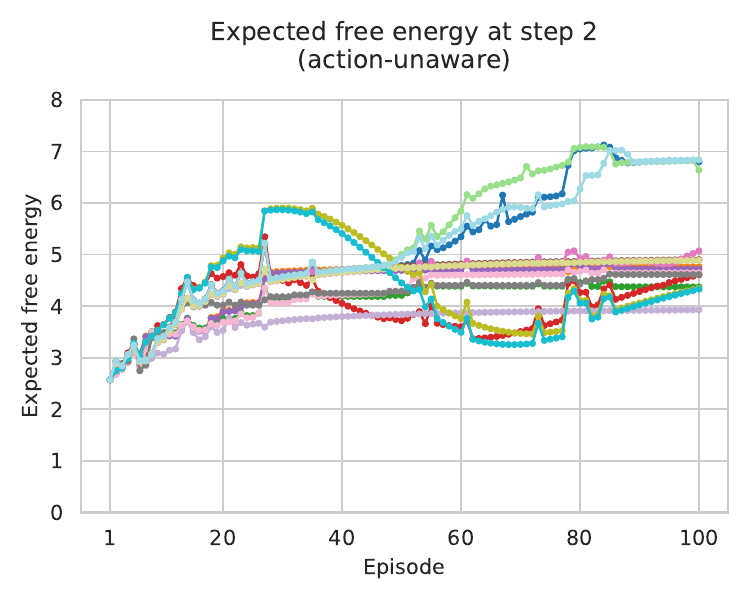}
  \end{subfigure}
  \begin{subfigure}{0.45\textwidth}\label{fig:tmaze4-aif-plans-efe-step1}
    \centering
    \includegraphics[width=\textwidth]{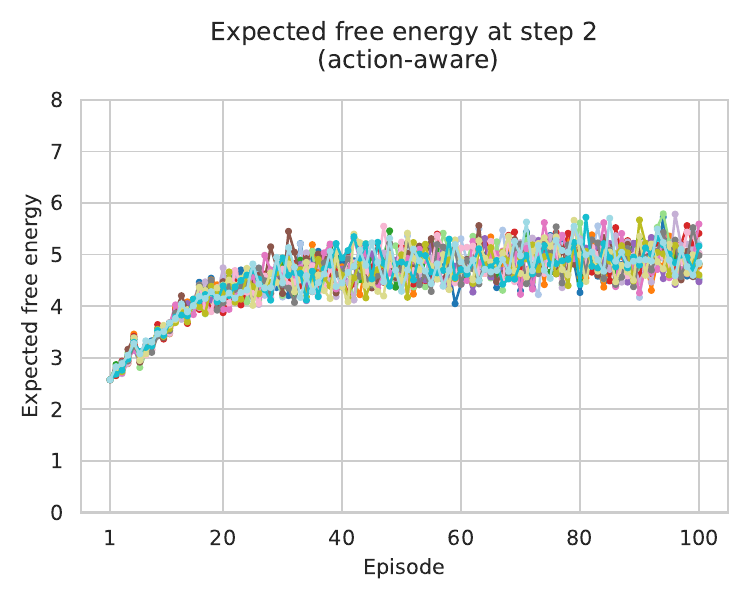}
  \end{subfigure}
    \begin{subfigure}{0.65\textwidth}
    \centering
    \includegraphics[width=\textwidth]{tmaze4_aif_policies_legend}
  \end{subfigure}
  \caption{Expected free energy at step 2 for each policy across episodes (showing average of 10 agents).}\label{fig:tmaze4-efe-step2}
\end{figure}

\begin{figure}[H]
  \centering
  \begin{subfigure}{0.45\textwidth}\label{fig:tmaze4-aif-paths-efe-step2}
    \centering
    \includegraphics[width=\textwidth]{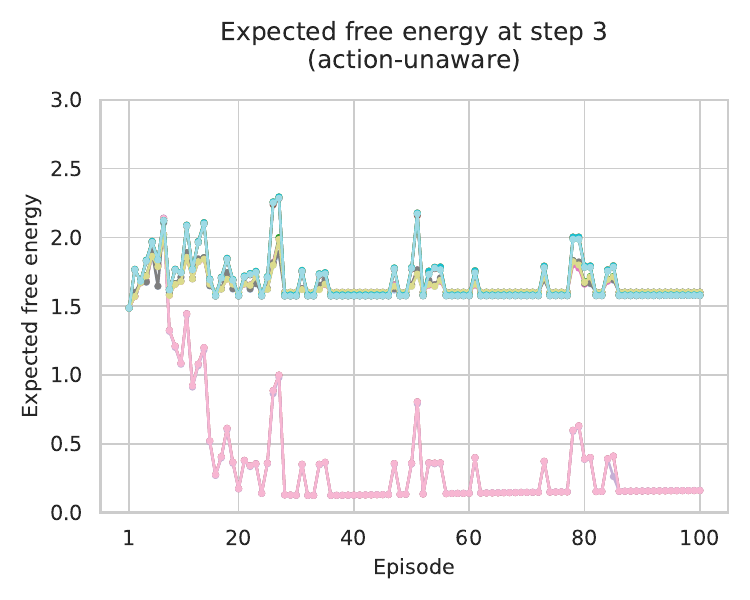}
  \end{subfigure}
  \begin{subfigure}{0.45\textwidth}\label{fig:tmaze4-aif-plans-efe-step2}
    \centering
    \includegraphics[width=\textwidth]{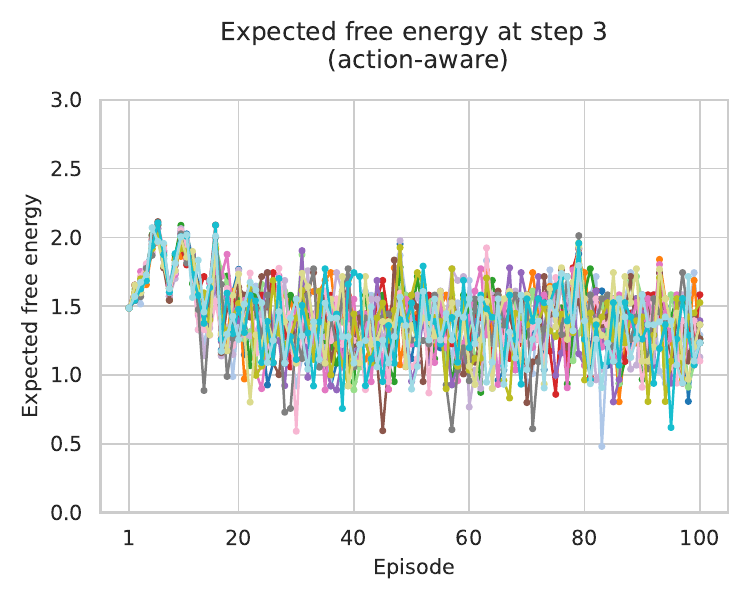}
  \end{subfigure}
    \begin{subfigure}{0.65\textwidth}
    \centering
    \includegraphics[width=\textwidth]{tmaze4_aif_policies_legend}
  \end{subfigure}
  \caption{Expected free energy at step 3 for each policy across episodes (showing average of 10 agents).}\label{fig:tmaze4-efe-step3}
\end{figure}

There is no expected free energy at step 4 because this is the step at which the environment terminates in the episodic setting considered in this work and, regardless of its location, the agent is no longer given the ability to plan forward in time.

\subsubsection{Expected free energy at step 1 breakdown}\label{sssecx:tmaze4-efe-breakdown}

\begin{figure}[H]
  \centering
  \begin{subfigure}{0.45\textwidth}
    \centering
    \includegraphics[width=\textwidth]{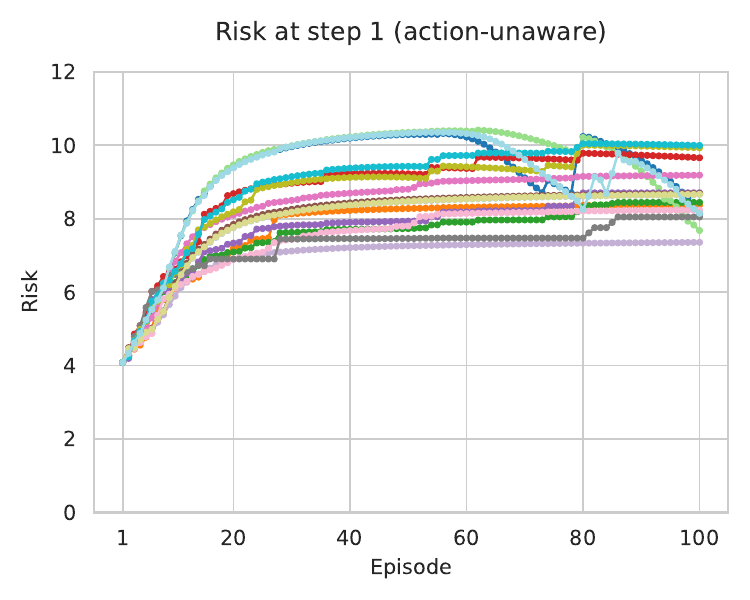}\label{fig:tmaze4-aif-paths-efe-risk}
  \end{subfigure}
  \begin{subfigure}{0.45\textwidth}
    \centering
    \includegraphics[width=\textwidth]{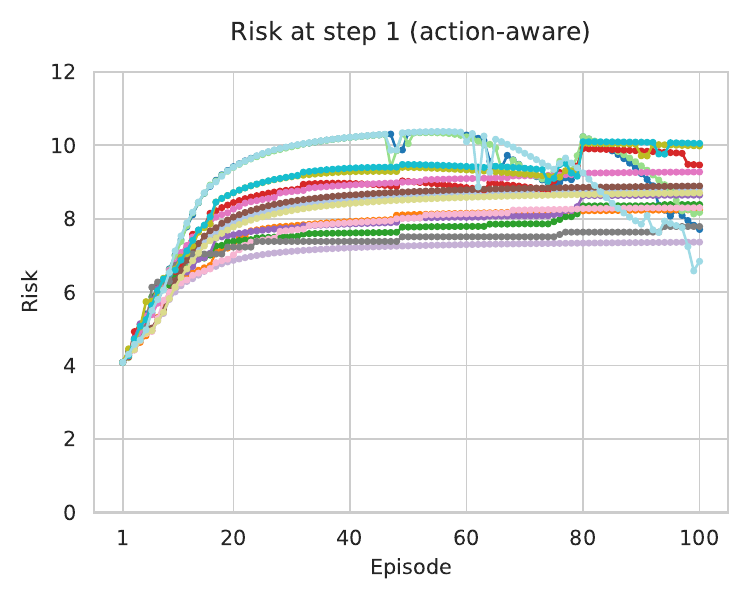}\label{fig:tmaze4-aif-plans-efe-risk}
  \end{subfigure}
    \begin{subfigure}{0.65\textwidth}
    \centering
    \includegraphics[width=\textwidth]{tmaze4_aif_policies_legend}
  \end{subfigure}
  \caption{Risk (expected free energy term) for each policy across episodes (showing average of 10 agents).}\label{fig:tmaze4-efe-risk}
\end{figure}

\begin{figure}[H]
  \centering
  \begin{subfigure}{0.45\textwidth}
    \centering
    \includegraphics[width=\textwidth]{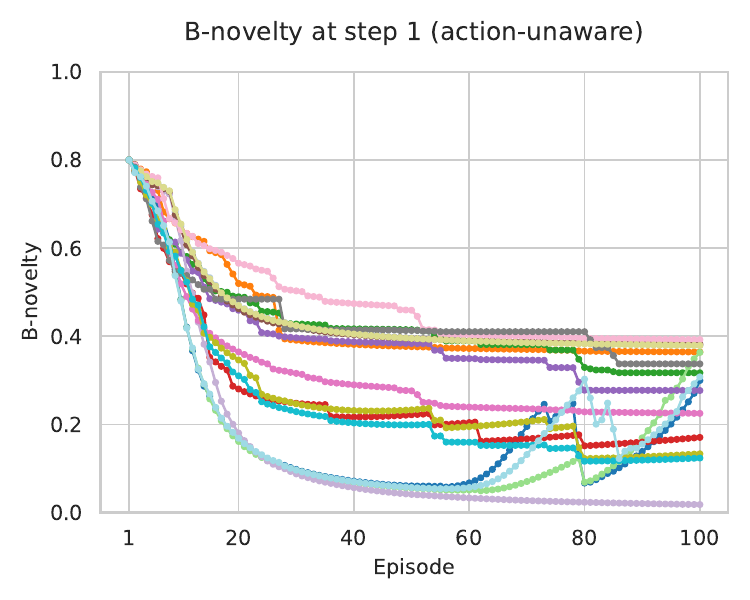}\label{fig:tmaze4-aif-paths-efe-bnov}
  \end{subfigure}
  \begin{subfigure}{0.45\textwidth}
    \centering
    \includegraphics[width=\textwidth]{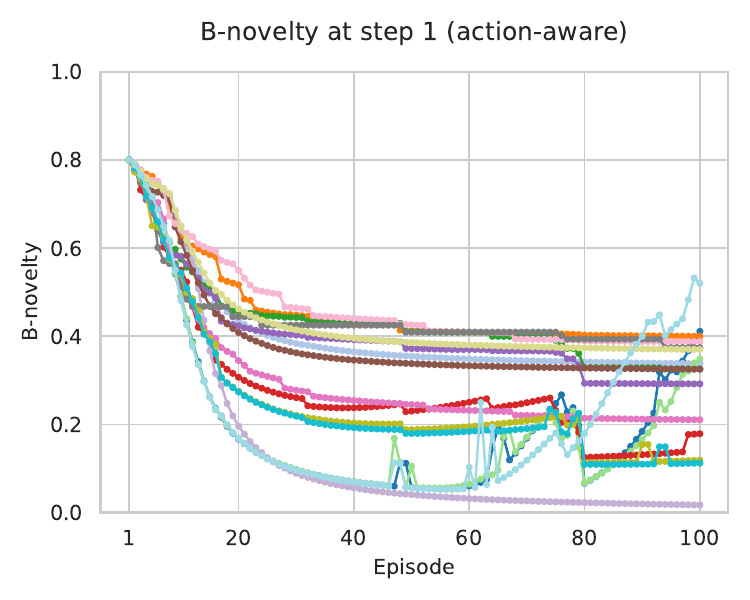}\label{fig:tmaze4-aif-plans-efe-bnov}
  \end{subfigure}
    \begin{subfigure}{0.65\textwidth}
    \centering
    \includegraphics[width=\textwidth]{tmaze4_aif_policies_legend}
  \end{subfigure}
  \caption{\(\transmap\)-novelty (expected free energy term) for each policy across episodes (showing average of 10 agents).}\label{fig:tmaze4-efe-bnov}
\end{figure}

\subsubsection{Ground truth transition maps}\label{sssecx:tmaze4-gtruth-trans-maps}

\begin{figure}[H]
\centering
\begin{subfigure}{0.45\textwidth}
    \centering
    \includegraphics[width=\textwidth]{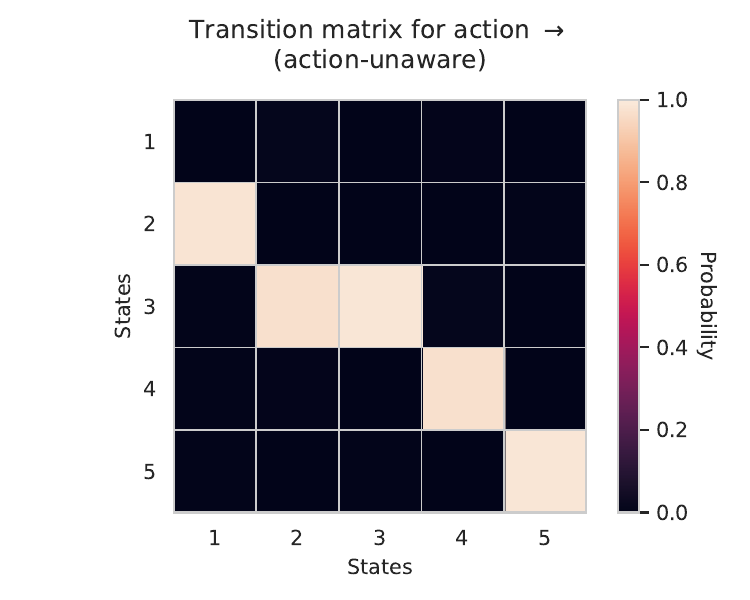}\label{fig:tmaze4-gtruth-matrix-B-a0}
\end{subfigure}
\begin{subfigure}{0.45\textwidth}
    \centering
    \includegraphics[width=\textwidth]{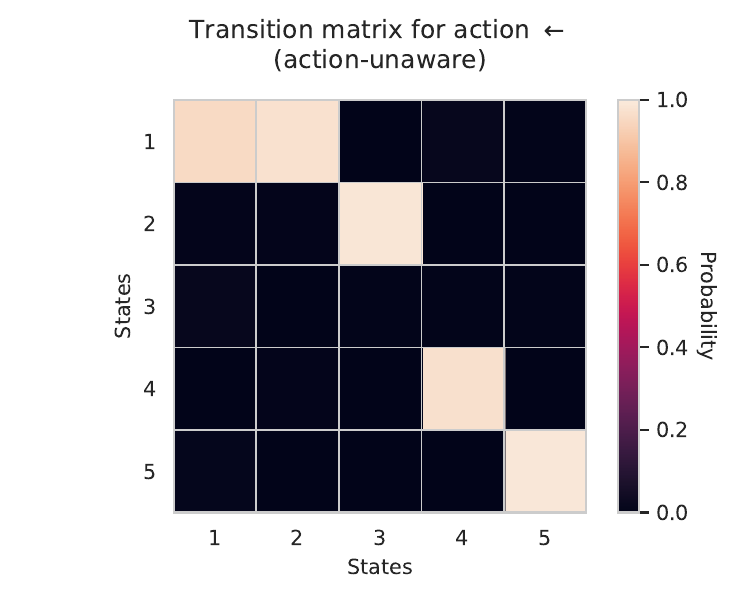}\label{fig:tmaze4-gtruth-matrix-B-a2}
\end{subfigure}
\caption{Ground truth transition maps for action \(\rightarrow\) and \(\leftarrow\) in the T-maze.}\label{fig:tmaze4-gt-arl}
\end{figure}

\begin{figure}[H]
\centering
\begin{subfigure}{0.45\textwidth}
    \centering
    \includegraphics[width=\textwidth]{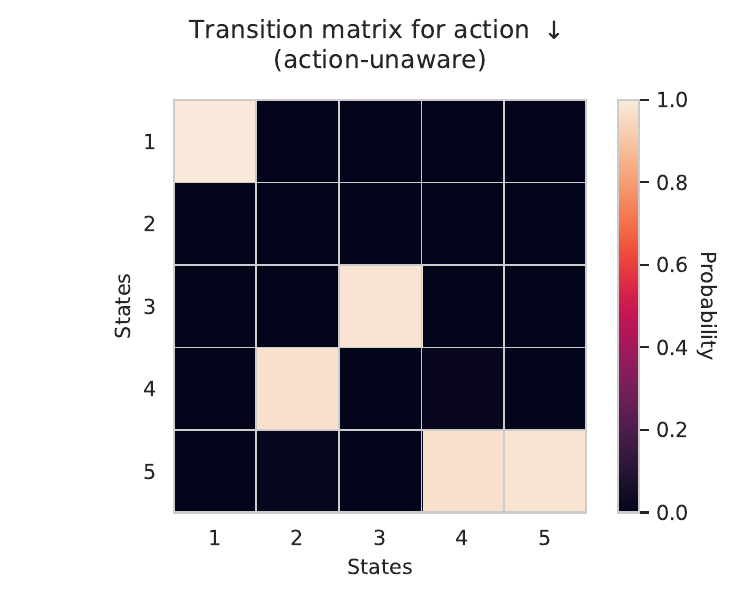}\label{fig:tmaze4-gtruth-matrix-B-a1}
\end{subfigure}
\begin{subfigure}{0.45\textwidth}
    \centering
    \includegraphics[width=\textwidth]{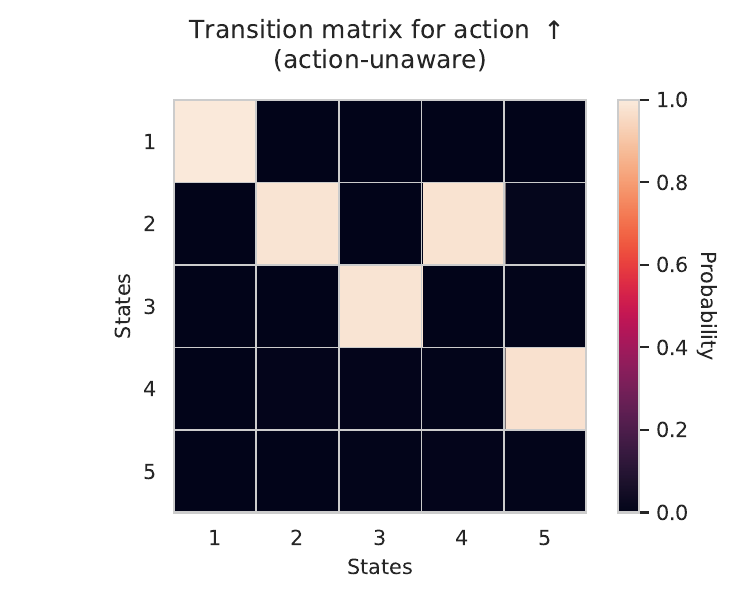}\label{fig:tmaze4-gtruth-matrix-B-a3}
\end{subfigure}
\caption{Ground truth ransition maps for action \(\downarrow\) and \(\uparrow\) in the T-maze.}\label{fig:tmaze4-a13}
\end{figure}

\subsubsection{Learned transition maps in action-unaware and action-aware agents}\label{sssecx:tmaze4-learned-trans-maps}

\begin{figure}[H]
\centering
\begin{subfigure}{0.45\textwidth}
    \centering
    \includegraphics[width=\textwidth]{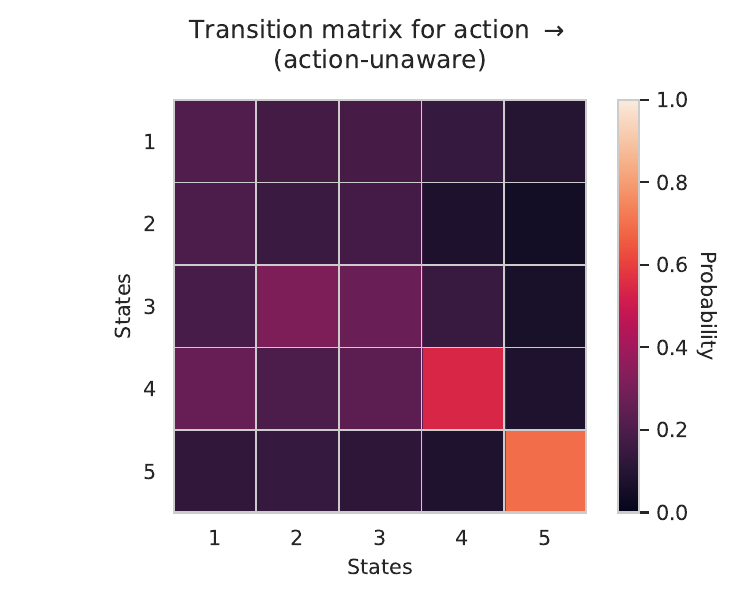}\label{fig:tmaze4-aif-paths-matrix-B-a0}
\end{subfigure}
\begin{subfigure}{0.45\textwidth}
    \centering
    \includegraphics[width=\textwidth]{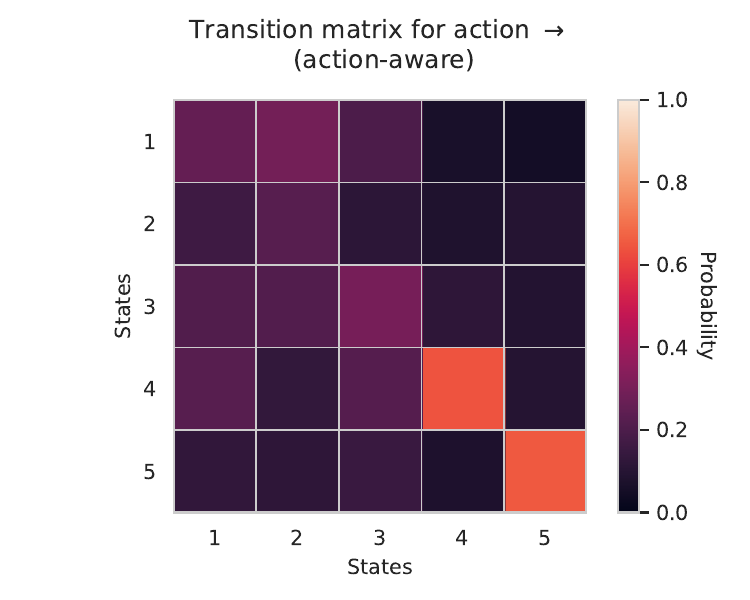}\label{fig:tmaze4-aif-plans-matrix-B-a0}
\end{subfigure}
\caption{Transition maps for action \(\rightarrow\).}\label{fig:tmaze4-a0r}
\end{figure}

\begin{figure}[H]
\centering
\begin{subfigure}{0.45\textwidth}
    \centering
    \includegraphics[width=\textwidth]{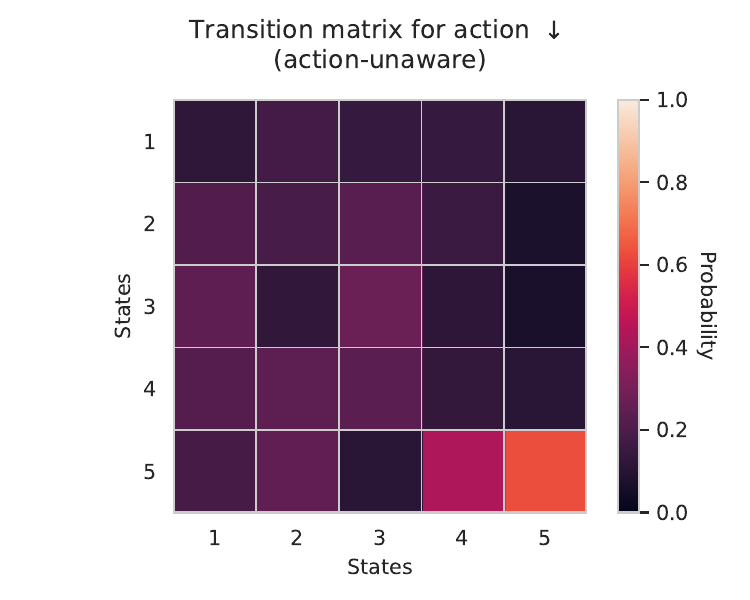}\label{fig:tmaze4-aif-paths-matrix-B-a1}
\end{subfigure}
\begin{subfigure}{0.45\textwidth}
    \centering
    \includegraphics[width=\textwidth]{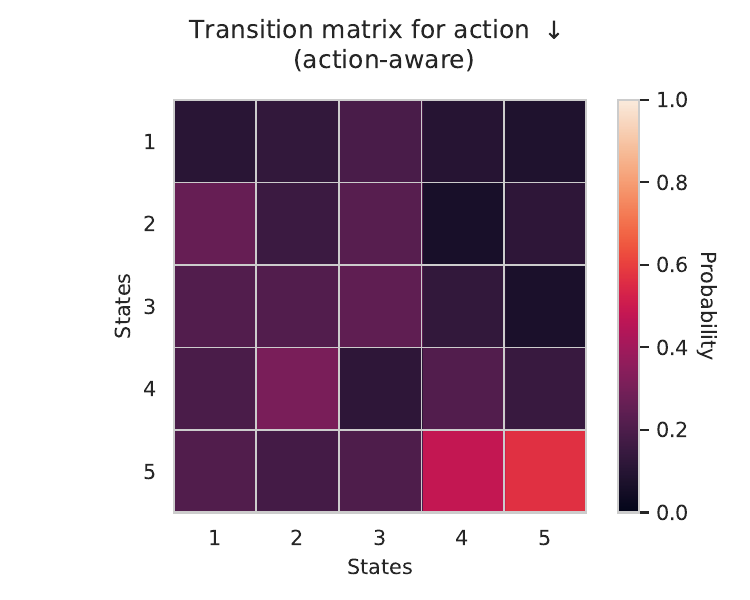}\label{fig:tmaze4-aif-plans-matrix-B-a1}
\end{subfigure}
\caption{Transition maps for action \(\downarrow\).}\label{fig:tmaze4-a1u}
\end{figure}

\begin{figure}[H]
\centering
\begin{subfigure}{0.45\textwidth}
    \centering
    \includegraphics[width=\textwidth]{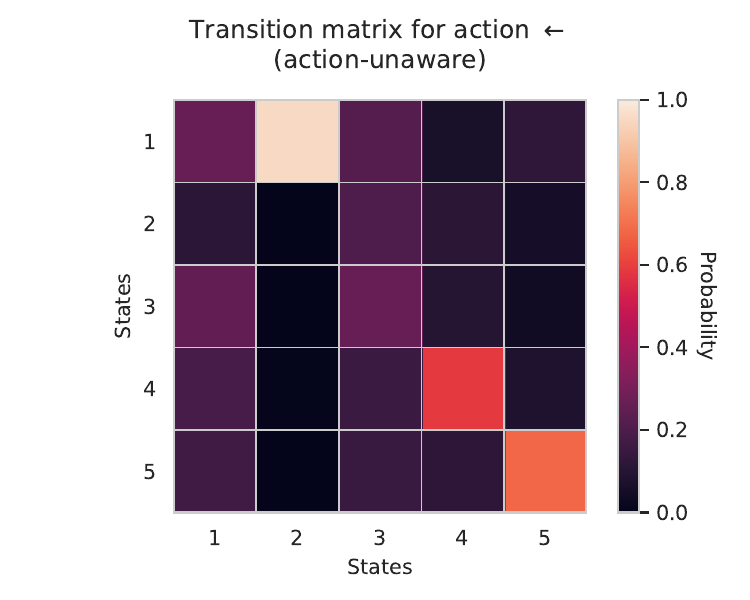}\label{fig:tmaze4-aif-paths-matrix-B-a2}
\end{subfigure}
\begin{subfigure}{0.45\textwidth}
    \centering
    \includegraphics[width=\textwidth]{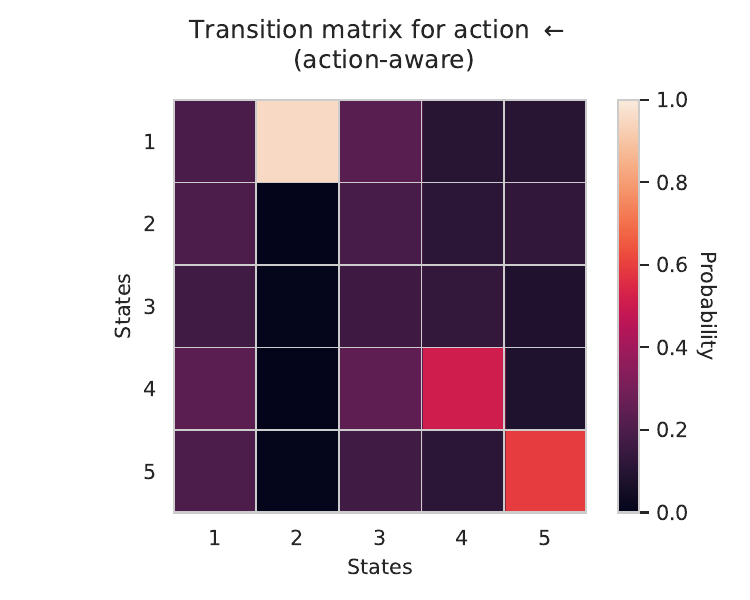}\label{fig:tmaze4a-aif-plans-matrix-B-a2}
\end{subfigure}
\caption{Transition maps for action \(\leftarrow\).}\label{fig:tmaze4-a2l}
\end{figure}

\begin{figure}[H]
\centering
\begin{subfigure}{0.45\textwidth}
    \centering
    \includegraphics[width=\textwidth]{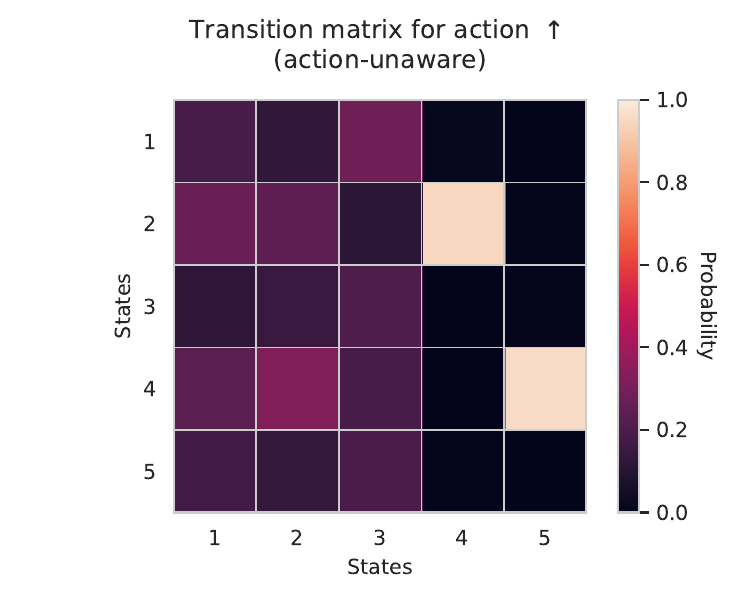}\label{fig:tmaze4-transmap-path-a3}
\end{subfigure}
\begin{subfigure}{0.45\textwidth}
    \centering
    \includegraphics[width=\textwidth]{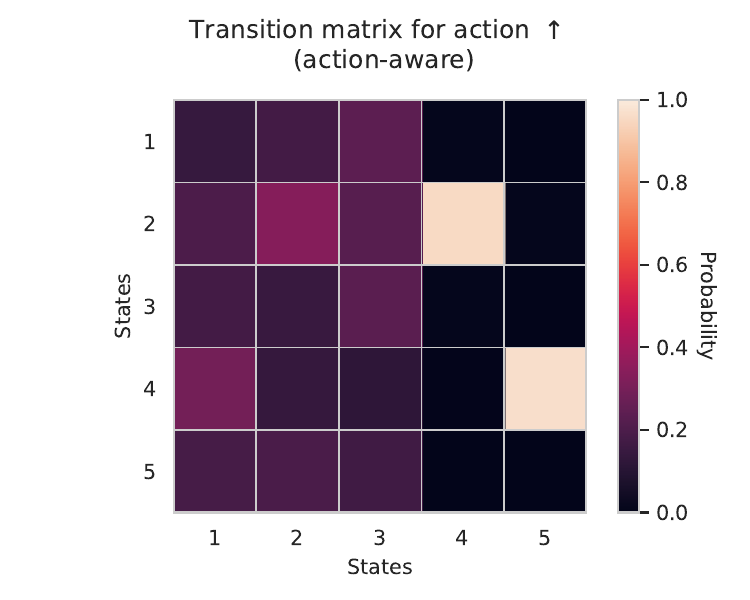}\label{fig:tmaze4-aif-plans-a3}
\end{subfigure}
\caption{Transition maps for action \(\uparrow\).}\label{fig:tmaze4-a3d}
\end{figure}

\subsection{Experiment 2: 5-step Gridw9}\label{ssecx:exp2-gridw9}
\subsubsection{How to Reproduce the Results of the Experiment}\label{sssecx:how-to-reproduce-gridw9}
The results reported in \cref{ssec:exp2-gridw9} were obtained by using the following command line arguments.

For the action-unaware agent:

\begin{lstlisting}[breaklines=true,]
main_aif_paths --exp_name aif_paths --gym_id gridworld-v1 --env_layout gridw9 --num_runs 10 --num_episodes 180 --num_steps 5 --inf_steps 10 --action_selection kd -lB --num_policies 256 --pref_loc all_goal
\end{lstlisting}

For the action-aware agent:

\begin{lstlisting}[breaklines=true,]
main_aif_plans_pi_cutoff --exp_name aif_plans --gym_id gridworld-v1 --env_layout gridw9 --num_runs 10 --num_episodes 180 --num_steps 5 --inf_steps 10 --action_selection kd -lB --num_policies 256
\end{lstlisting}

The plots were obtained using the following command line instructions:

\begin{lstlisting}[breaklines=true,]
vis_aif -gid gridworld-v1 -el gridw9 -nexp 2 -rdir episodic_e180_pol16_maxinf10_learnB -fpi 0 1 2 3 4 -i 4 -v 8 -ti 4 -tv 8 -vl 3 -hl 3 -xtes 20 -ph 4 -selrun 0 -selep 24 49 74 99 -npv 16 -sb 4 -ab 0 1 2 3
\end{lstlisting}

With these instructions, one can visualize more metrics than those reported in the main text. We offer a selection next.

\subsubsection{Free energy at steps 1-4}\label{sssecx:fes-gridw9}

\begin{figure}[H]
\centering
\begin{subfigure}{0.45\textwidth}
    \centering
    \includegraphics[width=\textwidth]{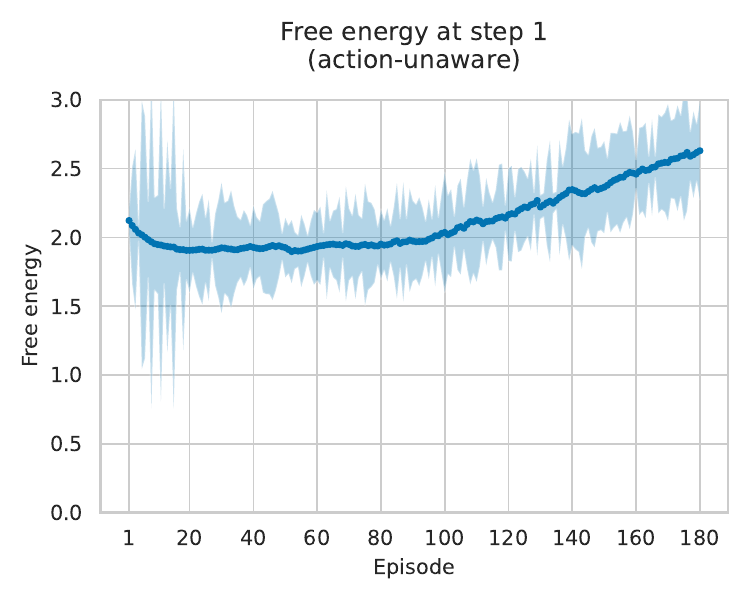}
    \caption{Free energy.}\label{fig:gridw9-aif-paths-marginal-fe-step1}
  \end{subfigure}
\begin{subfigure}{0.45\textwidth}
    \centering
    \includegraphics[width=\textwidth]{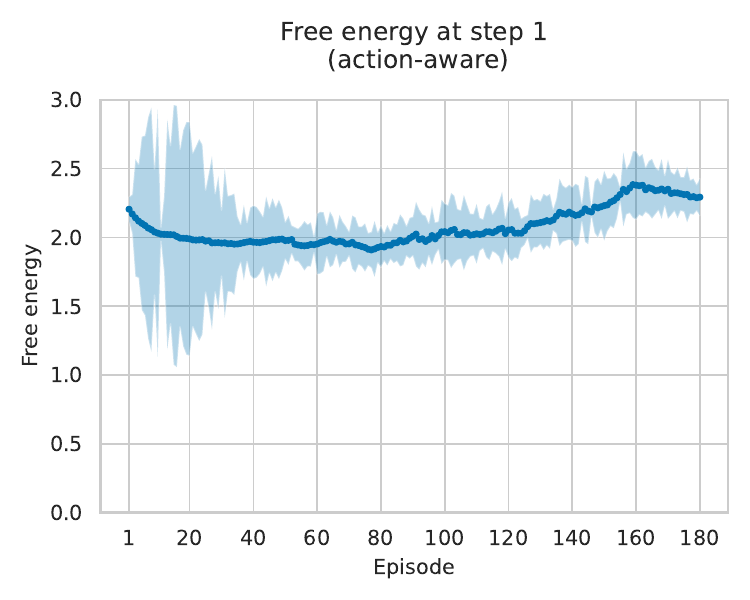}
    \caption{Free energy.}\label{fig:gridw9-aif-plans-marginal-fe-step1}
  \end{subfigure}
\caption{Free energy at step 1 across episodes (showing average of 10 agents).}\label{fig:gridw9-marginal-fe-step1}
\end{figure}

\begin{figure}[H]
\centering
\begin{subfigure}{0.45\textwidth}
    \centering
    \includegraphics[width=\textwidth]{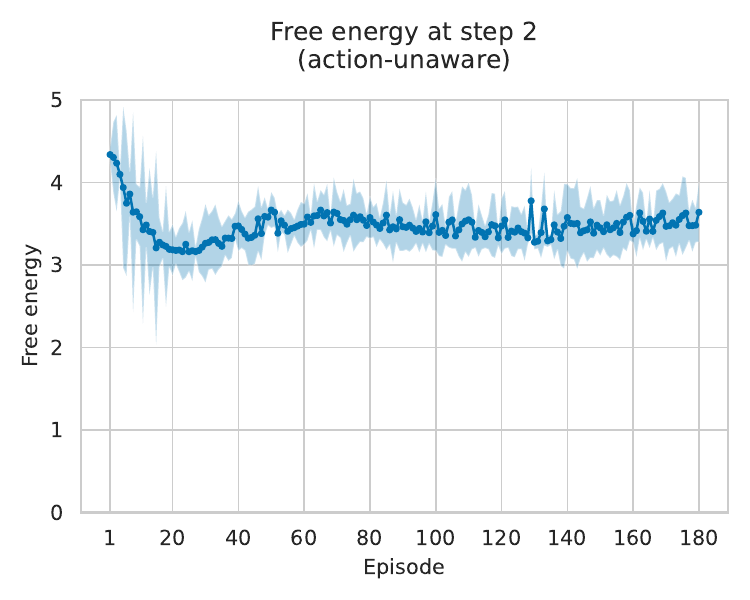}
    \caption{Free energy.}\label{fig:gridw9-aif-paths-marginal-fe-step2}
  \end{subfigure}
\begin{subfigure}{0.45\textwidth}
    \centering
    \includegraphics[width=\textwidth]{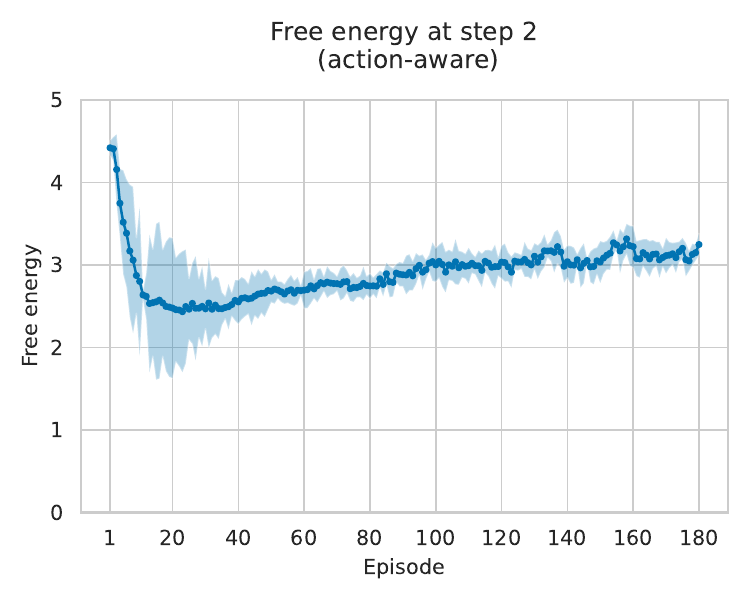}
    \caption{Free energy.}\label{fig:gridw9-aif-plans-marginal-fe-step2}
  \end{subfigure}
\caption{Free energy at step 2 across episodes (showing average of 10 agents).}\label{fig:gridw9-marginal-fe-step2}
\end{figure}

\begin{figure}[H]
\centering
\begin{subfigure}{0.45\textwidth}
    \centering
    \includegraphics[width=\textwidth]{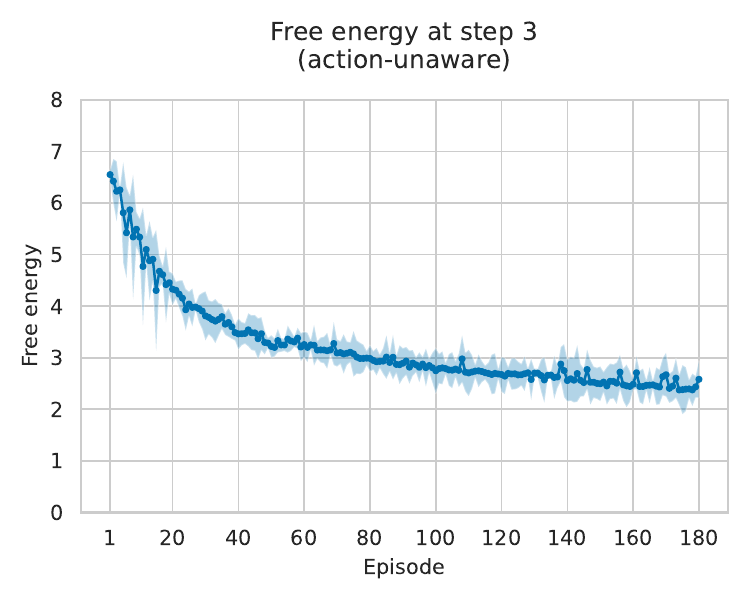}
    \caption{Free energy.}\label{fig:gridw9-aif-paths-marginal-fe-step3}
  \end{subfigure}
\begin{subfigure}{0.45\textwidth}
    \centering
    \includegraphics[width=\textwidth]{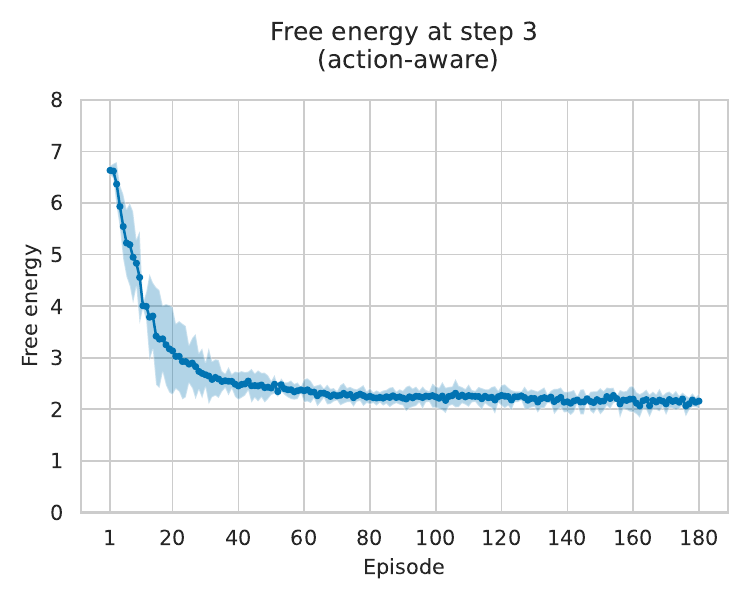}
    \caption{Free energy.}\label{fig:gridw9-aif-plans-marginal-fe-step3}
  \end{subfigure}
\caption{Free energy at step 3 across episodes (showing average of 10 agents).}\label{fig:gridw9-marginal-fe-step3}
\end{figure}

\begin{figure}[H]
\centering
\begin{subfigure}{0.45\textwidth}
    \centering
    \includegraphics[width=\textwidth]{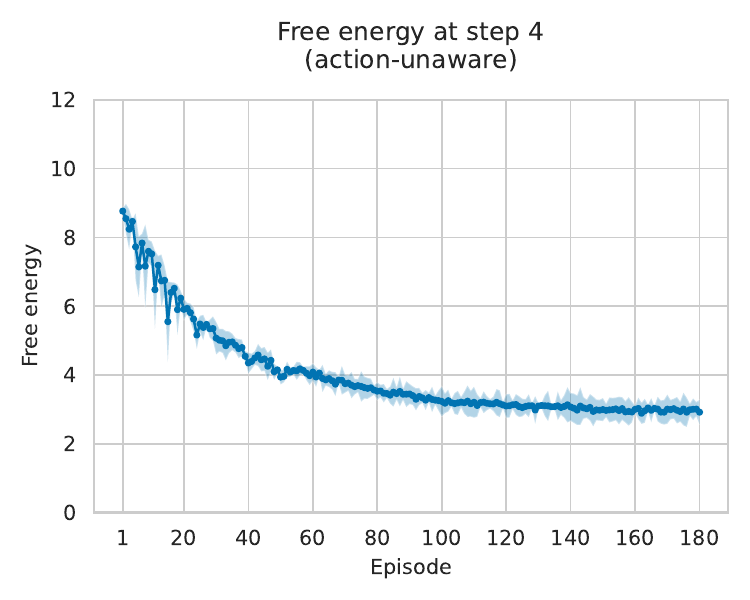}
    \caption{Free energy.}\label{fig:gridw9-aif-paths-marginal-fe-step4}
  \end{subfigure}
\begin{subfigure}{0.45\textwidth}
    \centering
    \includegraphics[width=\textwidth]{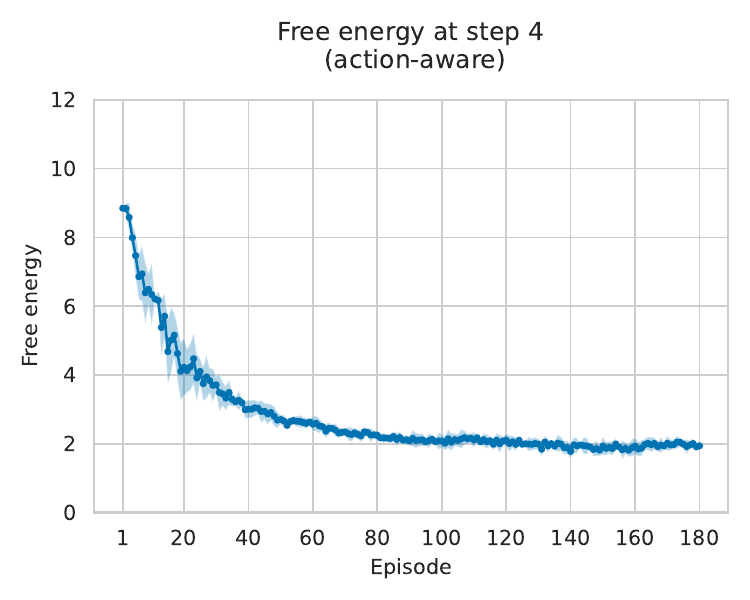}
    \caption{Free energy.}\label{fig:gridw9-aif-plans-marginal-fe-step4}
  \end{subfigure}
\caption{Free energy at step 4 across episodes (showing average of 10 agents).}\label{fig:gridw9-marginal-fe-step4}
\end{figure}

\subsubsection{Policy-conditioned free energy at steps 1--4}\label{sssecx:gridw9-pc-fe}

\begin{figure}[H]
  \centering
  \begin{subfigure}{0.45\textwidth}\label{fig:gridw9-aif-paths-policies-fe-step2}
    \centering
    \includegraphics[width=\textwidth]{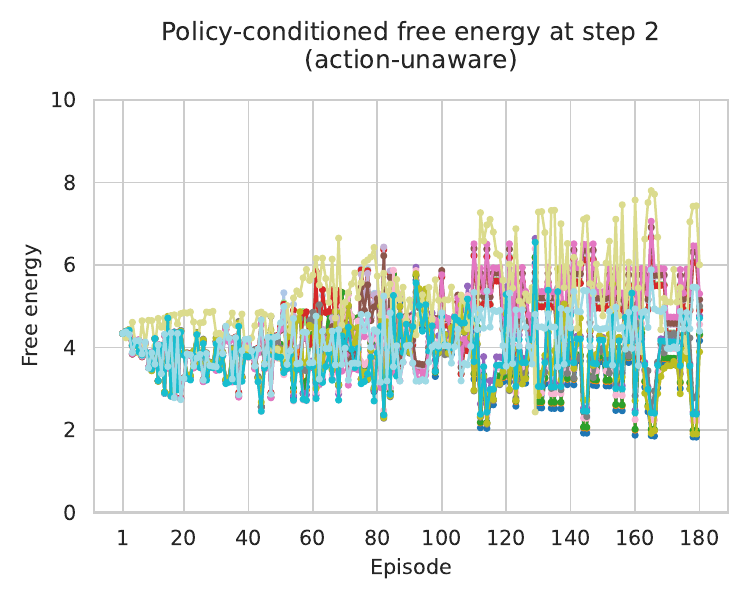}
  \end{subfigure}
  \begin{subfigure}{0.45\textwidth}\label{fig:gridw9-aif-plans-policies-fe-step2}
    \centering
    \includegraphics[width=\textwidth]{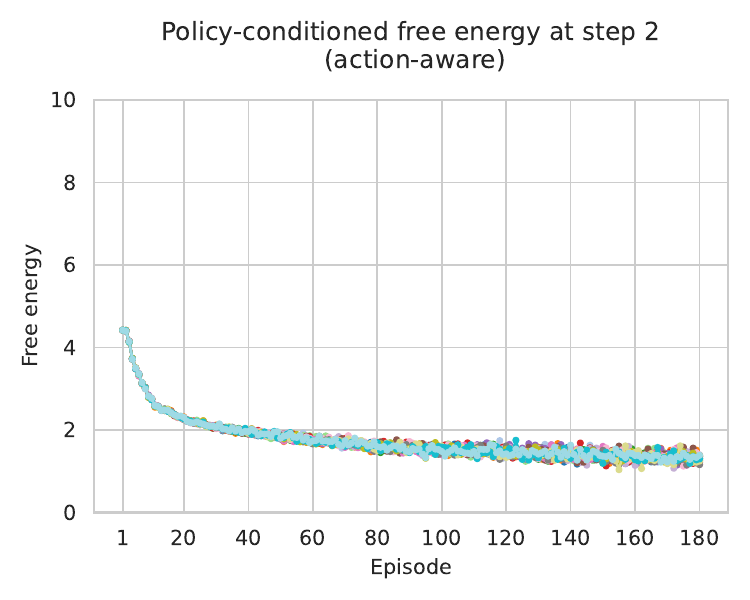}
  \end{subfigure}
    \begin{subfigure}{0.65\textwidth}
    \centering
    \includegraphics[width=\textwidth]{gridw9_aif_policies_legend}
  \end{subfigure}
  \caption{Policy-conditioned free energies at step 2 across episodes (showing average of 10 agents).}\label{fig:gridw9-policies-fes-step2}
\end{figure}

\begin{figure}[H]
  \centering
  \begin{subfigure}{0.45\textwidth}\label{fig:gridw9-aif-paths-policies-fe-step3}
    \centering
    \includegraphics[width=\textwidth]{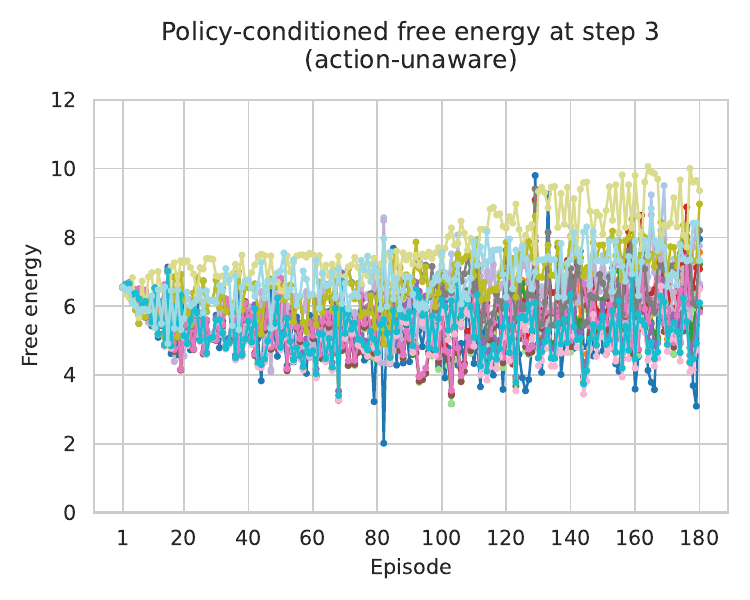}
  \end{subfigure}
  \begin{subfigure}{0.45\textwidth}\label{fig:gridw9-aif-plans-policies-fe-step3}
    \centering
    \includegraphics[width=\textwidth]{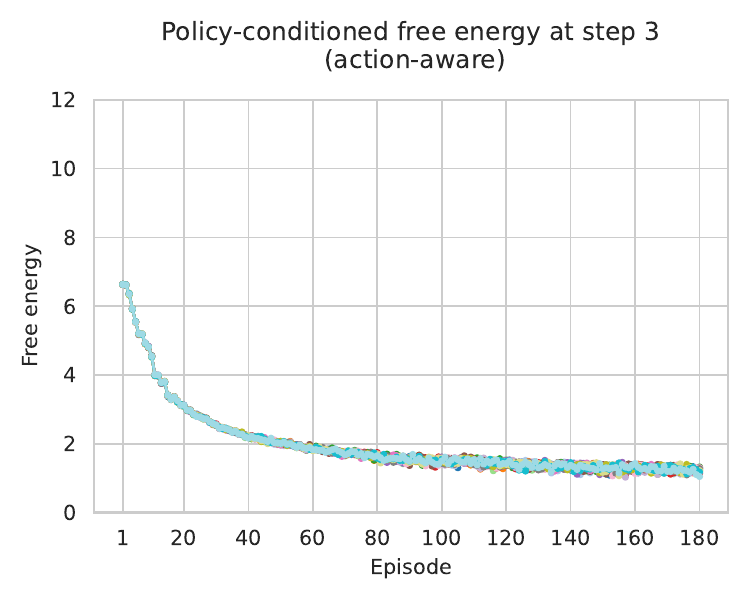}
  \end{subfigure}
    \begin{subfigure}{0.65\textwidth}
    \centering
    \includegraphics[width=\textwidth]{gridw9_aif_policies_legend}
  \end{subfigure}
  \caption{Policy-conditioned free energies at step 3 across episodes (showing average of 10 agents).}\label{fig:gridw9-policies-fes-step3}
\end{figure}

\begin{figure}[H]
  \centering
  \begin{subfigure}{0.45\textwidth}\label{fig:gridw9-aif-paths-policies-fe-step4}
    \centering
    \includegraphics[width=\textwidth]{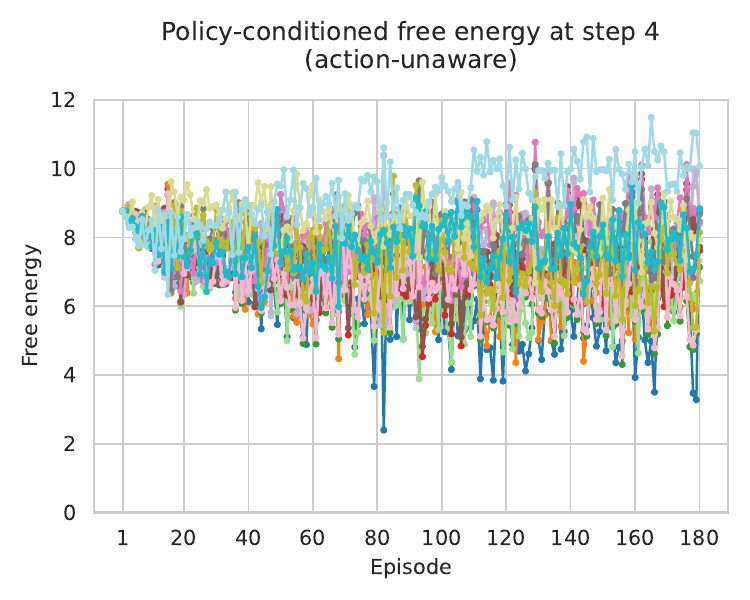}
  \end{subfigure}
  \begin{subfigure}{0.45\textwidth}\label{fig:gridw9-aif-plans-policies-fe-step4}
    \centering
    \includegraphics[width=\textwidth]{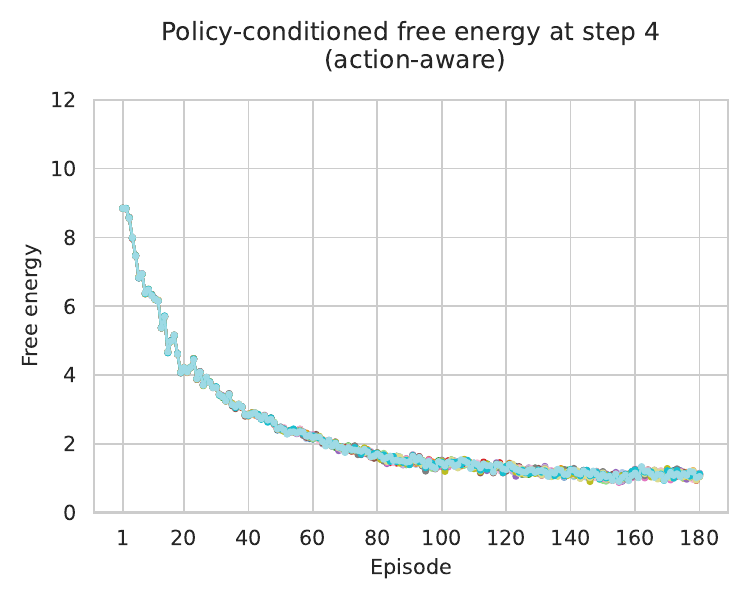}
  \end{subfigure}
    \begin{subfigure}{0.65\textwidth}
    \centering
    \includegraphics[width=\textwidth]{gridw9_aif_policies_legend}
  \end{subfigure}
  \caption{Policy-conditioned free energies at step 4 across episodes (showing average of 10 agents).}\label{fig:gridw9-policies-fes-step4}
\end{figure}

\subsubsection{Expected free energy at steps 2--4}\label{sssecx:gridw9-efe-steps}

\begin{figure}[H]
  \centering
  \begin{subfigure}{0.45\textwidth}\label{fig:gridw9-aif-paths-efe-step1}
    \centering
    \includegraphics[width=\textwidth]{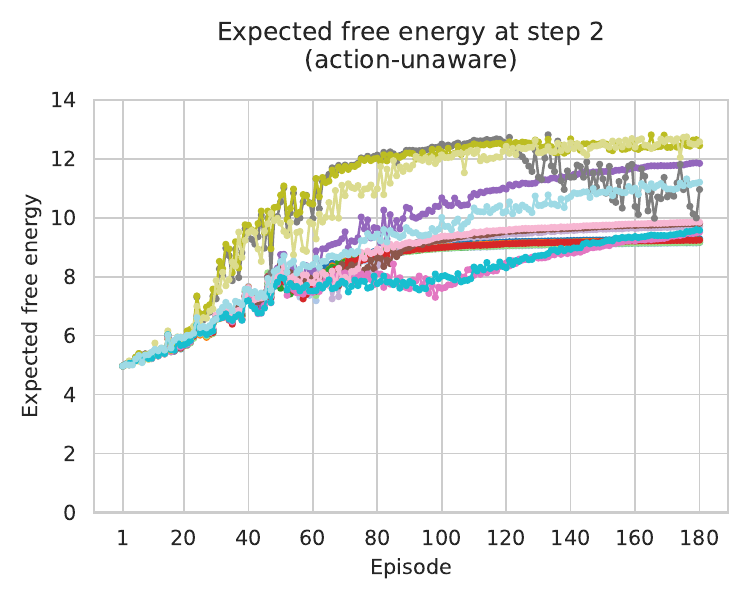}
  \end{subfigure}
  \begin{subfigure}{0.45\textwidth}\label{fig:gridw9-aif-plans-efe-step1}
    \centering
    \includegraphics[width=\textwidth]{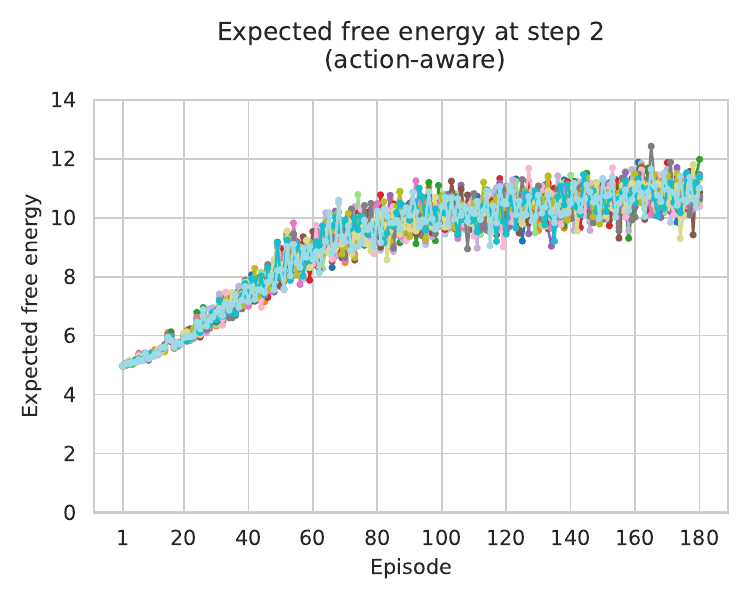}
  \end{subfigure}
    \begin{subfigure}{0.65\textwidth}
    \centering
    \includegraphics[width=\textwidth]{gridw9_aif_policies_legend}
  \end{subfigure}
  \caption{Expected free energy at step 2 for each policy across episodes (showing average of 10 agents).}\label{fig:gridw9-efe-step2}
\end{figure}

\begin{figure}[H]
  \centering
  \begin{subfigure}{0.45\textwidth}\label{fig:gridw9-aif-paths-efe-step2}
    \centering
    \includegraphics[width=\textwidth]{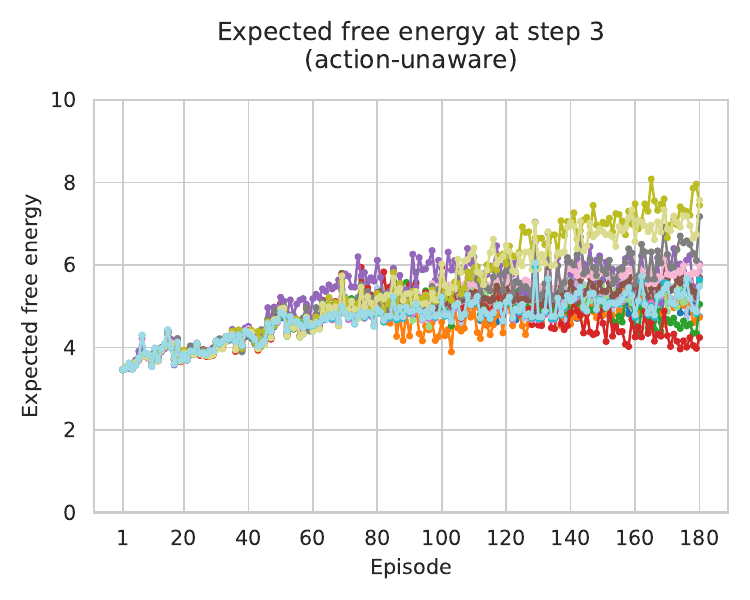}
  \end{subfigure}
  \begin{subfigure}{0.45\textwidth}\label{fig:gridw9-aif-plans-efe-step2}
    \centering
    \includegraphics[width=\textwidth]{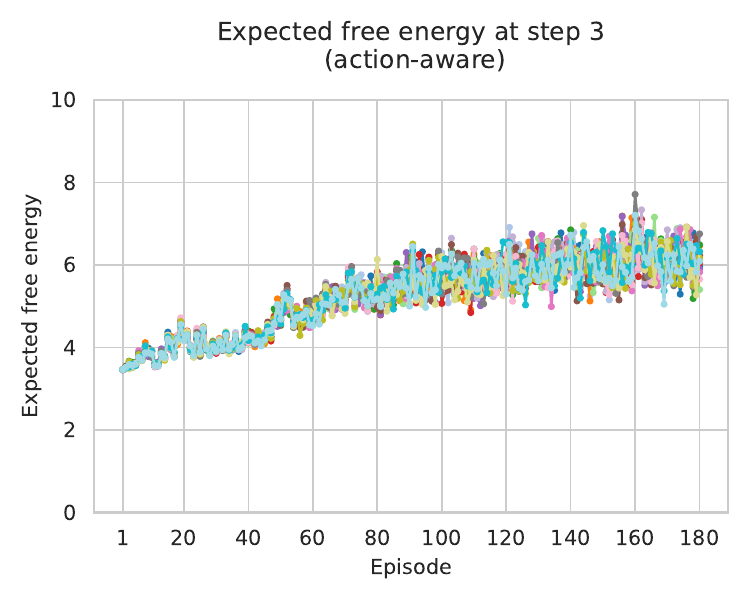}
  \end{subfigure}
    \begin{subfigure}{0.65\textwidth}
    \centering
    \includegraphics[width=\textwidth]{gridw9_aif_policies_legend}
  \end{subfigure}
  \caption{Expected free energy at step 3 for each policy across episodes (showing average of 10 agents).}\label{fig:gridw9-efe-step3}
\end{figure}

\begin{figure}[H]
  \centering
  \begin{subfigure}{0.45\textwidth}\label{fig:gridw9-aif-paths-efe-step3}
    \centering
    \includegraphics[width=\textwidth]{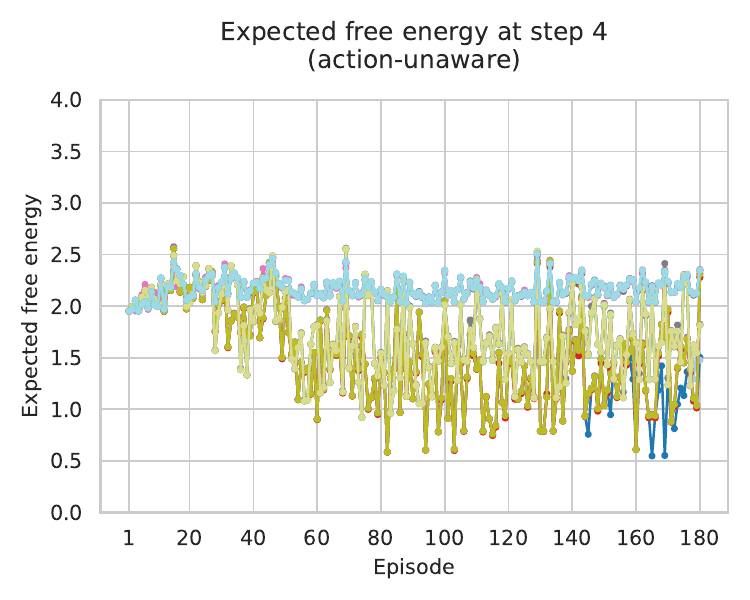}
  \end{subfigure}
  \begin{subfigure}{0.45\textwidth}\label{fig:gridw9-aif-plans-efe-step3}
    \centering
    \includegraphics[width=\textwidth]{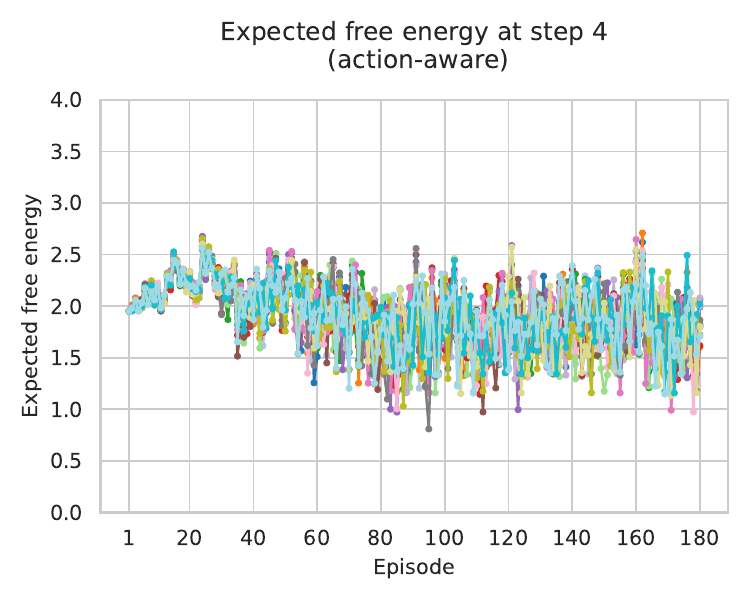}
  \end{subfigure}
    \begin{subfigure}{0.65\textwidth}
    \centering
    \includegraphics[width=\textwidth]{gridw9_aif_policies_legend}
  \end{subfigure}
  \caption{Expected free energy at step 4 for each policy across episodes (showing average of 10 agents).}\label{fig:gridw9-efe-step4}
\end{figure}

There is no expected free energy at step 5 because this is the step at which the environment terminates in the episodic setting considered in this work and, regardless of its location, the agent is no longer given the ability to plan forward in time.

\subsubsection{Expected free energy at step 0 breakdown}\label{sssecx:gridw9-efe-breakdown}

\subsubsection{Ground truth transition maps}\label{sssecx:gridw9-gtruth-trans-maps}

\begin{figure}[H]
\centering
\begin{subfigure}{0.45\textwidth}
    \centering
    \includegraphics[width=\textwidth]{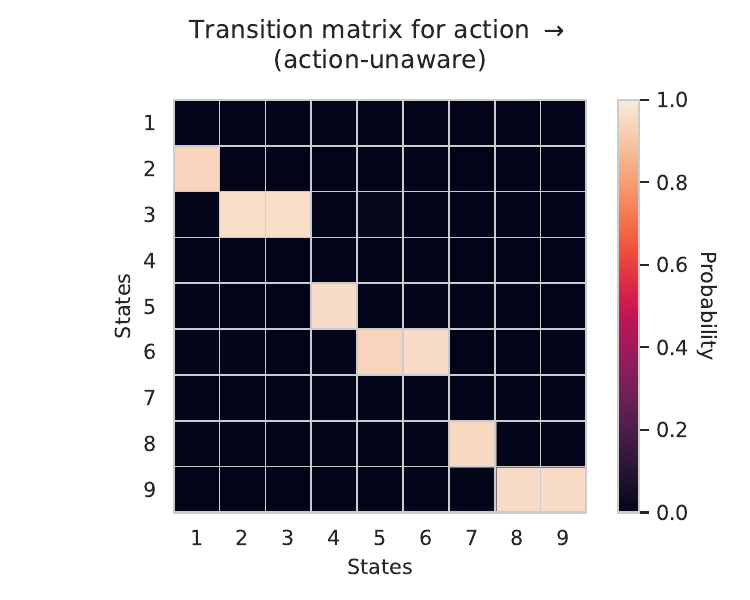}\label{fig:gridw9-gtruth-matrix-B-a0}
\end{subfigure}
\begin{subfigure}{0.45\textwidth}
    \centering
    \includegraphics[width=\textwidth]{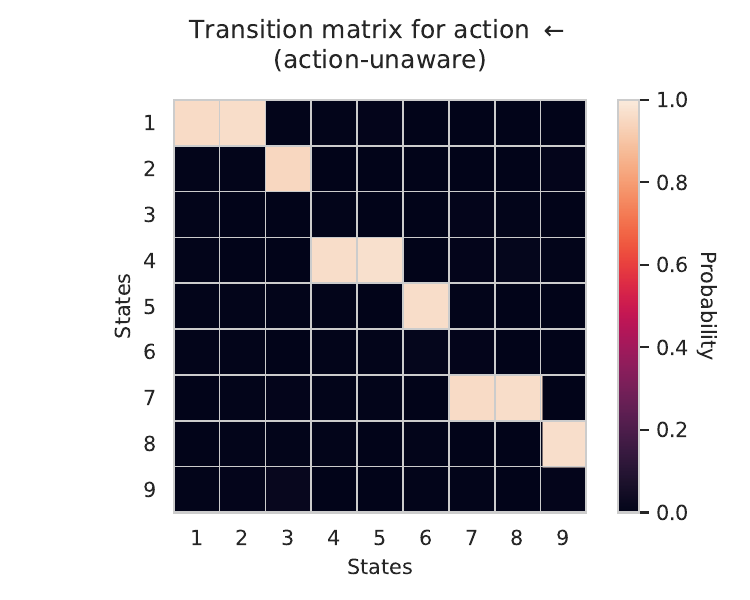}\label{fig:gridw9-gtruth-matrix-B-a2}
\end{subfigure}
\caption{Ground truth transition maps for action \(\rightarrow\) and \(\leftarrow\) in the grid world.}\label{fig:gridw9-gt-arl}
\end{figure}

\begin{figure}[H]
\centering
\begin{subfigure}{0.45\textwidth}
    \centering
    \includegraphics[width=\textwidth]{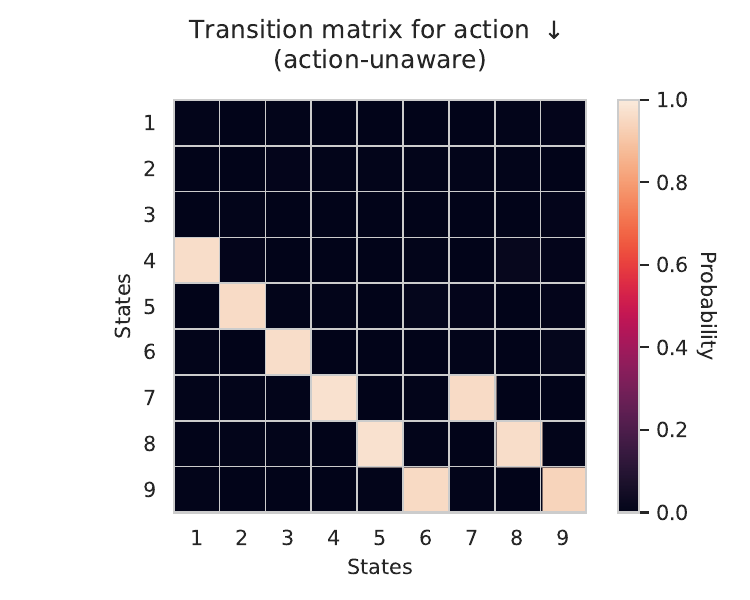}\label{fig:gridw9-gtruth-matrix-B-a1}
\end{subfigure}
\begin{subfigure}{0.45\textwidth}
    \centering
    \includegraphics[width=\textwidth]{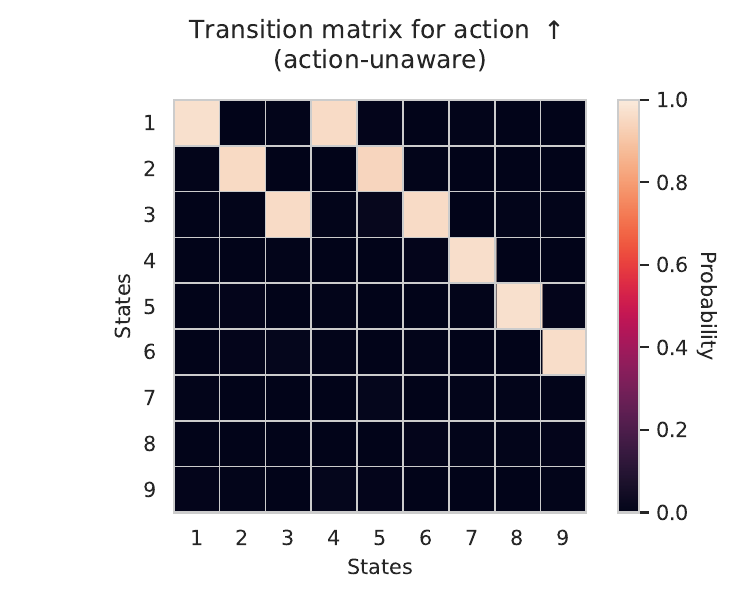}\label{fig:gridw9-gtruth-matrix-B-a3}
\end{subfigure}
\caption{Ground truth ransition maps for action \(\downarrow\) and \(\uparrow\) in the grid world.}\label{fig:gridw9-a13}
\end{figure}

\subsubsection{Learned transition maps in action-unaware and action-aware agents}\label{sssecx:gridw9-learned-trans-maps}

\begin{figure}[H]
\centering
\begin{subfigure}{0.45\textwidth}
    \centering
    \includegraphics[width=\textwidth]{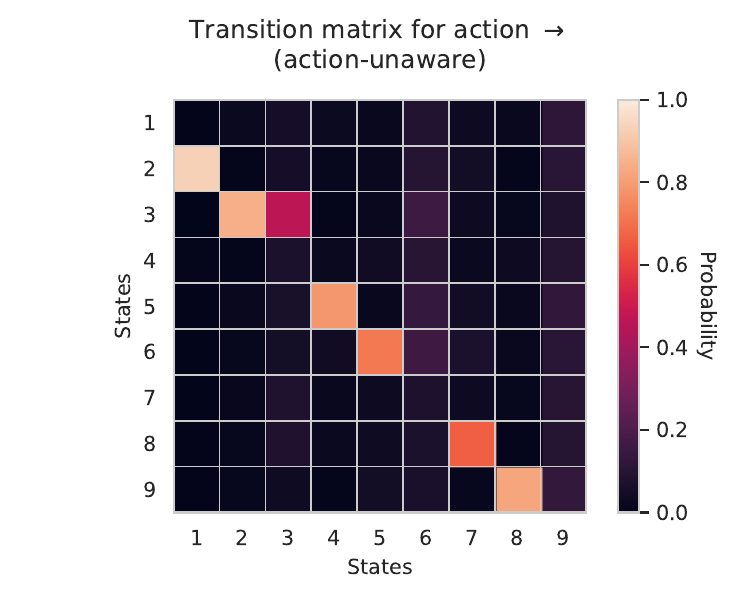}\label{fig:gridw9-aif-paths-matrix-B-a0}
\end{subfigure}
\begin{subfigure}{0.45\textwidth}
    \centering
    \includegraphics[width=\textwidth]{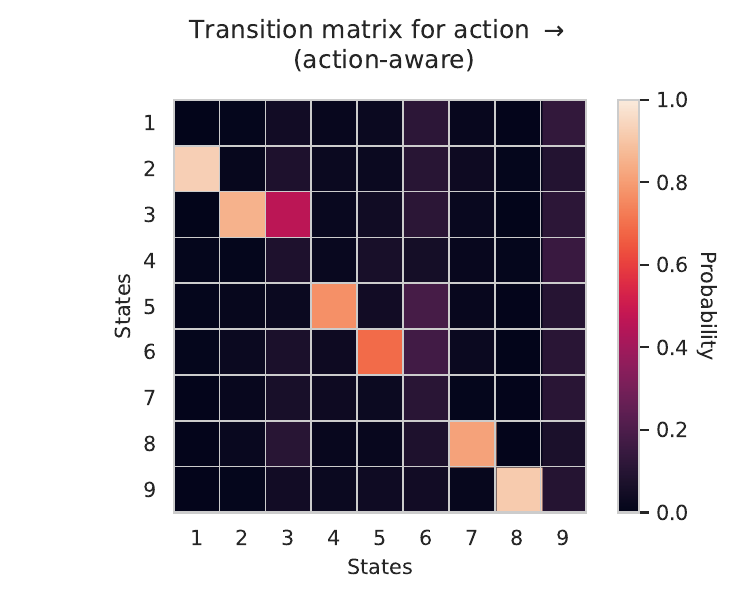}\label{fig:gridw9-aif-plans-matrix-B-a0}
\end{subfigure}
\caption{Transition maps for action \(\rightarrow\).}\label{fig:gridw9-a0r}
\end{figure}

\begin{figure}[H]
\centering
\begin{subfigure}{0.45\textwidth}
    \centering
    \includegraphics[width=\textwidth]{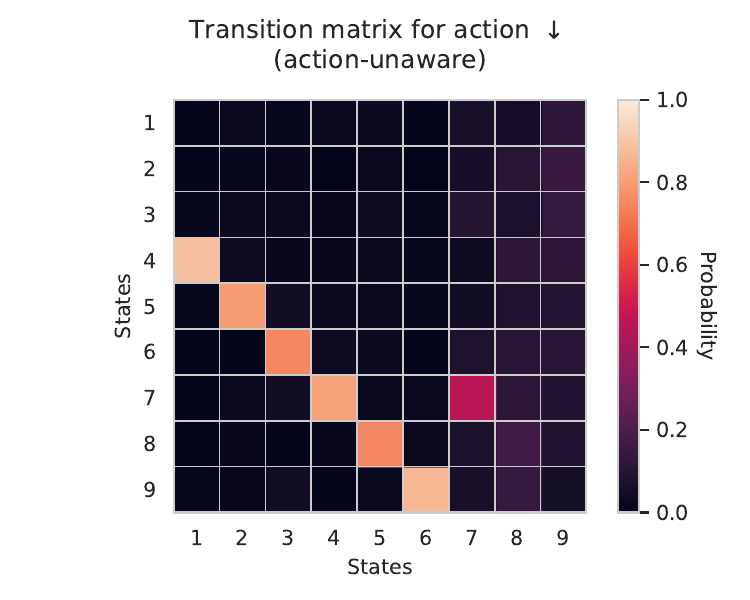}\label{fig:gridw9-aif-paths-matrix-B-a1}
\end{subfigure}
\begin{subfigure}{0.45\textwidth}
    \centering
    \includegraphics[width=\textwidth]{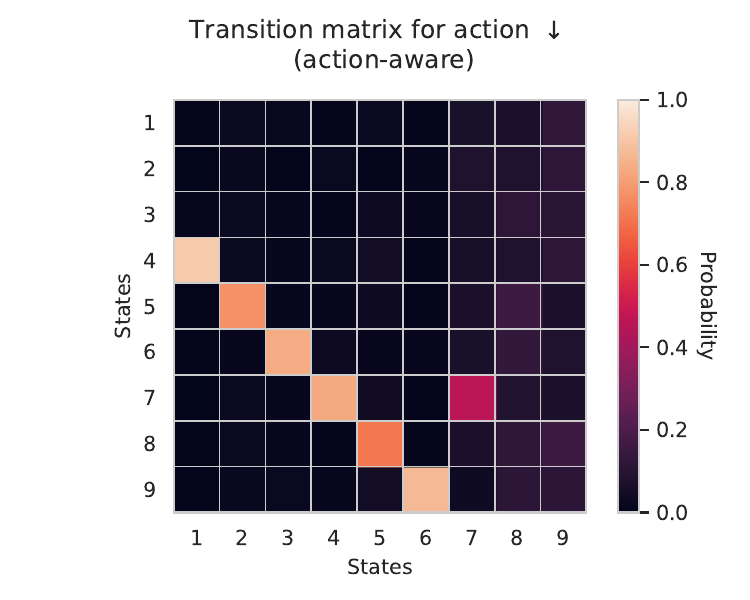}\label{fig:gridw9-aif-plans-matrix-B-a1}
\end{subfigure}
\caption{Transition maps for action \(\downarrow\).}\label{fig:gridw9-a1u}
\end{figure}

\begin{figure}[H]
\centering
\begin{subfigure}{0.45\textwidth}
    \centering
    \includegraphics[width=\textwidth]{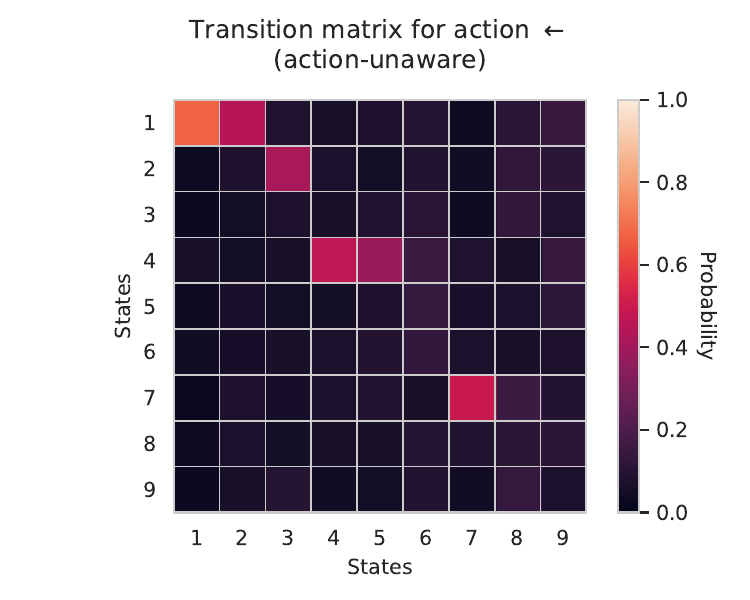}\label{fig:gridw9-aif-paths-matrix-B-a2}
\end{subfigure}
\begin{subfigure}{0.45\textwidth}
    \centering
    \includegraphics[width=\textwidth]{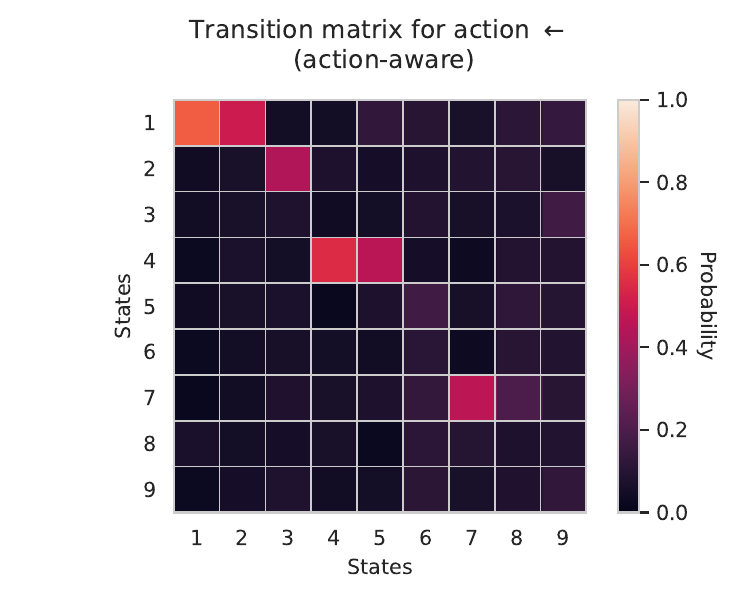}\label{fig:gridw9a-aif-plans-matrix-B-a2}
\end{subfigure}
\caption{Transition maps for action \(\leftarrow\).}\label{fig:gridw9-a2l}
\end{figure}

\begin{figure}[H]
\centering
\begin{subfigure}{0.45\textwidth}
    \centering
    \includegraphics[width=\textwidth]{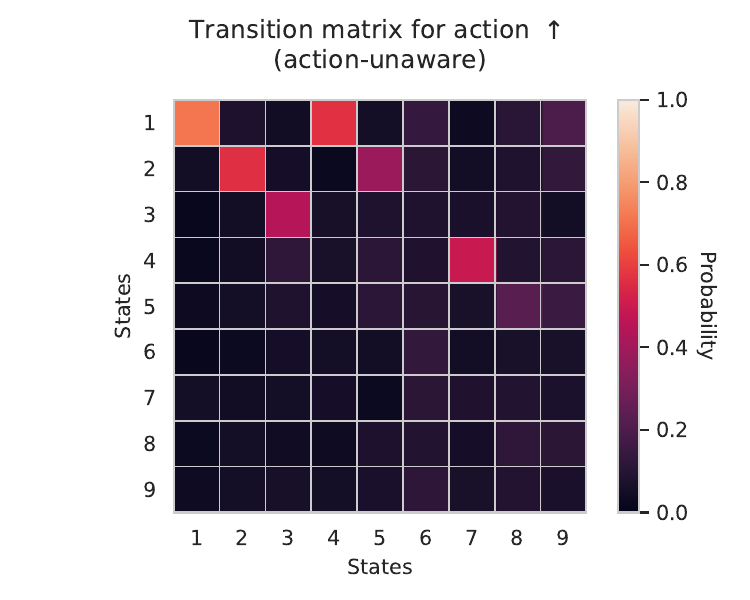}\label{fig:gridw9-transmap-path-a3}
\end{subfigure}
\begin{subfigure}{0.45\textwidth}
    \centering
    \includegraphics[width=\textwidth]{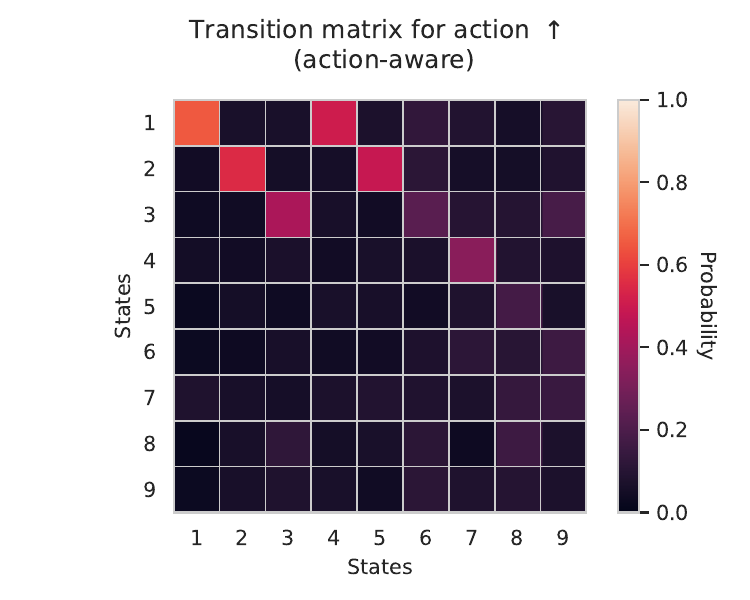}\label{fig:gridw9-aif-plans-a3}
\end{subfigure}
\caption{Transition maps for action \(\uparrow\).}\label{fig:gridw9-a3d}
\end{figure}

\end{document}